\documentclass[12pt]{article}

\usepackage{PRIMEarxiv}
\usepackage[utf8]{inputenc}    % Allow utf-8 input
\usepackage[T1]{fontenc}       % Use 8-bit T1 fonts
\usepackage{hyperref}          % Hyperlinks
\usepackage{url}               % Simple URL typesetting
\usepackage{booktabs}          % Professional-quality tables
\usepackage{amsfonts}          % Blackboard math symbols
\usepackage{amsmath}           % Advanced mathematical typesetting
\usepackage{graphicx}          % Graphics inclusion
\usepackage{caption}           % Custom captions
\usepackage{bm}                % Bold math symbols
\usepackage{subcaption}        % Sub-captions
\usepackage{geometry}          % Page geometry
\usepackage{multirow}          % Multi-row cells in tables
\usepackage{nicefrac}          % Compact symbols for 1/2, etc.
\usepackage{microtype}         % Microtypography
\usepackage{lipsum}            % Dummy text
\usepackage{fancyhdr}          % Custom headers and footers
\usepackage{enumitem}          % Custom lists
\usepackage{algorithm}         % Floating algorithms
\usepackage{bm}                % Bold math symbols
\usepackage{float}             % Improved floating objects
\usepackage{amsmath, amssymb}
\usepackage{graphicx}
\usepackage{tikz}
\usepackage{amsmath} % or \usepackage{mathtools}
\usepackage{makecell}

\usepackage{algorithm}
\usepackage{algpseudocode}
\usepackage{mathrsfs}

\usepackage{amsmath}
\usepackage{amssymb}
\usepackage{geometry}
\usepackage{hyperref}
\usepackage{amsfonts} % For mathbb
\usepackage{tikz} % For drawing diagrams
\usetikzlibrary{arrows.meta, positioning, 3d, decorations.pathreplacing, calligraphy}
\usepackage{mathtools}

\usepackage{amsmath}
\usepackage{amssymb} % For \therefore
\usepackage{tikz}
\usepackage{xcolor}
\usepackage[normalem]{ulem} % For strikethrough text with \sout
\usepackage{bm}

\usetikzlibrary{decorations.pathreplacing} % For the curly brace

\hypersetup{
    colorlinks=true,
    linkcolor=blue,
    filecolor=magenta,      
    urlcolor=cyan,
}

\title{eDCF: Estimating Intrinsic Dimension using Local Connectivity
}

\author{
    Dhruv Gupta, Aditya Nagarsekar, Vraj Shah\\
    Department of Computer Science and Information Systems \\
    BITS Pilani K K Birla Goa Campus\\
    403726, Goa, India\\
    \texttt{dhruvgupta110205@gmail.com, adityanagarsekar2108@gmail.com, vrajshah2785@gmail.com} \\
    \And
    Sujith Thomas\\
    Department of Computer Science and Information Systems \\
    BITS Pilani K K Birla Goa Campus\\
    403726, Goa, India\\
    \texttt{sujitht@goa.bits-pilani.ac.in} \\
}

\begin{document}

\maketitle

\begin{abstract}
Modern datasets often contain high-dimensional features exhibiting complex dependencies. To effectively analyze such data, dimensionality reduction methods rely on estimating the dataset's intrinsic dimension (id) as a measure of its underlying complexity. However, estimating id is challenging due to its dependence on scale: at very fine scales, noise inflates id estimates, while at coarser scales, estimates stabilize to lower, scale-invariant values. This paper introduces a novel, scalable, and parallelizable method called $e\mathcal{DCF}$, which is based on \emph{Connectivity Factor} $(\mathcal{CF})$, a local connectivity-based metric, to robustly estimate intrinsic dimension across varying scales. Our method consistently matches leading estimators, achieving comparable values of mean absolute error (MAE) on synthetic benchmarks with noisy samples. Moreover, our approach also attains higher exact intrinsic dimension match rates, reaching up to 25.0\% compared to 16.7\% for MLE and 12.5\% for TWO-NN, particularly excelling under medium to high noise levels and large datasets. Further, we showcase our method’s ability to accurately detect fractal geometries in decision boundaries, confirming its utility for analyzing realistic, structured data.
\end{abstract}

% keywords can be removed

\section{Introduction}

\subsection{A Brief Overview}

Topology, a branch of mathematics concerned with properties of space that remain invariant under continuous deformations such as stretching or bending (excluding tearing or gluing), has emerged as a powerful framework for analyzing complex data. In computer science and data analysis, topological methods provide insights into the structure and shape of high-dimensional datasets.
Topological or intrinsic dimension is defined as follows \cite{Devaney1992}:

\begin{itemize}
\item A set \( S \) has \emph{topological dimension} 0 if every point in \( S \) has arbitrarily small neighborhoods whose boundaries do not intersect \( S \).

\item For \( k \geq 1 \), \( S \) has topological dimension \( k \) if each point in \( S \) has arbitrarily small neighborhoods whose boundaries intersect \( S \) in a set of topological dimension \( k-1 \), and \( k \) is the smallest integer satisfying this property.
\end{itemize}

This inductive definition captures the intuitive notion of dimension by examining the structure of neighborhoods around points and how their boundaries meet the set.

One important application is in understanding intrinsic dimensionality. Under the manifold hypothesis \cite{FeffermanMitterNarayanan2016, Roweis2000, Tenenbaum2000}, high-dimensional data often lie near lower-dimensional manifolds embedded in ambient space. Estimating the intrinsic dimension of the low dimensional manifold is crucial for tasks such as dimensionality reduction, data visualization, and understanding data generation mechanisms, thereby helping mitigate the curse of dimensionality.

Another strength of topology is robustness to scale and deformation, captured by persistent homology \cite{EdelsbrunnerLetscherZomorodian2000}. Persistent homology tracks topological features - such as connected components, loops, and voids - across multiple spatial scales, identifying those that persist as meaningful structures distinct from noise. This multi-scale analysis obviates reliance on arbitrary scale parameters and ensures invariance to continuous transformations like bending or stretching.

Topological approaches also demonstrate resilience to noise. Although noise can introduce spurious, short-lived features, persistence-based methods effectively separate these from stable, significant topological signatures of the underlying data. This makes topological data analysis especially valuable for examining noisy real-world datasets.

Finally, topology contributes to characterizing fractal and complex geometric structures. Classical topology traditionally deals with integer-dimensional spaces, but extensions via measures such as Hausdorff and box-counting dimensions \cite{GneitingSevcikovaPercival2012} - estimable using topological techniques - allow quantification of fractal-like patterns with non-integer dimensions, common in natural phenomena and complex networks.

\subsection{Related Works and Differences\label{sec:relatedworks}}

\subsubsection{Literature Review}
Intrinsic dimension (ID) estimation is a foundational task in data analysis and machine learning, concerned with determining the minimum number of variables required to describe the underlying structure of high-dimensional data. This problem involves challenges such as robustness to noise, scalability to large datasets, and applicability to nonlinear or fractal manifolds. Recent literature has proposed diverse methods to address these issues, which are grouped according to their methodological principles, as below.

\textbf{Tangential and local geometric methods} exploit local properties of the data manifold, typically using nearest-neighbor statistics or tangent space approximations. The Maximum Likelihood Estimator (MLE) \cite{LevinaBickel2004} models the likelihood of distances within local neighborhoods, but its performance deteriorates in noisy environments or when intrinsic dimensionality is high \cite{LevinaBickel2004}. The Two Nearest Neighbors (TWO-NN) approach \cite{Facco2017} employs ratios of first and second nearest-neighbor distances, which makes it sample-efficient on smooth manifolds while maintaining sensitivity to perturbations \cite{Facco2017}. DANCo \cite{Ceruti2014} combines distance and angular distributions, capturing effects of curvature more effectively than MLE \cite{Ceruti2014}. Correlation Dimension, originally proposed in the Grassberger–Procaccia framework \cite{GrassbergerProcaccia1983}, estimates fractal dimension through scaling properties of correlation sums, making it well suited for complex or fractal geometries \cite{GrassbergerProcaccia1983}. Local PCA variants \cite{FeffermanMitterNarayanan2016} approximate tangent spaces by applying principal component analysis in neighborhoods, inferring dimension from eigenvalue spectra; however, these approaches are affected by manifold curvature and density heterogeneity \cite{GilbertONeill2023}.  

\textbf{Global and parametric approaches} rely on assumptions about the data's global structure. Principal Component Analysis (PCA) and other eigenspectrum-based techniques estimate intrinsic dimension from eigenvalue decay \cite{FeffermanMitterNarayanan2016}, making them computationally attractive though limited when applied to nonlinear or multi-scale manifolds \cite{FeffermanMitterNarayanan2016}. Projective methods \cite{Li2018} assume near-linear subspaces, which constrains their utility when manifolds exhibit curvature or heterogeneous scales \cite{GilbertONeill2023}.  

\textbf{Topological data analysis (TDA)} approaches provide a distinct framework by capturing shape and scale-invariant properties. The Persistent Homology Dimension (PHD) \cite{jaquette2020fractal} integrates local and global topological features through persistent homology, achieving robustness to noise while drawing upon a strong theoretical foundation that connects combinatorics and fractal geometry \cite{EdelsbrunnerLetscherZomorodian2000}.  

More recent advances leverage mathematical and physical modeling frameworks. \textbf{Diffusion-based and noise-perturbation approaches} \cite{Stanczuk2024} study how density and Jacobian ranks vary under noise \cite{Feng2021}, improving local estimations in realistic and noisy regimes \cite{Stanczuk2024}.  

\textbf{Specialized strategies} have also been proposed for discrete or highly sparse data. Intrinsic dimension estimation methods for discrete metrics \cite{Macocco2023} address the biases that emerge when continuous-space estimators are applied to non-continuous data spaces. Approaches based on packing numbers and combinatorial geometry \cite{Kegl2002} use geometric covering complexities to infer dimension without relying exclusively on neighborhood statistics \cite{CamastraStaiano2016}.

\subsubsection{Research Gap\label{sec:researchgap}}

\begin{itemize}
    \item \textbf{Research Gap 1: }Intrinsic dimension estimation in computerized spaces encounters challenges due to the discrete nature of data representations and the intrinsic scale- and perspective-dependence of topological features. Unlike continuous mathematical objects, point clouds comprise finite, often sparse, samples that inherently limit infinite resolution and connectivity assumptions. This limitation complicates the identification of neighbors and dimension, especially in high-dimensional settings.

    \item \textbf{Research Gap 2: }Thus, selecting an appropriate scale and neighborhood definition is non-trivial and critical for meaningful results. Existing approaches that rely on continuous metric spaces and naive distance thresholds often fail to generalize efficiently in high dimensions or handle scale ambiguity robustly, and often falter in high noise settings.

\end{itemize}

\subsubsection{Our Approach: \texorpdfstring{$e\mathcal{DCF}$}{eDCF} (Empirically-weighted Distributed Connectivity Factor)}

This paper introduces $e\mathcal{DCF}$, an empirically-weighted Distributed Connectivity Factor framework that addresses these challenges through several key innovations.

First, we employ a \textbf{grid-based neighbor framework}, where directions and neighbors are defined discretely via vector and set constructs instead of continuous distance metrics. This enables computationally efficient algorithms for neighbor generation and grid alignment. It also enables us to discretize space into independent subregions, enabling \textbf{fully parallel computation} of neighbor relations on separate grid cells without cross-dependencies. This facilitates scalable execution on large, high-dimensional datasets using multiprocessing and thread-based parallelism.

Instead of traditional scale selection, we introduce an alternative metric called \emph{Information Percentage (IP)}, which guides scale selection based on data coverage and information retention.

We also define a novel \emph{Connectivity Factor ($\mathcal{CF}$)}, a measure of local connectivity, based on the above gridded neighbor framework and the Information Percentage. We derive theoretical bounds for intrinsic dimension of manifolds using this $\mathcal{CF}$. Then, using a weighted average of this $\mathcal{CF}$, called \emph{Distributed Connectivity Factor ($\mathcal{DCF}$)}, we provide a theoretical method to estimate intrinsic dimension of arbitrary manifolds, under the assumption that the number of data points is sufficiently large.

To extend it to practical scenarios where the number of data points might be less, we use an empirical method built on $\mathcal{DCF}$, finally introducing $e\mathcal{DCF}$. This method robustly captures multi-scale connectivity while accommodating noise and sparsity typical in real-world data. It overcomes obstacles posed by scale sensitivity, neighborhood ambiguity, and computational complexity in high-dimensional intrinsic dimension estimation, and is validated through extensive synthetic manifold benchmarks against state-of-the-art methods like TWO-NN and MLE.

\section{Preliminaries and Problem Definition}

In this paper, we use the terms Intrinsic Dimension and Topological Dimension interchangeably.

\subsection{Notational Definition}

Basic Notations:

\begin{itemize}
    \item $\mathbb{N}$: Set of all natural numbers.
    \item $\mathbb{W}$: Set of all whole numbers.
    \item $\mathbb{Z}$: Set of all integers.
    \item $\mathbb{Z}^n$: $\{(x_1, x_2, ..., x_n) : x_i \in \mathbb{Z}, i \in \mathbb{N}, i \in [1, n]\}$
    \item $\mathbb{R}$: Set of all real numbers.
    \item $\mathbb{R}^+$: Set of all positive real numbers.
    \item $\mathbb{R}^n$: $\{(x_1, x_2, ..., x_n) : x_i \in \mathbb{R}, i \in \mathbb{N}, i \in [1, n]\}$
    \item $\sum$: Summation Operator
    \item $\cap$: Intersection over sets.
    \item $\cup$: Union over sets.
    \item $\bigcup$: Universal Union Operator, which unions over a family of sets.
    \item $\lvert{X}\rvert$: Cardinality of set X.
    \item $\setminus$: Set difference operator.
    \item $\| \mathbf{v} \|_0$: L0 Norm of a vector $\mathbf{v}$
    \item $argmax_t(W_t)$: The value of $t$ that maximizes $W_t$, where $W_t \in W$.
    \item $\delta_{xy}$: Kronecker Delta, which outputs 1 when x = y and 0 otherwise.
\end{itemize}

$IP$ - Information Percentage

$\mathcal{CF(S)}$ - Connectivity Factor of set $S$ :

Connectivity Factor is a measure which ranges between [0, 1] and $\mathcal{CF(S)} \in \mathbb{R}$

\begin{itemize}
    \item $\mathcal{CF}_m^n(S)$: Connectivity Factor of set $S$ with dimension $m$ embedded in ambient dimension $n$.

    \vspace{1.0em}

    \item ${}^L\mathcal{CF}_m^n$: Lower Bound of Connectivity Factor of an object with intrinsic dimension $m$ embedded in ambient dimension $n$.

    \item ${}^M\mathcal{CF}_m^n$: Middle Value of Connectivity Factor of an object with intrinsic dimension $m$ embedded in ambient dimension $n$.

    \item ${}^U\mathcal{CF}_m^n$: Upper Bound of Connectivity Factor of an object with intrinsic dimension $m$ embedded in ambient dimension $n$.

\end{itemize}

$S_{min}^m$ - Minimal Set of dimension $m$.

$S_{full}^m$ - Full Set of dimension $m$.

$S_{max}^{(n, m)}$ - Maximal Set of intrinsic dimension $m$, embedded in ambient dimension $n$.
\begin{itemize}
    \item $S_l^n$: Labeled Set of dimension $n$.
    \item $S_{l0}^n$: Labeled Set of dimension $n$ containing only 0 labeled points.
    \item $S_{l0}^{n'}$: Labeled Set of dimension $n$ containing all labeled points except the points labeled 0.
    \item $S_{lt}^n$: Labeled Set of dimension $n$ containing only $t$ labeled points.
\end{itemize}

$a_t$ - The number of neighbors of type $t$, which are in $S_{min}^m$.

$\alpha^m_t$ - The number of type $t$ points in set $S_{min}^m$.

${}^x\alpha^n_y$ - The number of type $y$ labeled points in the neighborhood of type $x$ labeled point, in set $S_l^n$.

$\Omega^n_x$ - System Space of dimension $x$ constructed from decomposing an $n$ dimensional space.

$\omega^r_k$ - Subsystem of dimension $r$ with a mapping from type $t \to t + k$.

\vspace{1.0em}

$\mathcal{N}(x) - \text{Neighbor Set of a Point }x$ (the set of immediate neighbors of a point in a given dimension according to a gridded space)

\vspace{1.0em}

$\mathcal{LMU-CF}$ - LMU Bound based Connectivity Factor

$\mathcal{DCF}$ - Distributed Connectivity Factor

$e\mathcal{DCF}$ - Empirically - Weighted Distributed Connecitivity Factor

$W$ - Vector of weights $W_i$

$F(x)$ - Vector of fraction functions (linear cap, gaussian, etc.).

$\hat{m}$ - Estimated Intrinsic Dimension.

\subsection{Problem Definition}

Let $X = \{x_1, x_2, \dots, x_N\} \subset \mathbb{R}^n$ be a finite set of $N$ data points, where each point $x_i = (x_{i1}, x_{i2}, \dots, x_{iN})$ represents an object.\\
\vspace{0.1em}Assume the points in $X \subset \mathbb{R}^n$ lie approximately on a low-dimensional topological space $M \subset \mathbb{R}^m$.\\
\vspace{0,1em}Let the intrinsic dimension of $M$ be $m = \dim(M)$, where $m \leq n$.

The problem is twofold:
\begin{enumerate}
    \item \textbf{Theoretical Derivation:} Derive bounds for a functional metric $f$ such that for a topological dimension $i$, the bounds are $L_i \leq f(X) \leq U_i$, where $L_i$ and $U_i$ are the minimum and maximum possible values of the function for an object of topological dimension $i$. The objective is to show that as the number of data points $N \to \infty$, the value of $f(X)$ provides a highly accurate estimate of the intrinsic dimension $m$.
    \item \textbf{Empirical Estimation:} Empirically estimate the intrinsic dimension $m$ from the given data set $X$.
\end{enumerate}

\section{Proposed Method}

\subsection{Formal Definition of Grid Neighbors}

In the context of an $n$-dimensional integer grid, $\mathbb{Z}^n$, we define the neighbors of a point based on the principle of single-step adjacency in all axial and diagonal directions. This set of neighbors is formally known as the \textbf{Moore neighborhood} \cite{Moore1962}. The definition leverages a linear combination of the grid's standard basis vectors to systematically characterize this relationship.

\subsubsection{Mathematical Formulation}

A point $\mathbf{p} \in \mathbb{Z}^n$ is considered a neighbor of a point $\mathbf{q} \in \mathbb{Z}^n$ if their displacement vector, $\mathbf{p} - \mathbf{q}$, is a non-zero vector that can be expressed as:
\begin{equation}
\mathbf{p} - \mathbf{q} = \sum_{i=1}^{n} c_i \mathbf{v}_i
\end{equation}
where:
\begin{itemize}
    \item $\{ \mathbf{v}_1, \mathbf{v}_2, \ldots, \mathbf{v}_n \}$ is the standard basis for $\mathbb{Z}^n$.
    \item The coefficients $c_i$ are drawn from the set $\{-1, 0, 1\}$. Each coefficient represents a single step backward, no step, or a single step forward along the $i$-th axis.
\end{itemize}

The condition that the displacement vector must be non-zero explicitly excludes the case where $\mathbf{p} = \mathbf{q}$, which corresponds to the trivial coefficient tuple $(c_1, \ldots, c_n) = (0, \ldots, 0)$.

\subsubsection{Worked Example in \texorpdfstring{$\mathbb{Z}^2$}{Zexp(2)}}

For a 2-dimensional grid, the basis vectors are $\mathbf{v}_1 = (1, 0)$ and $\mathbf{v}_2 = (0, 1)$. For a given point $\mathbf{q} = (x, y)$, its neighbors are determined by the 8 possible non-zero coefficient tuples $(c_1, c_2)$.

For instance, the top-left neighbor, $\mathbf{p} = (x-1, y+1)$, is generated by selecting $c_1 = -1$ (one step left) and $c_2 = 1$ (one step up):
\[
\mathbf{p} = \mathbf{q} + (-1)\mathbf{v}_1 + (1)\mathbf{v}_2 = (x, y) + (-1, 0) + (0, 1) = (x-1, y+1)
\]

The complete neighborhood is illustrated below.

\begin{minipage}{\textwidth}
 \centering
 \begin{tikzpicture}[scale=3]
 % Grid outline
 \draw[black, thick, step=1cm] (0,0) grid (3,3);

 % Center node
 \node at (1.5, 1.5) {$\mathbf{q}(x, y)$};

 % Neighbors of q
 \node at (0.5, 2.5) {$\mathbf{p}(x-1, y+1)$};
 \node at (1.5, 2.5) {$\mathbf{p}(x, y+1)$};
 \node at (2.5, 2.5) {$\mathbf{p}(x+1, y+1)$};
 \node at (0.5, 1.5) {$\mathbf{p}(x-1, y)$};
 \node at (2.5, 1.5) {$\mathbf{p}(x+1, y)$};
 \node at (0.5, 0.5) {$\mathbf{p}(x-1, y-1)$};
 \node at (1.5, 0.5) {$\mathbf{p}(x, y-1)$};
 \node at (2.5, 0.5) {$\mathbf{p}(x+1, y-1)$};
 \end{tikzpicture}
\end{minipage}

\subsubsection{Enumeration of Neighbors}

Having established the definition, we can determine the number of neighbors for any point in $\mathbb{Z}^n$ through a combinatorial argument.

Each of the $n$ coefficients, $c_i$, can independently take one of three possible values from the set $\{-1, 0, 1\}$. By the rule of product, the total number of unique coefficient tuples is:
\[
\underbrace{3 \times 3 \times \cdots \times 3}_{n \text{ times}} = 3^n
\]
This figure represents all possible points reachable under the formulation, including the point $\mathbf{q}$ itself (via the zero-displacement vector where all $c_i=0$). As a point is not its own neighbor, we exclude this single case.

Therefore, the number of neighbors for any point in $n$-dimensional space is 
$\boxed{3^n - 1}$.

\subsubsection{Generalization to Discretized Real Space}

This framework can be extended from the integer grid $\mathbb{Z}^n$ to a discretized real space $\mathbb{R}^n$ by introducing a uniform grid spacing, $s \in \mathbb{R}^+$. A point $\mathbf{p} \in \mathbb{R}^n$ is a neighbor of $\mathbf{q} \in \mathbb{R}^n$ if:
\begin{equation}
\boxed{
\mathbf{p} = \mathbf{q} + s \sum_{i=1}^{n} c_i \mathbf{v}_i
}
\end{equation}
where $c_i \in \{-1, 0, 1\}$, and not all $c_i$ are zero. This discretizes the continuous space into a uniform lattice where the neighborhood properties remain identical.

\newpage

For example:
Let, $n = 2$, $q = (1, 1)$, $s=0.5$:

Standard basis vectors(in tuple format) for $\mathbb{R}^2$ are $(1, 0)$ and $(0, 1)$.

\begin{minipage}{\textwidth}
 \centering
 \begin{tikzpicture}[
     scale=3 % Increased scale for better text accommodation
 ]
 % Grid outline is black and thick
 \draw[black, thick, step=1cm] (0,0) grid (3,3);

 % Center node
 \node at (1.5, 1.5) {{\boldmath$q(1, 1)$}};

 % Neighbors of q
 % Top row
 \node at (0.5, 2.5) {{\boldmath$p(0.5, 1.5)$}};
 \node at (1.5, 2.5) {{\boldmath$p(1, 1.5)$}};
 \node at (2.5, 2.5) {{\boldmath$p(1.5, 1.5)$}};

 % Middle row
 \node at (0.5, 1.5) {{\boldmath$p(0.5, 1)$}};
 \node at (2.5, 1.5) {{\boldmath$p(1.5, 1)$}};

 % Bottom row
 \node at (0.5, 0.5) {{\boldmath$p(0.5, 0.5)$}};
 \node at (1.5, 0.5) {{\boldmath$p(1, 0.5)$}};
 \node at (2.5, 0.5) {{\boldmath$p(1.5, 0.5)$}};

 \end{tikzpicture}
\end{minipage}

\vspace{1.0em}

\subsection{Information Percentage}
Real-world data rarely presents in a perfectly structured, grid-like format. Therefore, we employ a preprocessing step to transform the data to impose such a structure. For this purpose, we define a metric called Information Percentage ($IP$) to get a grid with a specific desired property discussed ahead.

\subsubsection{Global Normalization}
 Normalization is important to make sure that we are not dealing with spacing values which are arbitrarily large.

\vspace{1em}

Global normalization is employed to ensure that the relative scale and relationships between features are preserved. This is achieved by scaling all features by a single, common factor derived from the entire dataset. This factor is the maximum range observed across all individual features.

\vspace{1em}

The procedure is as follows:
\begin{enumerate}
    \item For each feature vector $\mathbf{X}_j \in X$ , calculate its range: \(R_j = x_{\max, j} - x_{\min, j}\).
    \item Identify the maximum of these ranges. This value becomes the global scaling factor: \(R_{\max} = \max(R_1, R_2, \dots, R_N)\).
\end{enumerate}

\vspace{1em}

Each component \(x\) of a given feature vector \(\mathbf{X}_j\) is then transformed into \(x'\) using the formula:
\begin{equation}
x' = \frac{x - x_{\min, j}}{R_{\max}}
\end{equation}
Here, \(x_{\min, j}\) is the minimum value of the specific feature \(j\) to which \(x\) belongs.

\vspace{1em}

This transformation results in the data being bounded within a unit hypercube. The feature that originally had the largest range will now be scaled to occupy the full interval [0, 1]. All other features will be scaled by the same amount, occupying a proportionally smaller sub-interval within [0, 1]. This preserves the ratio between the ranges of any two features after normalization.

\vspace{1em}

Formally, for any normalized n-dimensional data point \(\mathbf{x'} = (x'_1, x'_2, \dots, x'_N)\), each component is guaranteed to be within the range of 0 to 1:
\[ \forall \mathbf{x'} \in \mathbb{R}^n, \mathbf{x'} = (x'_1, x'_2, \dots, x'_N) \text{ such that } 0 \le x'_i \le 1 \text{ for } i = 1, 2, \dots, N. \]

\subsubsection{Defining the Grid from Raw Data}

Let \(X\) be the raw data set, where \(X\) is a collection of points in an n-dimensional space (\(X \subset \mathbb{R}^n\)). We aim to create a new set of points, \(S\), by mapping each point from \(X\) to the center of a corresponding cell in a discrete, uniform grid.

\vspace{1em}

This grid is characterized by the spacing \(s\), \(s > 0\), which defines the uniform size (or resolution) of the grid cells along each axis. The grid is centred at the origin.

\vspace{1em}

For every point \(\mathbf{x} = (x_1, x_2, \dots, x_N)\) in the original dataset \(X\), we compute a new point \(\mathbf{p} = (p_1, p_2, \dots, p_N)\). The coordinates of \(\mathbf{p}\) are calculated component-wise using the following transformation for each dimension \(i\):
\begin{equation}
p_i = \left\lfloor \frac{x_i}{s} \right\rfloor \cdot s + \frac{s}{2}
\end{equation}
\vspace{1em}

This operation effectively ``snaps'' each raw data point to the center of its enclosing grid cell. The term \(\lfloor x_i / s \rfloor\) identifies the integer index of the grid cell, which is then multiplied by \(s\) to find the lower boundary (or ``floor'') of that cell. Adding \(s/2\) positions the new point precisely in the center of the cell.

\vspace{1em}

The resulting set \(S\) is the collection of all such grid center points derived from the points in \(X\). Formally, it is defined as:
\begin{equation}
\boxed{
S = \left\{ \mathbf{p} \in \mathbb{R}^n \mid \forall \mathbf{x} \in X \text{ such that } p_i = \left\lfloor \frac{x_i}{s} \right\rfloor \cdot s + \frac{s}{2} \text{ for } i=1, \dots, N \right\}
}
\end{equation}

\subsubsection{Mathematical Definition and Information Retention}

The process of converting raw data to a grid-based format is a form of quantization. A direct consequence of this is that multiple, distinct raw data points can be mapped to the exact same grid point. This phenomenon, where unique points become indistinguishable after transformation, represents a potential loss of information.

\vspace{1em}

The degree of this information loss is fundamentally controlled by the grid spacing parameter, \(s\).
\begin{itemize}
    \item A \textbf{large} value of \(s\) creates a coarse grid, increasing the likelihood that many raw data points will fall within the same cell and map to a single grid point. This leads to a significant reduction in data complexity but at the cost of losing detail.
    \item A \textbf{small} value of \(s\) creates a fine grid, reducing the number of these ``collisions'' and thus preserving more of the original data's granularity.
\end{itemize}

\vspace{1em}

Therefore, the amount of information preserved can be effectively measured by the number of unique grid points generated after the conversion. A higher count of unique grid points signifies greater information retention.

\vspace{1em}

Let \(X\) be the original, raw data set and \(S\) be the set of unique grid points obtained after applying the grid conversion process. The \emph{Information Percentage (IP)} can be defined as a metric to quantify this retention:
\begin{equation}
\boxed{
\text{IP} = \frac{|S|}{|X|} \times 100
}
\end{equation}
Here, \(|S|\) is the cardinality (number of unique elements) of the set \(S\), and \(|X|\) is the cardinality of the original set \(X\). An IP of 100\% would imply that every raw data point mapped to a unique grid point, indicating no information was lost through quantization.

\vspace{1em}

\subsubsection{Search Method for Optimal Grid Spacing}

Having established the relationship between the \emph{Information Percentage (IP)} and the grid spacing (\(s\)), we can now define a procedure to find an appropriate value of \(s\) that aligns with a desired level of data granularity.

\vspace{1em}

The user specifies a target IP range, \([IP_{\min}, IP_{\max}]\).
\begin{itemize}
    \item A \textbf{lower IP} is chosen when the goal is to capture the general, high-level structure of the data, effectively smoothing out noise and minor variations by mapping more points to the same grid cell.
    \item A \textbf{higher IP} is chosen when preserving fine details and maximizing data fidelity is critical, as this corresponds to a finer grid where fewer points are consolidated.
\end{itemize}

\vspace{1em}

The search for the optimal \(s\) is guided by the fundamental property that the IP is monotonically non-decreasing as \(s\) decreases. This allows for an efficient search. Since the data has been normalized to a unit hypercube, the maximum possible spacing is 1, providing a natural starting point for our search.

\vspace{1em}

The search employs a coarse-to-fine strategy:
\begin{enumerate}
    \item \textbf{Phase 1: Coarse Search (Galloping/Exponential Search)} \\
    This phase quickly identifies a promising interval for \(s\). Starting with an initial coarse spacing (e.g., \(s = 1\)), we iteratively decrease \(s\) (for instance, by dividing it by 5 at each step) until the calculated IP meets or exceeds the lower bound of our target range, \(IP_{\min}\). This rapidly narrows the search space from \([0, 1]\) to a much smaller, relevant bracket.

    \item \textbf{Phase 2: Fine Search (Binary Search)} \\
    Once the coarse search has found a range \([s_{\text{low}}, s_{\text{high}}]\) that brackets the target IP, a binary search is performed within this interval. This allows for an efficient and precise identification of a spacing \(s\) that yields an IP within the desired \([IP_{\min}, IP_{\max}]\) range. Iteration limits can be used to guarantee termination.
\end{enumerate}

\vspace{1em}

This two-phase approach ensures that we can efficiently find a grid spacing \(s\) that balances data simplification with information retention, tailored to the specific requirements of the analysis.

\subsection{Connectivity Factor\label{sec:Connectivity Factor}}

\subsubsection{Formal Definition\label{sec:ConnectivityFactorformal}}

Let $S = \{s_1, s_2, \dots, s_N\} \subset \mathbb{R}^n$ be a finite set of $N$ points in an $n$-dimensional ambient space. Each point $s_i$ is a vector of dimension $n$, denoted as $s_i = (s_{i1}, s_{i2}, \dots, s_{in})$ for $i \in \{1, 2, \dots, N\}$. Thus, $s_{ij}$ represents the $j$-th feature of the $i$-th data point.

For each point $s_i \in S$, we define its neighborhood, $\mathcal{N}(s_i)$, as the set of all its neighboring points:
\begin{equation}
\mathcal{N}(s_i) = \{y \in \mathbb{R}^n \mid y \text{ is a neighbor of } s_i\}
\end{equation}
As established previously, the cardinality of the neighborhood for any point in an $n$-dimensional space is given by:
\begin{equation}
|\mathcal{N}(s_i)| = 3^n - 1
\end{equation}

The \emph{Connectivity Factor} of the set $S$, denoted as $\mathcal{CF}(S)$, is defined as the average proportion of neighbors for each point in $S$ that are also members of $S$. This is formally expressed as:
\[
\mathcal{CF}(S) = \frac{1}{N} \sum_{i=1}^{N} \frac{|\mathcal{N}(s_i) \cap S|}{|\mathcal{N}(s_i)|}
\]
Substituting the constant cardinality of the neighborhood, the formula can be simplified to:
\begin{equation}
\boxed{
\mathcal{CF}(S) = \frac{\sum_{i=1}^{N} |\mathcal{N}(s_i) \cap S|}{N \cdot (3^n - 1)}
}
\end{equation}

\subsubsection{Space Conversion Formula:}

% This is a comment below
\iffalse
CF is not invariant under rotation. This is because the grid structure of points is broken down when we use a Rotation transformation on the points. Thus CF cannot be transferred to a higher dimension if the object is rotated in the higher space wrt to the m axes. It is however possible to change the location of the object in the grid, however we see fit via a transformation. This causes the same set to have a different connectivity factor if rotated since the points do not fit in the old lattice when it comes to rotation. We mitigate this by using Forced Gridding and 50\% IP which helps us get a good estimate of the structure of points without assuming any structure. Largely, the CF of the two structures (original and rotated) would be the same under a large number of points.
\fi

The connectivity factor of objects naturally changes, for the same object when embedded in a higher dimension due to the change in the number of maximum possible neighbours.

Let $S = \{s_1, s_2, \dots, s_N\} \subset \mathbb{R}^m$ be a finite set of $N$ points in an $m$-dimensional ambient space. Each point $s_i$ is a vector of dimension $m$, denoted as $s_i = (s_{i1}, s_{i2}, \dots, s_{im})$ for $i \in \{1, 2, \dots, N\}$. Thus, $s_{ij}$ represents the $j$-th feature of the $i$-th data point.

Therefore as previously defined, $\mathcal{CF}(S) = \frac{\sum_{i=1}^{N} |\mathcal{N}(s_i) \cap S|}{N \cdot (3^m - 1)}$

Let $v_1, \dots, v_n \in \mathbb{R}^n$ be a set of standard basis vectors of $\mathbb{R}^n$. An affine transformation $T: \mathbb{R}^m \to \mathbb{R}^n$, where $m \le n$ is defined as:
\begin{equation*}
T(s) = As + b
\end{equation*}
where the matrix $A \in \mathbb{R}^{n \times m}$ and the vector $b \in \mathbb{R}^n$ are given by:
\begin{equation*}
A := \begin{pmatrix} | & | & & | \\ v_1 & v_2 & \cdots & v_m \\ | & | & & | \end{pmatrix}
\end{equation*}

The set $S' \subset \mathbb{R}^n$ is the image of $S$ under the transformation $T$:
\begin{equation*}
S' := \{T(s) \mid s \in S\} = \{As + b \mid s \in S\}
\end{equation*}

Now we define a notation to represent the \emph{Connectivity Factor} of $S'$ composed of the transformed points of the set $S$, which is representative of mapping the structure from an $m$ dimensional space to an $n$ dimensional space.

$\mathcal{CF}_m^n(S) = \mathcal{CF}(S')$

This notation also implies that, $\mathcal{CF}_m^m(S) = \mathcal{CF}(S)$, since $S = S'$.

Due to the nature of the transformation, we preserve the grid properties such as the neighbor connections of the set $S$ in set $S'$. Since the neighbors of points are preserved and there is no addition of points in set $S'$, we can say that $\lvert \mathcal{N}(s') \cap S' \rvert = \lvert \mathcal{N}(s) \cap S \rvert$.

Therefore for the transformation $T: S \to S'$:

\[
  \mathcal{CF}^n_m(S) = \mathcal{CF}(S') = \frac{\sum_{i=1}^{N} |\mathcal{N}(s_i') \cap S'|}{N \cdot (3^n - 1)} = \frac{3^m - 1}{3^m - 1} \cdot \frac{\sum_{i=1}^{N} |\mathcal{N}(s_i) \cap S|}{N \cdot (3^n - 1)}
\]

\[
  \implies \mathcal{CF}^n_m(S) = \frac{3^m - 1}{3^n - 1} \cdot \frac{\sum_{i=1}^{N} |\mathcal{N}(s_i) \cap S|}{N \cdot (3^m - 1)} = \frac{3^m - 1}{3^n - 1} \cdot \mathcal{CF}_m^m(S)
\]

Here, $0 \le m \le n$, and $n > 0$. When $n=0$, the denominator $3^n - 1 = 0$, and we define this base case explicitly as
\[
  \mathcal{CF}^0_0 = 1.0,
\]
(note that when $n = 0$, since we have $m \le n$, thus $m = 0$).

Therefore, the final form of $CF_m^n(S)$ is as follows, 

\begin{equation}
\boxed{
\mathcal{CF}^n_m(S) =
\begin{cases}
    \dfrac{3^m - 1}{3^n - 1} \cdot \mathcal{CF}_m^m(S), & \text{if } 0 \leq m \leq n, n > 0 \\[8pt]
    1, & \text{if } n = 0, m = 0
\end{cases}, \quad \text{where } m, n \in \mathbb{W}.
}
\end{equation}

\subsubsection{Minimal Set in \texorpdfstring{$m$}{m} Dimensions} \label{minimalsetdef}
A \emph{minimal set} in $m$-dimensional space is defined as the smallest collection of points on an $m$-dimensional grid that satisfies the following conditions: 
\begin{enumerate}
    \item It contains exactly one point, referred to as the \emph{central point}, whose connectivity factor is $1.0$.
    \item By construction of its neighborhood, all immediate neighbors of the central point, as determined by the adjacency relations in the $m$-dimensional grid, are included in the set.
\end{enumerate}

Formally, 
\begin{equation}
    \boxed{
    S_{min}^m = \{ \mathbf{q} \} \cup \mathcal{N}(\mathbf{q}), \text{where } \mathbf{q} \in \mathbb{R}^m
    }
\end{equation}
The minimality requirement guarantees that no proper subset of this configuration satisfies the same conditions. Therefore, the cardinality of a minimal set $S_{min}^m$ is given by
\begin{equation}
|S_{min}^m| = 3^m.
\end{equation}

\vspace{0.5em}

\textbf{Examples of minimal sets:}
\newcommand{\circled}[1]{\tikz[baseline=(char.base)]{\node[shape=circle,draw,inner sep=1.5pt] (char) {\(\displaystyle #1\)};}}

\begin{minipage}[t]{\textwidth}
\centering
$m = 0$: \\[0.1cm]
\begin{tabular}{c c c}
% Left-hand side TikZ diagram
\begin{tikzpicture}[baseline=(current bounding box.center)]
\def\m{1}
\def\n{1}
\draw[step=1cm, gray, very thin] (0,0) grid (\n,\m);

% Create a node with a circle shape and place the 'X' inside it
\node[draw, circle, inner sep=1pt] at (0.5, 0.5) {\Large \(X\)};

\end{tikzpicture}
&
% Equivalence sign
\(\vcenter{\hbox{\Huge $\equiv$}}\)
&
% Right-hand side matrix
% \(\vcenter{\hbox{\displaystyle
$
\begin{bmatrix}
\circled{1}
\end{bmatrix}
$
% }}\)
\end{tabular}
\end{minipage}

\vspace{0.3cm} % Increased spacing between minipages

\begin{minipage}[t]{\textwidth}
\centering
$m = 1$: \\[0.3cm]
\begin{tabular}{c c c}
% TikZ diagram
\begin{tikzpicture}[baseline=(current bounding box.center), scale=1]
\def\m{1}
\def\n{3}
\draw[step=1cm, gray, very thin] (0,0) grid (\n,\m);
\foreach \j in {1, 3} {
    \node at (\j-0.5, 0.5) {\Large \(X\)};
}
\node[draw, circle, inner sep=1pt] at (1.5, 0.5) {\Large \(X\)};
\end{tikzpicture}
&
% Equivalence sign
\(\vcenter{\hbox{\Huge $\equiv$}}\)
&
% Matrix

$
\frac{1}{3}
\begin{bmatrix}
\frac{1}{2} & \circled{\frac{2}{2}} & \frac{1}{2}
\end{bmatrix}
$

\end{tabular}
\end{minipage}

\vspace{0.3cm} % Increased spacing between minipages

\begin{minipage}[t]{\textwidth}
\centering
$m = 2$: \\[0.1cm]
\begin{tabular}{c c c}
\begin{tikzpicture}[baseline=(current bounding box.center)]
\def\m{3}
\def\n{3}
\draw[step=1cm, gray, very thin] (0,0) grid (\n,\m);
\foreach \y in {1,...,3} {
    \foreach \x in {1,...,3} {
        \node at (\x-0.5, \y-0.5) {\Large \(X\)};
    }
}
\node[draw, circle, inner sep=1pt] at (1.5, 1.5) {\Large \(X\)};
\end{tikzpicture}
&
\(\vcenter{\hbox{\Huge $\equiv$}}\)
&
$
\frac{1}{9}
\begin{bmatrix}
\frac{3}{8} & \frac{5}{8} & \frac{3}{8} \\[1.5ex]
\frac{5}{8} & \circled{\frac{8}{8}} & \frac{5}{8} \\[1.5ex]
\frac{3}{8} & \frac{5}{8} & \frac{3}{8}
\end{bmatrix}
$
\end{tabular}
\end{minipage}
\vspace{1em}

\paragraph{Types of Points in a Minimal Set}

We can define different types of points in a minimal set, allowing us to group points which have the same properties.

The parameter to determine the type of a point in a minimal set $S_{min}^m$, where $m$ is the dimension of the space for the minimal set, is that with respect to the central point $\mathbf{q} \in S_{min}^m$, which has a contribution of 1.0 in the connectivity factor of the whole set, the distance vector $\mathbf{d}=\mathbf{p}-\mathbf{q}, \mathbf{p} \in S_{min}^m$ has how many non-zero components. Thus point $\mathbf{p}$ would be of type $t$ if $\mathbf{d}$ has $t$ non-zero components. The type of point is essentially, the L0 Norm of the difference vector $\mathbf{d}$ divided by the spacing $s$ of the grid in set $S_{min}^m$, i.e. $\left\|\frac{\mathbf{p}-\mathbf{q}}{s}\right\|_0$.

\paragraph{Example in 3D:}
Consider the minimal set \( S_{\min}^3 \) in 3-dimensional space with central point \( q = (0,0,0) \).

\begin{itemize}
    \item Type \( t=0 \): The \emph{central point} \( q = (0,0,0) \).
    \item Type \( t=1 \): Points with exactly one non-zero coordinate, e.g., \( (1,0,0), (0,1,0), (0,0,-1) \). These correspond to \emph{face centers} on the cube.
    \item Type \( t=2 \): Points with exactly two non-zero coordinates, e.g., \( (1,1,0), (1,0,-1), (0,-1,-1) \). These correspond to \emph{edge centers} of the cube.
    \item Type \( t=3 \): Points with exactly three non-zero coordinates, e.g., \( (1,1,1), (1,-1,-1), (-1,1,-1) \). These correspond to the \emph{vertices} of the cube.
\end{itemize}

\paragraph{Diagram:}

\begin{center}
\begin{tikzpicture}[scale=3]
% Draw cube edges
\draw[thick] (0,0,0) -- (1,0,0) -- (1,1,0) -- (0,1,0) -- cycle;
\draw[thick] (0,0,1) -- (1,0,1) -- (1,1,1) -- (0,1,1) -- cycle;
\draw[thick] (0,0,0) -- (0,0,1);
\draw[thick] (1,0,0) -- (1,0,1);
\draw[thick] (1,1,0) -- (1,1,1);
\draw[thick] (0,1,0) -- (0,1,1);

% Central point t=0
\filldraw[color=red!80!black, fill=red!80!black] (0.5,0.5,0.5) circle (0.03) node[below] {(centre)};

% Face centers t=1
\foreach \x/\y/\z in {0.5/0/0.5, 1/0.5/0.5, 0.5/1/0.5, 0/0.5/0.5, 0.5/0.5/1, 0.5/0.5/0}
  \filldraw[color=blue!80!black, fill=blue!80!black] (\x,\y,\z) circle (0.025);

% Edge centers t=2 (slightly bigger)
\foreach \x/\y/\z in {1/1/0.5, 1/0.5/1, 0.5/1/1, 0/1/0.5, 0/0.5/0, 0.5/0/0,
                      1/0/0.5, 0.5/0/1, 0/0/0.5, 1/0.5/0, 0/1/1, 0.5/1/0, 0/0.5/1}
  \filldraw[color=green!70!black, fill=green!70!black] (\x,\y,\z) circle (0.03);

% Vertices t=3 (bigger)
\foreach \x/\y/\z in {0/0/0,1/0/0,1/1/0,0/1/0,0/0/1,1/0/1,1/1/1,0/1/1}
    \filldraw[color=violet!85!black, fill=violet!85!black] (\x,\y,\z) circle (0.025);

% Smaller Legend box
\begin{scope}[xshift=1.2cm, yshift=0.5cm, scale=0.6]
    \draw[fill=white] (0,0) rectangle (2,1.5);
    \filldraw[color=red!80!black, fill=red!80!black] (0.3,1.2) circle (0.05);
    \node[right] at (0.4,1.2) {\(t=0\): centre};
    \filldraw[color=blue!80!black, fill=blue!80!black] (0.3,0.85) circle (0.05);
    \node[right] at (0.4,0.85) {\(t=1\): face centres};
    \filldraw[color=green!70!black, fill=green!70!black] (0.3,0.5) circle (0.05);
    \node[right] at (0.4,0.5) {\(t=2\): edge centres};
    \filldraw[color=violet!85!black, fill=violet!85!black] (0.3,0.15) circle (0.05);
    \node[right] at (0.4,0.15) {\(t=3\): vertices};
\end{scope}

\end{tikzpicture}
\end{center}

\paragraph{Number of Types of Points in $S_{min}^m$}
\mbox{}\\

The set $S_{min}^m$ contains a central point and all its neighbors, by the definition established previously. Therefore, with respect to the central point we will have points of the type 0 (corresponding to the central point itself having $\mathbf{d} = [0, 0, \dots , 0]^T$) to type $m$ (corresponding to the central point itself having $\mathbf{d} = [c_1, c_2, \dots , c_m]^T$, where $c_i \in \{s, -s\}$ and $s$ is the spacing considered in the minimal set). Therefore there are $m + 1$ type of points in set $S_{min}^m$.

\paragraph{Number of Points of Type t in $S_{min}^m$}
\mbox{} \\

$\alpha^m_t$ represents the number of points which are of type $t$ with respect to the central point of the minimal set $S_{min}^m$, in $m$ dimensional space.

A type $t$ point contains $t$ coordinates which are non-zero, therefore we can choose $t$ coordinates out of $m$ by ${}^mC_t$ and each of the $t$ selected coordinates have further 2 choices which are from the set $\{s, -s\}$, where $s$ is the spacing of the minimal set.
\begin{equation}
\boxed{
\therefore \alpha^m_t= 2^t \cdot {}^mC_t
}
\end{equation}

We can verify it's correctness by summing over all types of points which should give us the cardinality of the minimal set which is $3^m$.

\[
\sum_{t = 0}^m\alpha_t^m  = \sum_{t = 0}^m 2^t \cdot {}^mC_t= \sum_{t = 0}^m 2^t \cdot 1^{m - t} \cdot {}^mC_t
\]

\[
\\
\text{By Binomial Theorem, }
\sum_{t = 0}^m\alpha_t^m = (2 + 1)^m = 3^m
\]

\subsubsection{Lower Bound for \texorpdfstring{$\mathcal{CF}$}{CF}}

We represent the lower bound for the \(\mathcal{CF}\) of a structure with intrinsic dimension \textit{m} and ambient dimension \textit{n} to be ${}^L\mathcal{CF}^n_m$. \\
For a minimal set the lower bound on the connectivity factor is given by:
\[
{}^L\mathcal{CF}^m_m = \frac{\sum_{i=1}^{3^m} \lvert \mathcal{N}(s_i) \cap S \rvert}{(3^m-1)(3^m)}
\]

The value $\sum_{i=1}^{3^m} \lvert \mathcal{N}(s_i) \cap S \rvert$ counts all neighbor-interactions for this fully packed minimal neighborhood.

To compute the term $\sum_{i=1}^{3^m} \lvert \mathcal{N}(s_i) \cap S \rvert$, we consider a grid of size $3^m$, where $m$ denotes the ambient dimension, and every point belongs to a minimal set with uniform spacing (assumed to be $1$ for simplicity). Each such point can be naturally represented as a vector:
\[
V^T = [c_1, c_2, c_3, \ldots, c_m]
\]
where the $c_i$ specifies the component along each axis.

We now formalize the interaction counts associated with different types of points defined above through a recursive series:
\begin{itemize}
    \item $a_0$ captures the interactions for the central element, i.e., the point with all $c_i = 0$ for $i \in [1, m]$. This central point possesses the maximal number of possible neighbors, so $a_0 = 3^m - 1$.
    \begin{center} % Center the TikZ picture
    \begin{tikzpicture} %[xshift=0cm, 3d/disable] % Adjust xshift if needed, removed from scope
       % Draw the bounding box
    \draw[gray!80, thick] (-1.5, -1.5) rectangle (1.5, 1.5);

    % Center point, black 'X'
    \node[black, thick] at (0,0) {\Large X};

    % Neighboring points, filled black with circles
    \foreach \pos in {(-1,-1), (-1,0), (-1,1), (0,-1), (0,1), (1,-1), (1,0), (1,1)} {
        \node[draw, circle, fill=black, inner sep=1pt] at \pos {};
        % Add an extra circle around the node
        \draw \pos circle (0.2); % Adjust radius as needed
    }

    % Add labels
    \node at (0, 2) {\textbf{Configuration for $m=2$}};
    \node at (0, -2) {\textbf{8 Neighbors} = $3^2 - 1$};
    \end{tikzpicture}
    \end{center}
    \item $a_1$ counts the interactions for those points lying on the axes, i.e., having $c_1 \in \{-1, 1\}$ and $c_i = 0$ for $i \geq 2$, without loss of generality. Since the "pivot" axis can be chosen in $m$ ways and each can take two signs, there are $2^1 \cdot {}^mC_1$ such points.
    \begin{center}
\begin{tabular}{cc}
% First Figure
\begin{tikzpicture}
    % Draw the bounding box
    \draw[gray!80, thick] (-1.5, -1.5) rectangle (1.5, 1.5);

    % Central point, 'X'
    \node[black, thick] at (-1,0) {\Large X};
    \node[black, thick] at (1,0) {\Large X};
    \node[black, thick] at (0,1) {\Large X};
    \node[black, thick] at (0,-1) {\Large X};;

    % Neighboring points, filled black circles
    \foreach \pos in {(-1,-1), (0,0), (-1,1), (1,-1), (1,1)} {
        \node[draw, circle, fill=black, inner sep=1pt] at \pos {};
    }

    % Labels
    \node at (0, 2) {\textbf{Configuration for $m=2$}};
    \node at (0, -2) {\textbf{Locations of 4 type-a1 points}};
\end{tikzpicture}

\begin{tikzpicture}
    % Draw the bounding box
    \draw[gray!80, thick] (-1.5, -1.5) rectangle (1.5, 1.5);

    % Four central 'X' points
    \node[black, thick] at (-1,0) {\Large X};

    % Neighboring points with circles
    \foreach \pos in {(-1,-1),(1,0),(0,1), (0,0), (-1,1), (1,-1),(0,-1), (1,1)} {
        \node[draw, circle, fill=black, inner sep=1pt] at \pos {};
    }

    \foreach \pos in {(-1,1), (0,0), (-1,-1), (0,1), (0,-1)} {
        \draw[black] \pos circle (0.2); % Adjust radius as needed
    }
    
    % Labels (Rewritten for clarity and better presentation)
    \node at (0, 2) {\textbf{Configuration for $m=2$}};
    \node at (0, -2) {\textbf{Set of 5 Neighbors for Point Type $\text{a}_1$ (Minimal Covering Set)}};
\end{tikzpicture}
\end{tabular}
\end{center}
    \item Extending this logic, $a_t$ counts points where $t$ of the $m$ coordinates are nonzero (each $\pm1$), and the remainder are zero. Thus, there are $\alpha_t^m$ such points for each $t = 0, 1, \ldots, m$ as derived previously.
\end{itemize}

Given these definitions, the total number of neighbors for the central point is $a_0 = 3^m - 1$. To understand the structure recursively, consider what occurs when making a single step away from the center along any one axis: precisely one third of the grid points become inaccessible as neighbors, which yields the relationship
\[
a_1 = a_0 - \frac{1}{3}(a_0 + 1).
\]
This recursive structure extends further; at each level, a similar exclusion applies, producing the general recurrence
\[
a_t = a_{t-1} - \frac{1}{3}(a_{t-1} + 1).
\]

By solving this recurrence (proof given in Appendix \ref{proofofat}), we obtain the closed-form:
\begin{equation}
\boxed{
a_t= 2^t 3^{m - t} - 1.
}
\label{an_formula}
\end{equation}

Accumulating the contributions for all point types yields the aggregate interaction sum, and thus,
\[
{}^L\mathcal{CF}^m_m = \mathcal{CF}(S_{min}^m) = \frac{\sum_{t=0}^{m} a_t\cdot \alpha_t^m}{(3^m - 1)(3^m)} = \frac{\sum_{t=0}^{m} [2^t 3^{m-t} - 1]\cdot 2^t \cdot {}^mC_t}{(3^m - 1)(3^m)}
\]

\[
\implies {}^L\mathcal{CF}^m_m = \frac{\sum_{t=0}^{m} 4^t \cdot 3^{m-t}\cdot {}^mC_t - \sum_{t = 0}^m 1^{m - t} \cdot 2^t \cdot {}^mC_t}{(3^m - 1)(3^m)}
\]

We can simplify this further using the binomial theorem, giving:
\[
{}^L\mathcal{CF}^m_m = \frac{(4 + 3)^m - (2 + 1)^m}{(3^m - 1)(3^m)} = \frac{7^m - 3^m}{(3^m - 1)(3^m)}.
\]

Therefore, for minimal sets in $m$ dimensions, the lower bound for the connectivity factor is:
\begin{equation}
{}^L\mathcal{CF}^m_m =
\begin{cases}
    \dfrac{7^m - 3^m}{(3^m - 1)(3^m)}, & \text{if } m > 0 \\[8pt]
    1, & \text{if } m = 0
\end{cases}, \quad \text{where } m \in \mathbb{W}.
\end{equation}

Using the Space Conversion Formula defined previously, we can compute the connectivity factor of this minimal set when embedded in higher dimensions as well.

\[
\implies {}^L\mathcal{CF}^n_m =
\begin{cases}
    \dfrac{3^m - 1}{3^n - 1} \cdot {}^L\mathcal{CF}_m^m, & \text{if } 0 \leq m \leq n, n > 0 \\[8pt]
    1, & \text{if } n = 0, m = 0
\end{cases}, \quad \text{where } m, n \in \mathbb{W}.
\]

\[
\implies {}^L\mathcal{CF}^n_m =
\begin{cases}
    \dfrac{3^m - 1}{3^n - 1} \cdot \dfrac{7^m - 3^m}{(3^m - 1)(3^m)}, & \text{if } 0 \leq m \leq n, n > 0 \\[8pt]
    1, & \text{if } n = 0, m = 0
\end{cases}, \quad \text{where } m, n \in \mathbb{W}.
\]

\begin{equation}
\therefore 
\boxed{
{}^L\mathcal{CF}^n_m =
\begin{cases}
    \dfrac{7^m - 3^m}{(3^n - 1)(3^m)}, & \text{if } 0 \leq m \leq n, n > 0 \\[8pt]
    1, & \text{if } n = 0, m = 0
\end{cases}, \quad \text{where } m, n \in \mathbb{W}.
}
\end{equation}

\subsubsection{Middle Value for \texorpdfstring{$\mathcal{CF}$}{CF}:}

We define the middle value (middle bound) for $\mathcal{CF}$ of a structure with intrinsic dimension $m$ and ambient dimension $n$ as ${}^M\mathcal{CF}^n_m$.

\paragraph{Full Set in \texorpdfstring{$m$}{m} dimensions}
\mbox{}\\

Consider the scenario where, in an $m$-dimensional space, our set contains all possible points corresponding to a given grid spacing $s$, effectively filling the space entirely. In this setting, as the set size $\lvert S \rvert$ tends to infinity, every point in the $m$-dimensional structure has all of its possible neighbors within the set.

Mathematically, we define it as:

\begin{equation}
\boxed{
S_{full}^m = \{ p \in \mathbb{R}^m \mid p = q + s \cdot z, \text{ for some } z \in \mathbb{Z}^m \}
}
\end{equation}

As a consequence of having all possible points in our infinite set, we get the following property for the full set $S_{full}^m$:
\[
\lvert \mathcal{N}(p) \cap S \rvert = 3^m - 1, \text{where } p \in S_{full}^m
\]
This yields a connectivity factor of
\[
{}^M\mathcal{CF}_m^m = \mathcal{CF}(S_{full}^m) = \frac{\sum_{i=1}^{N} |\mathcal{N}(s_i) \cap S_{full}^m|}{N \cdot (3^m - 1)}, \text{where } N = \lvert S_{full}^m \rvert
\]

Due to the above described property, we have the equation:
\[
{}^M\mathcal{CF}_m^m = \frac{N \cdot (3^m - 1)}{N \cdot (3^m - 1)} = 1
\]

\begin{equation}
\therefore {}^M\mathcal{CF}_m^m = 1.
\end{equation}

To relate the intrinsic $m$-dimensional structure to its representation when embedded in a higher-dimensional space of dimension $n$, we employ the space conversion formula for this configuration:
\[
{}^M\mathcal{CF}_m^n = \frac{3^m - 1}{3^n - 1} \cdot {}^M\mathcal{CF}_m^m = \frac{3^m - 1}{3^n - 1} \cdot 1
\]

\begin{equation}
\boxed{
{}^M\mathcal{CF}_m^n = \frac{3^m - 1}{3^n - 1} 
}
\end{equation}

We note that we have empirically observed this value to be representative of the most probable connectivity factor for a given structure, assuming a lack of noise.

\paragraph{Intuition for \texorpdfstring{${}^M\mathcal{CF}_m^n$}{MCF} being the most probable value of \texorpdfstring{$\mathcal{CF}$}{CF}}
\mbox{}\\

As the density of points of a given structure increases, the local neighborhood configuration for the constituent points converges to a state reflective of the structure's intrinsic dimension, $m$. For a 1-dimensional manifold, an increasing number of interior points will exhibit a local neighborhood count that approaches the theoretical value of $3^1 - 1 = 2$ adjacent neighbors within a discretized grid. Similarly, for a 3-dimensional manifold, the observed neighbor count for a vast majority of points will converge to $3^3 - 1=26$. This convergence is predicated on the principle that in the limit of infinite point density, the topological properties of the manifold's interior dominate the metric effects of its boundaries or junctions. The ratio of boundary points to interior points approaches zero, and thus, the statistical measure of local neighborhood cardinality becomes an increasingly accurate estimator of the manifold's true dimensionality.

\subsubsection{Upper Bound for \texorpdfstring{$\mathcal{CF}$}{CF}:}

The goal of this derivation is to establish a precise mathematical formula for \textbf{${}^U\mathcal{CF}_{m}^{n}$}, which represents the theoretical \textbf{maximum achievable} \emph{Connectivity Factor} for an idealized, non-fractal structure possessing topological dimension $m$ embedded within a higher-dimensional space of ambient dimension $n$.

\paragraph{Labeled Grid Set in \texorpdfstring{$n$}{n} dimensions}
\mbox{}\\
Let $S_{l0}^n$ represent a set of labeled points around a reference point $q$, where $q \in \mathbb{R}^n$ having a spacing $2s$ between them where $s \in \mathbb{R}$, where the label of each point is $0$.

Formally,
\begin{equation}
S_{l0}^n = \{ (q + 2s \cdot z, 0) \mid z \in \mathbb{Z}^n \}
\end{equation}

By $S_{l0}^n$, we represent a set containing label of type $0$ only and it is used to define a rigid structure for our final labeled set that we aim to derive.

Let $S_{l0}^{n'}$ represent the unique set of point label pairs such that for a point $p$ from the set $S_{l0}^n$ we include all its neighbors $r \in \mathcal{N}(p)$ (with spacing $s$), and label them with a type $t$. The type $t$ corresponds to the number of non-zero coordinates in the scaled difference vector between $r$ and $p$.

Formally,
Let $P_0 = \{ p \mid (p, 0) \in S_{l0}^n \}$. The set of labeled neighbors is:
\begin{equation}
S_{l0}^{n'} = \left\{ \left(r, \left\|\frac{r-p}{s}\right\|_0\right) \;\middle|\; p \in P_0, \; r \in \mathcal{N}(p) \right\}
\end{equation}

By $S_{l0}^{n'}$, we represent a set containing all possible labels for dimension $n$, except the label $0$ i.e. a complement set.

Let set $S_l^n$ denote the fully labeled set of points, with a fixed reference $q$, with a spacing between points of $s$.

Formally,
\begin{equation}
\boxed{
S_l^n = S_{l0}^n \cup S_{l0}^{n'}
}
\end{equation}

\newpage

\textbf{Example of $S^2_{l}$:}

\begin{center}
\begin{tikzpicture}
    % Define the grid data as a nested array for pgfmath.
    \def\griddata{{
        {2, 1, 2, 1, 2},
        {1, 0, 1, 0, 1},
        {2, 1, 2, 1, 2},
        {1, 0, 1, 0, 1},
        {2, 1, 2, 1, 2}
    }}

    % Draw grid lines from (0,0) to (5,5).
    % The intersections will now be at integer coordinates.
    \draw[step=1.0, gray, thin] (0,0) grid (5,5);

    % Iterate through rows from 0 to 4
    \foreach \y in {0,...,4} {
        % Iterate through columns from 0 to 4
        \foreach \x in {0,...,4} {
            % Correctly access the array element at row \y, column \x
            \pgfmathsetmacro{\label}{\griddata[\y][\x]}

            % Place the label as a node at the center of the cell.
            % We add 0.5 to both x and y coordinates.
            % The y-coordinate is calculated as (4.5 - \y) to ensure
            % the data appears top-to-bottom.
            \node at (\x + 0.5, 4.5 - \y) {\pgfmathprintnumber{\label}};
        }
    }
\end{tikzpicture}
\end{center}

Note that this grid is infinite and extends in all directions, the example above contains only a $5\times5$ section of $S_l^2$.

In general, we can define $S_{lt}^n \subset S_l^n$ which contains only labels of type $t$.

From this the property follows:

\begin{equation}
\boxed{
S_l^n = \bigcup_{t = 0}^n S^n_{lt}
}
\end{equation}

The ultimate goal of constructing $S_l^n$ is to establish a formal upper bound for the connectivity factor, ${}^U\mathcal{CF}_m^n$, using our $n$-dimensional labeled grid set, $S_l^n$. A critical prerequisite for determining connectivity is to first have a precise understanding of the local environment around any given point.

The central challenge is to determine the number of points of type $y$ are in the immediate neighborhood of a point of type $x$. We denote this quantity as ${}^x\alpha^n_y$. A naive, brute-force enumeration of these neighbors is not feasible. Such an approach would be specific to a single dimension $n$ and a single \emph{perspective} (labeled neighborhood with respect to) $x$, and it would become combinatorially intractable and error-prone as $n$ grows. To solve this problem formally, we require a general and analytical formula for ${}^x\alpha^n_y$ that holds for any $n$, $x$, and $y$.

\paragraph{The System Space Decomposition Strategy}
\mbox{}\\

To derive this general formula, we introduce a decomposition strategy. The core idea is to imagine the $n$-dimensional neighborhood of a point as a composite object built from smaller, simpler pieces. We achieve this by partitioning the $n$ dimensions into two distinct groups:
\begin{enumerate}
    \item A set of $x$ dimensions, which we use to define a high-level "scaffold" or blueprint.
    \item The remaining $n-x$ dimensions, which describe the "internal structure" of the components placed on that scaffold.
\end{enumerate}

This leads us to our two primary constructs:
\begin{itemize}
    \item \textbf{The System Space ($\Omega^n_x$):} This is the $x$-dimensional scaffold. Each location on this scaffold represents and organizes one of our simpler components.
    \item \textbf{The Subsystem ($\omega_k^r$):} These are the $(n-x)$-dimensional components that are placed \textit{onto} the scaffold. They represent a standardized, lower-dimensional piece of the total neighborhood.
\end{itemize}

The key insight of this framework is that we can systematically count the points of type $y$ in the total $n$-dimensional neighborhood by first choosing a location on the $x$-dimensional scaffold and then counting the relevant points within the $(n-x)$-dimensional subsystem that resides there. By summing over all possible locations on the scaffold, we can build a complete and accurate count for the entire neighborhood. This deconstructive method allows us to transform a complex, high-dimensional counting problem into a simple summation. We will now proceed with the formal definitions of these constructs.

\subparagraph{Subsystems:}

A subsystem $\omega_k^r$ is defined as a construct of a mapped minimal set of dimension $r$ such that type $t$ points are mapped to type $t + k$ points. Here $t + k$ is not limited by the value of $r$ and is simply a mapping of type of points.

\textbf{Example 1: Subsystem $\omega_1^2$}

This subsystem is a construct based on a 2-dimensional minimal set ($r=2$) with a mapping where the type is shifted by $k=1$.

\begin{itemize}
    \item \textbf{Underlying Minimal Set}: A 2-dimensional minimal set, which is a 3x3 grid of points. It has a total of $3^2=9$ points.

    \item \textbf{Point Types Present}:
    \begin{itemize}
        \item \textbf{Type 0}: The single central point.
        \item \textbf{Type 1}: The four points with one non-zero coordinate relative to the center.
        \item \textbf{Type 2}: The four points with two non-zero coordinates (the corners).
    \end{itemize}

    \item \textbf{Mapping ($k=1$)}: The rule is to map points of type $t$ to the concept of type $t+1$.
    \begin{itemize}
        \item \textbf{Type 0} points are mapped to \textbf{Type 1}.
        \item \textbf{Type 1} points are mapped to \textbf{Type 2}.
        \item \textbf{Type 2} points are mapped to \textbf{Type 3}. Although no points of Type 3 exist in this 2D minimal set, the mapping itself is a valid mental construct as per the definition.
    \end{itemize}
\end{itemize}

\textbf{Example 2: Subsystem $\omega_3^2$}

This subsystem is also based on a 2-dimensional minimal set ($r=2$), but with a larger mapping shift of $k=3$.

\begin{itemize}
    \item \textbf{Underlying Minimal Set}: A 2-dimensional minimal set (a 3x3 grid) with 9 points.

    \item \textbf{Point Types Present}:
    \begin{itemize}
        \item \textbf{Type 0}: The central point.
        \item \textbf{Type 1}: The four points adjacent to the center.
        \item \textbf{Type 2}: The four corner points.
    \end{itemize}

    \item \textbf{Mapping ($k=3$)}: The rule is to map points of type $t$ to the concept of type $t+3$.
    \begin{itemize}
        \item \textbf{Type 0} points are mapped to \textbf{Type 3}.
        \item \textbf{Type 1} points are mapped to \textbf{Type 4}.
        \item \textbf{Type 2} points are mapped to \textbf{Type 5}.
    \end{itemize}
    In this case, all mappings from the existing point types are to conceptual types that are not physically present in the underlying 2-dimensional set.
\end{itemize}

\textbf{System Space:}

A System Space $\Omega^n_x$ is defined as a construct of a minimal set of dimension $x$ such that each element of the set is mapped to a \textbf{subsystem} $\omega^{n - x}_{x - t}$, i.e each vector in this minimal set of type $t$ where type is consistent with the previously defined types, is mapped to a subsystem aforementioned.

Formally, for a minimal set $S_{\text{min}}^x$ with $q \in \mathbb{R}^x$ as the center and $s \in \mathbb{R}$ as the spacing, the System Space is the set of these specified subsystems:
\begin{equation}
\boxed{
\Omega^n_x = \left\{ \omega^{n - x}_{x - t} \;\middle|\; p \in S_{\text{min}}^x, \; t = \left\|\frac{p-q}{s}\right\|_0 \right\}
}
\end{equation}

\textbf{Example: System Space $\Omega_2^3$}

This example demonstrates the decomposition of a 3-dimensional problem ($n=3$) into a 2-dimensional arrangement ($x=2$) of subsystems.

\begin{itemize}
    \item \textbf{Decomposition Parameters}:
    \begin{itemize}
        \item Total problem dimension, $n=3$.
        \item System Space dimension, $x=2$.
    \end{itemize}

    \item \textbf{Underlying Minimal Set}: Per the definition, $\Omega_2^3$ is constructed upon an underlying minimal set of dimension $x=2$, which is $S_{\text{min}}^2$. This set consists of $3^2=9$ points, whose types ($t$) are arranged as follows:
    
    \begin{center}
    \begin{tabular}{|c|c|c|}
        \hline
        $t=2$ & $t=1$ & $t=2$ \\ \hline
        $t=1$ & $t=0$ & $t=1$ \\ \hline
        $t=2$ & $t=1$ & $t=2$ \\ \hline
    \end{tabular}
    \end{center}

    \item \textbf{Mapping to Subsystems}: Each point $p$ in the $S_{\text{min}}^2$ grid is mapped to a subsystem $\omega_k^r$. The subsystem parameters $r$ and $k$ are derived from the System Space rules:
    \begin{itemize}
        \item The dimension of each resulting subsystem is $r = n - x = 3 - 2 = 1$. Thus, every subsystem will be 1-dimensional ($\omega^1_k$).
        \item The type-shift mapping for each subsystem is $k = x - t = 2 - t$. The specific shift $k$ is therefore dependent on the type $t$ of the corresponding point in the $S_{\text{min}}^2$ grid.
    \end{itemize}

    \item \textbf{Calculating the Specific Subsystems}:
    \begin{itemize}
        \item \textbf{Central point (t=0)}: This position maps to a subsystem with $k = 2 - 0 = 2$. The subsystem is $\omega_2^1$.
        \item \textbf{Four adjacent points (t=1)}: Each of these four positions maps to a subsystem with $k = 2 - 1 = 1$. The subsystem is $\omega_1^1$.
        \item \textbf{Four corner points (t=2)}: Each of these four positions maps to a subsystem with $k = 2 - 2 = 0$. The subsystem is $\omega_0^1$.
    \end{itemize}

    \item \textbf{Resulting System Space}: The System Space $\Omega_2^3$ is the structured collection of these nine 1-dimensional subsystems. The spatial arrangement of these subsystems mirrors the $S_{\text{min}}^2$ grid that generated them:

    \begin{center}
    \begin{tabular}{|c|c|c|}
        \hline
        $\omega_0^1$ & $\omega_1^1$ & $\omega_0^1$ \\ \hline
        $\omega_1^1$ & $\omega_2^1$ & $\omega_1^1$ \\ \hline
        $\omega_0^1$ & $\omega_1^1$ & $\omega_0^1$ \\ \hline
    \end{tabular}
    \end{center}
    
    Formally, the set of subsystems is $\Omega_2^3 = \{\omega_2^1, \omega_1^1, \omega_1^1, \omega_1^1, \omega_1^1, \omega_0^1, \omega_0^1, \omega_0^1, \omega_0^1 \}$. This exemplifies the decomposition of a 3D problem into a structured set of 1D constructs.
\end{itemize}

\subparagraph{Number of Subsystems of type $t$}
\mbox{}\\

A subsystem $\omega^{n - x}_{x - t}$ corresponds to type $t$, in an $x$ dimensional scaffold, therefore there are $\alpha^x_t=2^t \cdot {}^xC_t$ subsystems $\omega^{n - x}_{x - t}$ in our system space.

\subparagraph{Number of Points of type $y$ in $\omega^{n - x}_{x - t}$}
\mbox{}\\

A subsystem $\omega^{n - x}_{x - t}$ contains points of type $x - t$ and above therefore if $y < x - t$ then we have 0 types of these points in our subsystem. Another case is we search for a point of type $y$ such that $y > n - t$, since we can have max internal type of the subsystem as $n - x + x - t = n - t$, we will have 0 such points.

For a subsystem $\omega^{n - x}_{x - t}$, when $x - t \le y \le n - t$, we will have $\alpha^{n - x}_{y - (x - t)} = 2^{y - (x - t)} \cdot {}^{n - x}C_{y - (x - t)}$ points of type $y$.

\paragraph{Perspective and the Formulation of ${}^x\alpha^n_y$}
\mbox{}\\

To determine the number of points of type $y$ in the immediate neighborhood of a point of type $x$, a quantity we denote ${}^x\alpha^n_y$, we employ the System Space Decomposition Strategy. We establish our framework by setting the System Space dimension equal to the perspective type, $x$. This choice effectively centers our analysis on a type $x$ point and structures its $n$-dimensional neighborhood as an $x$-dimensional System Space, $\Omega^n_x$.

The total count of type $y$ neighbors is obtained by summing the contributions from all subsystems within this System Space. The summation is performed over all scaffold-point types $i$ (from $0$ to $x$):
\[
\text{Gross Count} = \sum_{t=0}^x \alpha^x_t \cdot \alpha^{n - x}_{y - (x - t)}
\]

This summation constructs the entire local environment from the perspective of the type $x$ point. However, in doing so, it inherently includes the perspective point itself in the enumeration. This occurs in the specific case where the type of the neighbor being counted is the same as the type of the perspective point ($y=x$). The $i=0$ term of the sum corresponds to the center of the scaffold, and within the subsystem at that location, the $\alpha^{n-x}_0$ term counts the subsystem's center. This combination represents the perspective point itself.

Since a point cannot be a member of its own neighborhood, this self-count must be subtracted. We introduce a correction term that is equal to 1 if and only if $y=x$. The Kronecker delta $\delta_{xy}$, is ideal for this purpose.

This leads to the final, precise formula for the number of neighbors of type $y$ in perspective $x$, in a $n$ dimensional space:
\[
    {}^{x}\alpha^{n}_y = \left[ \sum_{t=0}^x \alpha^x_t \cdot \alpha^{n - x}_{y - (x - t)} \right] - \delta_{xy}
\]
Substituting the combinatorial expressions for $\alpha$ yields the complete analytical solution:
\begin{equation}
\boxed{
    {}^{x}\alpha^{n}_y = \left[ \sum_{t=0}^x \left( 2^{t} \cdot {}^{x}C_t \right) \left( 2^{y - (x - t)} \cdot {}^{n - x}C_{y - (x - t)} \right) \right] - \delta_{xy}
    }
\end{equation}

This formulation is a direct consequence of our decomposition strategy, as it systematically sums the contributions of type $y$ points from every subsystem arranged on the $x$-dimensional scaffold.

It can be trivially proved that for $x = 0$, we have a single subsystem in our system space which is $\Omega^n_0 = \{ \omega^n_0 \}$. Therefore, we have a single term of $t = 0$ in our summation which is $\alpha_0^0 \cdot \alpha^n_y = \alpha^n_y$.

Therefore we prove that ${}^0\alpha_y^n = \alpha^n_y$, which can be verified via our initial definition of alpha for our minimal set $S_{min}^m$ centered around the type $0$ point.

\paragraph{Constructing the Maximal Set for topological dimension $m$ in $n$ dimensions}
\mbox{}\\

The set $S_l^n$ represents a labeled set of all points with a spacing $s$ in $n$ dimensions as established before.

This means that $S_l^n$ contains every point available in the $n$ dimensional space. Thus, to construct a maximal set of intrinsic dimension $n$ in a $n$ dimensional space, we keep the entire set $S_l^n$ and take all points in it.

\begin{equation}
S_{max}^{(n, n)} = S_l^n
\end{equation}

Now, we want to make a maximal set such that $n - 1$ is the intrinsic dimension with $n$ as the dimension of space. Thus, we must remove certain points from the labeled set $S_l^n$.

When we remove all the points which are centers of the $n$ dimensional hypercubes in our labeled set, we are left with no single point which gives an identity of $n$ dimension since not a single point contributes $3^n - 1$ anymore. The perfect candidate for this removal is our type $0$ point in our labeled grid.

\begin{equation}
\implies S_{max}^{(n, n - 1)} = S_l^n \setminus S^n_{l0}
\end{equation}

Similarly to reduce the dimension by one more, we must reduce the points in the set once more. This time the center of the $n - 1$ boundaries thus formed in the previous set is eliminated, for which type $1$ labeled points are the perfect candidate, reducing our set to the topological dimension $n - 2$

\begin{equation}
\implies S_{max}^{(n, n - 2)} = S_{max}^{(n, n - 1)} \setminus S^n_{l1} = S_l^n \setminus (S^n_{l0} \cup S_{l1}^n) = \bigcup_{t = 2}^{n} S_{lt}^n
\end{equation}

We see that, the type of point that we must remove from the previous reduced set is the point which is the center of the current reduced set structure, which is the point with type $m$ when we want to go from maximal set of topological dimension $n - m$ to topological dimension $n - m - 1$ with an ambient dimension $n$.

Therefore, we arrive at the following formulation for a topological dimension $m$ with ambient space $n$ the maximal set is represented as,

\begin{equation}
S_{max}^{(n, m)} = S_{max}^{(n, m + 1)} \setminus S^n_{l(n - m - 1)} = S_l^n \setminus \bigcup_{t = 0}^{n - m - 1}S^n_{lt}, \text{where } 0 \le m < n, S_{max}^{(n, n)} = S_l^n
\end{equation}

We can use a constructive form, instead of destructive form of the expression to incorporate the base case into the formula as well, 
\begin{equation}
\boxed{
\therefore S_{max}^{(n, m)} = \bigcup_{t = n - m}^{n} S_{lt}^n
}
\end{equation}

\subparagraph{Connecting Maximal Sets to ${}^x\alpha_y^n$}
\mbox{}\\

We have established that our Maximal Set, will constructively consist of points which are labeled from $n - m \to n$. This means that, for each type of point $t$ which still exists in the set i.e. $n - m \le t \le n$, the number of neighbors in the maximal set of $m$ topological dimension with ambient space as $n$, would be $\sum_{i = n - m}^n {}^t\alpha^n_i$.

\paragraph{Contribution Formula (${}^m\chi_t^n$)}
\mbox{}\\

The contribution of a single point of type $t$ to the connectivity factor in the maximal set $S_{max}^{(n, m)}$:
\begin{equation}
\boxed{
{}^m\chi_t^n = \frac{\sum_{i = n-m}^n {}^t\alpha_i^n}{3^n - 1}, \quad \text{where } n-m \leq t \leq n.
}
\end{equation}

\paragraph{Frequency Formula ($f_t$)}
\mbox{}\\

We have ${}^m\chi^n_t$ i.e. contribution of a point of type $t$ in the maximal set of intrinsic dimension $m$ in ambient space of dimension $n$ however every maximal set is an infinite set therefore, it is crucial to get the proportion of these points with respect to each other in the maximal set.

\subparagraph{Calculation of equivalents ($e_t$)}: 
\mbox{}\\

\[
e_t = \frac{\text{Number of type } t \text{ points under fixed perspective } (x=0)}{\text{Number of hypercubes the point } t \text{ contributes to } (x = t)},
\]
which simplifies to
\begin{equation}
e_t = \begin{cases}
\frac{{}^0\alpha_t^n}{{}^t\alpha_0^n} & 0 < t \leq n \\
1 & t=0 \\
\text{Invalid} & \text{otherwise}
\end{cases} = {}^nC_t.
\end{equation}

The frequency of points of type $t$ is then 
\begin{equation}
f_t = \frac{{}^nC_t}{\displaystyle \sum_{i = n - m}^n {}^nC_i}
\end{equation}

\paragraph{Final Upper Bound Formula}
\mbox{}\\

The overall upper bound on the connectivity factor is the weighted average of contributions, weighted by frequencies, of all types of points in the $S_{max}^{(n, m)}$ set:
\begin{equation}
\boxed{
{}^U\mathcal{CF}_m^n = \mathcal{CF}(S_{max}^{(n, m)}) = \sum_{t = n - m}^n f_t \cdot {}^m\chi_t^n
}.
\end{equation}

Explicitly,
\begin{equation}
\boxed{
{}^U\mathcal{CF}_m^n = \sum_{t = n - m}^n \left( \frac{{}^nC_t}{\displaystyle \sum_{i = n - m}^n {}^nC_i} \cdot \frac{\sum_{i = n-m}^n (\left[ \sum_{j=0}^t \left( 2^{j} \cdot {}^{t}C_j \right) \left( 2^{i - (t - j)} \cdot {}^{n - t}C_{i - (t - j)} \right) \right] - \delta_{ti}
)}{3^n - 1}
\right)
}.
\end{equation}

\vspace{1em}

\textbf{Note:} While the upper and middle bounds both achieve a maximum value of 1.0 (when a structure is embedded in its intrinsic dimension), the middle bound is still less than the upper bound when a structure is not embedded in its intrinsic dimension.
\\
Appendix \ref{lmudiscussion} contains more information about the LMU (Lower, Middle, Upper) bounds.
\\
We now propose two methods for theoretical estimation of intrinsic dimension, namely the $\mathcal{LMU-CF}$ (LMU Bound based $\mathcal{CF}$) and the $\mathcal{DCF}$ (Distributed $\mathcal{CF}$).

\vspace{1em}

\subsection{Overlap of LMU Bounds}
\label{subsubsec:lmu_flag_overlaps_influence}

We note that the lower, middle and upper bounds for $\mathcal{CF}$ have overlaps and do not behave as strict bounds for intrinsic dimension classification (an example is provided below of the same). Thus, there is a need to develop a weighing function in order to correctly determine the influence each bound has over the estimation of the intrinsic dimension. The weighing function should have the following characteristics:

\begin{itemize}
    \item It accepts as input a scalar $\mathcal{CF}$ value, and optionally the number of points in the dataset.
    \item The output is a vector of weights over all candidate intrinsic dimensions, typically of size \( n + 1 \), reflecting the or degree of membership of the structure in each possible dimension.
    \item Common implementations of such influence functions include triangular cap functions, Gaussian distributions, etc. or learned weights via machine learning techniques.
    \item Aggregating the weight vector over all points allows computation of a final intrinsic dimension estimate for the dataset.
\end{itemize}

We note that in the following sections we use a weighting function as specified above; however, this weighting function is not used for the \(\mathcal{LMU-CF}\) method. Unlike the \(\mathcal{DCF}\) method introduced next, the \(\mathcal{LMU-CF}\) method is entirely theoretical and does not lend itself well to empirical estimation of bounds. This implies that without a sufficient number of points (approximately of the order of \(3^{d}\), where \(d\) is the ambient dimension of the dataset), we cannot use \(\mathcal{LMU-CF}\). In contrast, \(\mathcal{DCF}\) can be modified to \(e\mathcal{DCF}\) to work effectively with a lower number of points.
\vspace{1em}

\textbf{Illustrative Examples:}

\begin{figure}[H]
\centering
\begin{tikzpicture}[
    font=\sffamily\small,
    % Define colors
    myblue/.style={color=blue},
    myred/.style={color=red},
    myorange/.style={color=orange},
    mygreen/.style={color=green!60!black},
    mypurple/.style={color=purple}
]
% --- n=0 Line ---
\begin{scope}[yshift=6cm]
    \node[anchor=west, font=\bfseries] at (-1.5, 0) {$n=0$};
    \draw[-{Stealth}] (0, 0) -- (12, 0);
    \draw (0, 0.1) -- (0, -0.1) node[below] {0};
    \draw (10, 0.1) -- (10, -0.1) node[below] {1};
    % L0, M0, U0 at the end
    \draw[myblue, thick] (10, 0.2) -- (10, 0.2);
    \node[myblue, above, font = \tiny] at (10, 0.2) {$L_0$, $M_0$, $U_0$};
\end{scope}
% --- n=1 Line ---
\begin{scope}[yshift=4cm]
    \node[anchor=west, font=\bfseries] at (-1.5, 0) {$n=1$};
    \draw[-{Stealth}] (0, 0) -- (12, 0);
    \draw (0, 0.1) -- (0, -0.1) node[below] {0};
    \draw (10, 0.1) -- (10, -0.1) node[below] {1};
    % L0, M0, U0 at the beginning
    \draw[myblue, thick] (0, 0) -- (0, 0);
    \node[myblue, above, font = \tiny] at (0, 0) {$L_0$, $M_0$, $U_0$};
    
    % Line from L1 to end
    \draw[myred, thick] (6.67, 0.2) -- (10, 0.2);
    \node[myred, above, font=\tiny] at (6.67, 0.2) {$L_1$};
    \node[myred, above, font=\tiny] at (10, 0.2) {$M_1$, $U_1$};
    \node[myred, below=1pt, font = \tiny] at (6.67, 0) {\tiny 0.667};
\end{scope}
% --- n=2 Line ---
\begin{scope}[yshift=1.5cm]
    \node[anchor=west, font=\bfseries] at (-1.5, 0) {$n=2$};
    \draw[-{Stealth}] (0, 0) -- (12, 0);
    \draw (0, 0.1) -- (0, -0.1) node[below] {0};
    \draw (10, 0.1) -- (10, -0.1) node[below] {1};
    % L0, M0, U0 at the beginning
    \draw[myblue, thick] (0, 0) -- (0, 0);
    \node[myblue, above, font = \tiny] at (0, 0) {$L_0$, $M_0$, $U_0$};
    % Line from L1 to U1
    \draw[myred, thick] (1.67, 0.2) -- (6.7, 0.2);
    \node[myred, above, font=\tiny] at (1.67, 0.2) {$L_1$};
    \node[myred, above, font=\tiny] at (2.5, 0.2) {$M_1$};
    \node[myred, above, font=\tiny] at (6.7, 0.4) {$U_1$};
    \node[myred, below=1pt] at (1.67, 0) {\tiny 0.1667};
    \node[myred, below=1pt] at (2.5, 0) {\tiny 0.25};
    \node[myred, below=1pt] at (6.7, 0) {\tiny 0.67};
    % Line from L2 to U2 (end)
    \draw[myorange, thick] (5.5, 0.4) -- (10, 0.4);
    \node[myorange, above, font=\tiny] at (5.5, 0.4) {$L_2$};
    \node[myorange, above, font=\tiny] at (10, 0.4) {$M_2$, $U_2$};
    \node[myorange, below=1pt] at (5.5, 0) {\tiny 0.55};
    % Overlap annotation
    \draw[mygreen, decorate, decoration={calligraphic brace, amplitude=3pt}] (5.5, 0.7) -- (6.7, 0.7);
    \node[mygreen, above=2pt, font = \tiny] at (6.1, 0.6) {overlap};
\end{scope}
% --- n=3 Line ---
\begin{scope}[yshift=-1.2cm]
    \node[anchor=west, font=\bfseries] at (-1.5, 0) {$n=3$};
    \draw[-{Stealth}] (0, 0) -- (12, 0);
    \draw (0, 0.1) -- (0, -0.1) node[below] {0};
    \draw (10, 0.1) -- (10, -0.1) node[below] {1};
    % L0 M0 U0 at the beginning
    \draw[myblue, thick] (0, 0) -- (0, 0);
    \node[myblue, above, font = \tiny] at (-0.3, 0) {$L_0$, $M_0$, $U_0$};
    % Range 1: Line from L1 to U1
    \draw[myred, thick] (0.5, 0.2) -- (3.4, 0.2);
    \node[myred, above, font=\tiny] at (0.5, 0.2) {$L_1$};
    \node[myred, above, font=\tiny] at (0.5, 0.4) {0.05};
    \node[myred, above, font=\tiny] at (3.4, 0.4) {$U_1$};
    \node[myred, below=1pt] at (0.7, 0) {\tiny $M_1$};
    \node[myred, below=1pt] at (0.7, -0.2) {\tiny 0.07};
    \node[myred, below=1pt] at (3.4, 0) {\tiny 0.34};
    
    % Range 2: Line from L2 to U2
    \draw[myorange, thick] (1.7, 0.4) -- (8.57, 0.4);
    \node[myorange, above, font=\tiny] at (1.7, 0.4) {$L_2$};
    \node[myorange, above, font=\tiny] at (3.0, 0.4) {$M_2$};
    \node[myorange, above, font=\tiny] at (8.57, 0.55) {$U_2$};
    \node[myorange, below=1pt] at (1.7, 0) {\tiny 0.17};
    \node[myorange, below=1pt] at (3.0, 0) {\tiny 0.3};
    \node[myorange, below=1pt] at (8.57, 0) {\tiny 0.857};
    % Range 3: Line from L3 to U3
    \draw[mypurple, thick] (4.5, 0.6) -- (10, 0.6);
    \node[mypurple, above, font=\tiny] at (4.5, 0.55) {$L_3$};
    \node[mypurple, above, font=\tiny] at (10, 0.6) {$M_3$, $U_3$};
    \node[mypurple, below=1pt] at (4.5, 0) {\tiny 0.45};
    % Overlap 1: 0.17 to 0.34
    \draw[mygreen, decorate, decoration={calligraphic brace, amplitude=3pt}] (1.7, 0.7) -- (3.4, 0.7);
    \node[mygreen, above=2pt, font=\tiny] at (2.55, 0.7) {overlap};
    % Overlap 2: 0.45 to 0.857
    \draw[mygreen, decorate, decoration={calligraphic brace, amplitude=3pt}] (4.5, 0.9) -- (8.57, 0.9);
    \node[mygreen, above=2pt, font=\tiny] at (6.53, 0.9) {overlap};
\end{scope}
\end{tikzpicture}
\caption{Ranges and potential overlaps of the Lower (L), Middle (M), and Upper (U) \emph{Connectivity Factor} ($\mathcal{CF}$) values for various topological dimensions within an \( n \)-dimensional ambient space (wher $K_i$ denotes the $K$ bound (L(lower), M(middle), U(upper)) for embedded dimension \( i \)), for $n$ = 0, 1, 2 and 3.}
\label{fig:cf-scale-overlaps-updated}
\end{figure}
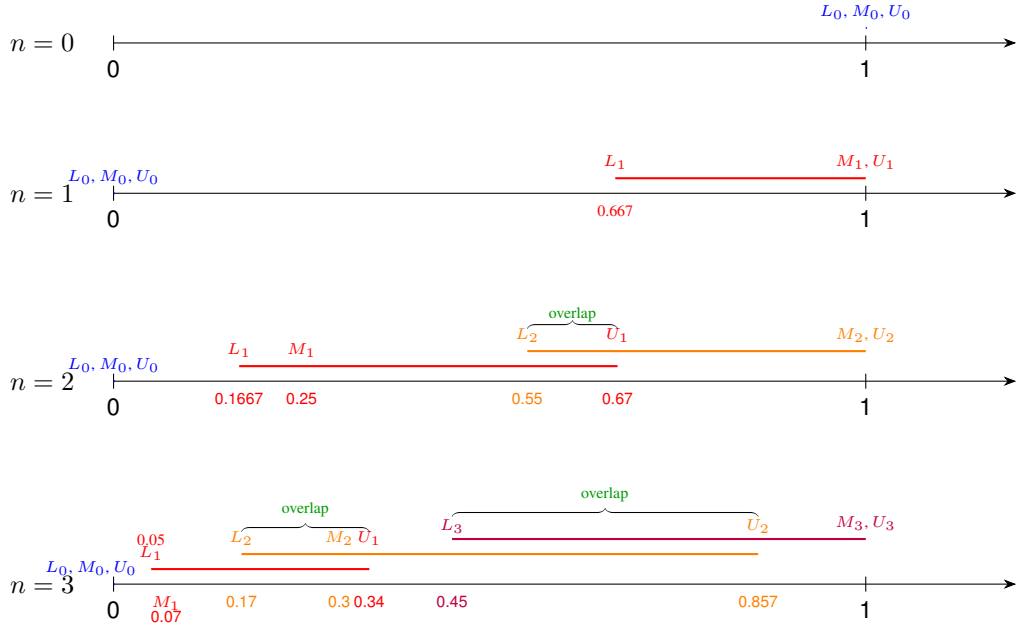

\vspace{1em}

\subsection{Distributed Connectivity Factor (\texorpdfstring{$\mathcal{DCF}$}{DCF})}
\label{dcf_defn}

The Distributed Connectivity Factor ($\mathcal{DCF}$) method acknowledges that individual points may not be purely of a single topological dimension but instead contribute fractionally to multiple dimensions. This framework models each point as having a fractional membership across different topologies.

Fractional contributions are computed using a cap function, which bounds the influence of a point toward each topology based on its local connectivity. For simplicity and interpretability, a linear cap function is employed here, though other weighing functions can easily be swapped in.

\vspace{1em}

\section*{Example: \( n=2 \)}
\begin{tikzpicture}[font=\small]
    % Coordinates for the points
    \coordinate (p1) at (0,0);
    \coordinate (p2) at (2,0);
    \coordinate (p3) at (8,0);
    \coordinate (topleft) at (0, 2);
    \coordinate (p2_hump) at (2, 2);
    \coordinate (p3_hump) at (8, 2);

    % Main line segment and dashed line
    \draw[purple, thick] (p1) -- (p3);
    \draw[purple, dashed] (-0.5, 2) -- (8.5, 2);

    % Fractional connection illustrations
    \draw[red, thick] (topleft) -- (p1);
    \draw[red, thick] (topleft) -- (p2);
    \draw[red, thick] (p2) -- (p3);

    \draw[orange, thick] (p1) -- (p2_hump);
    \draw[orange, thick] (p2_hump) -- (p3);

    \draw[green, thick] (p1) -- (p2);
    \draw[green, thick] (p2) -- (p3_hump);
    \draw[green, thick] (p3_hump) -- (p3);

    % Points
    \fill[black] (p1) circle (2pt);
    \fill[black] (p2) circle (2pt);
    \fill[black] (p3) circle (2pt);

    % Label '1'
    \node at (topleft) {1};

    % Labels below points
    \node[below=2pt, red] at (p1) {$0$};
    \node[below=14pt, red] at (p1) {$(3^0 - 1)$};

    \node[below=2pt, orange] at (p2) {$2$};
    \node[below=14pt, orange] at (p2) {$(3^1 - 1)$};

    \node[below=2pt, green] at (p3) {$8$};
    \node[below=14pt, green] at (p3) {$(3^2 - 1)$};

    % Decorative brace and label
    \draw [decorate, decoration={brace, amplitude=10pt, mirror}, purple] 
        (p1 |- {0,-1.5}) -- (p3 |- {0,-1.5}) 
        node [midway, below=8pt] {Number of interactions of a point = $3^m - 1$};

    % Arrow and explanatory text
    \node[right] at (8.7, 1.2) {$\rightarrow$};
    \node[text width=5cm, align=left, right, anchor=west] at (9.5, 1.2)
        {\small Based on this function, we determine each point's fractional contribution to each topology.};

    % Vertical dashed line at x=3 and label below
    \draw[dashed, thick] (3,0) -- (3,2);
    \node[below=6pt] at (3,0) {$x = 3$};

\end{tikzpicture}

\vspace{1em}

For instance, in a two-dimensional ambient space (\( n=2 \)), a point with \( x = 3 \) neighbours results in fractional memberships:
\[
1 (orange \ line) \to 0.833 \quad \text{(83.3\% contribution to 1-D topology)},
\]
\[
2 (green \ line) \to 0.167 \quad \text{(16.7\% contribution to 2-D topology)}.
\]

Thus, the fractionally assigned memberships reflect the mixed topological nature of the point.

Aggregating this fractional influence \( f(x,t) \) over all points and normalizing enables a probabilistic estimation of the dataset's topological structure.

\vspace{1em}

\subsubsection{Weighted Membership Assignment}

We determine the weight of dimension $m$ such that if a point $\mathbf{p}$ has $\lvert \mathcal{N}(\mathbf{p}) \cap S \rvert = \lvert\mathcal{N}(\mathbf{p}) \rvert = 3^m - 1, \text{where } \mathbf{p} \in S, S \subset  \mathbb{R}^m$, it has the highest probability of being $m$ dimensional, and this weight (for being $m$ dimensional) linearly decays as we get further away from having number of neighbors equal to $3^m - 1$.

\begin{equation}
r_t = 3^t - 1, \quad t=0,1,\ldots,n, \quad
r_{-1} \coloneqq r_0 = 0, \quad r_{n+1} \coloneqq r_{n} = 3^{n-1}.
\end{equation}
\begin{equation}
f_t(x) =
\begin{cases}
0, & x \notin [\,r_{t-1},\, r_{t+1}\,], \\[6pt]
\min \left\{1, \ \dfrac{x - r_{t-1}}{r_t - r_{t-1}} \right\},
& r_{t-1} \le x \le r_t, \\[10pt]
\min \left\{1, \ \dfrac{r_{t+1} - x}{r_{t+1} - r_t} \right\},
& r_t \le x \le r_{t+1}, \\[10pt]
\mathbf{1}\{x = r_t\}, & r_{t-1} = r_t = r_{t+1}.
\end{cases}
\end{equation}

$D_{max}$ can be set to the highest intrinsic dimension the data could be, which is capped by the ambient dimension.

\subsubsection{Intrinsic Dimension Estimation}
\label{subsubsec:fraction_vector_function_f}

The dataset's total weight vector \( W \) is defined as the sum of the fractional membership vectors \( F(\lvert \mathcal{N}(\mathbf{p}) \cap S\rvert) \) over all points \( \mathbf{p} \in S \). Formally,
\begin{equation}
W = \sum_{\mathbf{p} \in S} F(\lvert \mathcal{N}(\mathbf{p}) \cap S \rvert) = \begin{bmatrix} W_0 \\ W_1 \\ \vdots \\ W_n \end{bmatrix},
\end{equation}
where
\begin{itemize}
    \item \( S \) is the set of sampled points,
    \item \( F(x) = [f_0(x), f_1(x), \ldots, f_n(x)]^\top \) is the fractional membership vector for point $\mathbf{p}$ with $\lvert \mathcal{N}(\mathbf{p}) \cap S \rvert = x$,
    \item \( W_t \) is the total weight (sum of influences) for topology dimension \( t \) over the entire dataset.
\end{itemize}

The estimated intrinsic dimension \( \hat{m} \) is then computed as the weighted average of all possible dimensions, rounded to the nearest integer:
\begin{equation}
\hat{m} = argmax_t (W_t)
\end{equation}

\subsubsection{Algorithm for \texorpdfstring{$\mathcal{DCF}$}{DCF}}

Below is the algorithm for $\mathcal{DCF}$:

\begin{algorithm}[H]
\caption{$\mathcal{DCF}$}
\begin{algorithmic}[1]
\Require Dataset $X=\{x_i\in\mathbb{R}^d\}_{i=1}^n$, Information Percentage Range $IP_{range}=[IP_{low}, IP_{high}]$
\Ensure Estimated intrinsic dimension $\widehat{m}\in\mathbb{W}$

\Function{dcf}{$X,\,IP_{range}$}
  \State \textbf{Normalize to unit box:} Affinely rescale each axis so $X\subset[0,1]^d$.
  \State \textbf{Search over spacing $s$:} Find $s^\star$ such that gridding at step $s^\star$ retains an $IP\%$ of points, $IP \in IP_{range}$.
  \State \textbf{Grid and extract representatives:} Map each $x\in X$ to its grid representative at step $s^\star$ and keep the unique set $\mathcal{C}\subset\mathbb{R}^d$.

  \State \textbf{Calculate neighbor counts:} For each point $u \in \mathcal{C}$, compute its neighbor count $c(u)$ as:
    \[
        c(u) \gets \big|\{v\in\mathcal{C}\setminus\{u\}:\ \|u-v\|_\infty\le s^\star\}\big|.
    \]
    \Comment{Calculated using efficient grid-based neighbour counting}

    \State \textbf{Calculate weights:} $\mu_k \gets 3^k - 1$, where $k = 0, 1,\dots, n$.
    
  \State \textbf{Dimensional scoring (summation of influences):} For $m=0,\dots,n$ set
    \[
        W[m]\ \gets\ \sum_{u \in \mathcal{C}} \kappa_{\triangle}\!\big(c(u);\,\mu_{m-1},\,\mu_m,\,\mu_{m+1}\big),
    \]
    where $c(u)$ is the neighbor count for each point $u$.
    
  \Statex
  \Statex \textbf{Notation for $\kappa_{\triangle}$}: The function $\kappa_{\triangle}(x; l, c, r)$ is the triangular (hat) basis function defined by a left point $l$, a center (peak) $c$, and a right point $r$. It has a value of $1$ at $x=c$ and decreases linearly to $0$ at $x=l$ and $x=r$.
    \[
    \kappa_{\triangle}(x; l, c, r) = 
    \begin{cases}
    0, & x \notin [l, r], \\[6pt]
    \dfrac{x - l}{c - l}, & l \le x \le c, \\[10pt]
    \dfrac{r - x}{r - c}, & c \le x \le r, \\[10pt]
    \mathbf{1}\{x = c\}, & l = c = r.
    \end{cases}
    \]
    
  \State \textbf{Weighted mean estimate:} $\widehat{m}\gets argmax_t (W_t))$
  \State \Return $\widehat{m}$
\EndFunction
\end{algorithmic}
\end{algorithm}

\subsection{Empirically-weighted Distributed Connectivity Factor (\texorpdfstring{$e\mathcal{DCF}$}{eDCF})}
\label{subsec:edcf}

The $e\mathcal{DCF}$ method extends the Distributed Connectivity Factor ($\mathcal{DCF}$) framework by incorporating an empirically generated reference model for neighbor counts, improving robustness and accuracy under real-world noise and finite sampling conditions.

\vspace{1em}

\subsubsection{Relation to \texorpdfstring{$\mathcal{DCF}$}{DCF}}

Definitions and notation for intrinsic dimension, neighbor counts, and point-wise connectivity contributions remain unchanged from $\mathcal{DCF}$. The $e\mathcal{DCF}$ replaces the fixed theoretical bounds with empirically derived neighbor count references obtained from a pre-processing step. This calibration uses synthetic datasets sampled from \( t \)-dimensional hyperspheres with comparable size and noise as the input data to accurately model expected neighbor counts \( r_t \) for each intrinsic dimension \( t \). \\
We note that in a standard hat-function framework with ordered reference points, a point can contribute to at most two dimensions. However, in $e\mathcal{DCF}$, the empirically generated \( r_t \) values are not guaranteed to be monotonic. This allows the support intervals \( [r_{t-1}, r_{t+1}] \) of the hat functions to overlap, meaning a single point's neighbor count could theoretically contribute to more than two dimensions.

\vspace{1em}

\subsubsection{Empirical Reference Model Generation}

For each \( t \in [1, D_{\max}] \), a synthetic \( t \)-dimensional hypersphere is sampled with the same point count and noise level as the dataset. The average neighbor count \( r_t \) is computed by pairwise comparisons within a grid of spacing \( \epsilon \), and then taking their mean:
\[
\text{For } i = 1, \ldots, N, \quad \text{for } j = i+1, \ldots, N:
\]
\[
\quad \text{Increment counts if } \|x_i - x_j\|_{\infty} \leq \epsilon \text{ and } i \neq j.
\]
We then take the mean to get \( r_t \). These \( r_t \) values form an empirical lookup that replaces theoretical bounds for connectivity-based dimension estimation. A map is made from \{number of points, noise, dimension\} to \{average neighbor count\}, which is then stored to act as a look-up table for future runs. By caching this way, and creating buckets for values (such as number of points), we can reduce the size of this map while maintaining accuracy. \\
To pass the value of noise to the reference model generator, we require the amount of noise in the dataset, which we obtain using fast graph-based denoising (FGBD) by \cite{Watanabe2024} for lower ambient dimensions (upto 3), and SURE \cite{Stein1981} for higher ambient dimensions (local PCA / subspace methods can also be used). We note that the time taken for noise estimation is negligible compared to the rest of the algorithm, and thus is not included in our analysis.

If you have enough compute to store $(2.33)^{D_{max}}$ number of points, you can directly use $\mathcal{}^L{CF}_m^n * (3^m - 1)$ as the average neighbor count, which reduces the model generation time complexity to $\Theta\!(D_{max} / k)$. But, this method is computationally infeasible for us due to limited compute, and thus we use the empirical weights.

\vspace{1em}

\subsubsection{Weighted Membership Assignment}

Using the empirical counts \( r_t \), $e\mathcal{DCF}$ employs triangular cap functions \( f_t(x) \). For boundary conditions, we define \( r_{-1} \coloneqq r_0 \) and \( r_{D_{\max}+1} \coloneqq r_{D_{\max}} \). The function is defined as:
\begin{equation}
f_t(x)=
\begin{cases}
0, & x \notin [\,r_{t-1},\, r_{t+1}\,],\\[6pt]
\min\!\Bigl\{1,\ \dfrac{x - r_{t-1}}{\,r_t - r_{t-1}\,}\Bigr\},
& r_{t-1} \le x \le r_t\ \ \text{and}\ \ r_t > r_{t-1},\\[10pt]
\min\!\Bigl\{1,\ \dfrac{r_{t+1} - x}{\,r_{t+1} - r_t\,}\Bigr\},
& r_t \le x \le r_{t+1}\ \ \text{and}\ \ r_{t+1} > r_t,\\[10pt]
\mathbf{1}\{x = r_t\}, & r_{t-1} = r_t = r_{t+1}\ \text{(degenerate plateau).}
\end{cases}
\end{equation}

The absolute neighbor count \( c_i \) of each point is then converted into the fractional membership vector \( F(c_i) = [f_0(c_i), \ldots, f_{D_{\max}}(c_i)]^\top \).

\vspace{1em}

\subsubsection{Intrinsic Dimension Estimation}

The dataset's total weight vector \( W \) is defined as the sum of the fractional membership vectors \( F(c_i) \) over all points \( x_i \in S \), where \( c_i \) is the observed neighbor count of the point. Formally,
\begin{equation}
W = \sum_{i=1}^{|S|} F(c_i) = \begin{bmatrix} W_0 \\ W_1 \\ \vdots \\ W_d \end{bmatrix},
\end{equation}
where
\begin{itemize}
    \item \( S \) is the set of sampled points,
    \item \( |S| \) is the number of points,
    \item \( D_{max} \leq n \) is the maximum intrinsic dimension considered,
    \item \( c_i \) is the absolute neighbor count of the \(i\)-th point,
    \item \( F(c_i) = [f_0(c_i), f_1(c_i), \ldots, f_{D_{max}}(c_i)]^\top \) is the fractional membership vector for point \(i\),
    \item \( W_t \) is the total weight (sum of influences) for topology dimension \( t \) over the entire dataset.
\end{itemize}

The estimated intrinsic dimension \( \hat{m} \) is then computed as the weighted average of all possible dimensions, rounded to the nearest integer:
\begin{equation}
\hat{m} = \mathrm{round}\left(\frac{\sum_{t=0}^{D_{max}}t \cdot W_t}{\sum_{t=0}^{D_{max}}W_t}\right).
\end{equation}

This weighted average captures the expected intrinsic dimension by aggregating contributions from all candidate dimensions according to their total weight.

\vspace{1em}

\subsubsection{Computational Complexity and Optimizations}

The original theoretical bounds and neighbor count calculations exhibit runtime complexity on the order of \( O(N^2 d) \), where \( N \) is the number of points and \( d \) the ambient dimension. Another basic method, which is exponential in dimensions and uses recursive building of neighbour sets can also be used, but it is expensive.

Introducing the empirical reference model adds an amortized overhead for computing neighbor counts across multiple intrinsic dimensions \( D_{\max} \). However, this cost is amortized across multiple runs by caching reference values keyed on intrinsic dimension, point count, and noise level.

Further computational improvements include:

\begin{itemize}
\item Highly parallelized pairwise neighbor calculations, yielding effective runtime
\begin{equation}
T_{\mathrm{total}}^{\mathrm{pair}}(N,d,D_{\max},k)
\;=\;
\Theta\!\left(
\frac{N^{2} \, D_{\max}^{2}}{k}
\;+\;
\mathbf{1}_{[d\le \gamma]}\,\frac{N\,3^{d}}{k}
\;+\;
\mathbf{1}_{[d>\gamma]}\,\frac{N^{2}\,d}{k}
\right) 
\end{equation}
with k processors.

\item Potential replacement of brute-force neighborhood counts with KD-tree \cite{Bentley1975} methods, lowering complexity to approximately:
\begin{equation}
T_{\mathrm{total}}^{\mathrm{kd}}(N,d,D_{\max},k)
\;=\;
\Theta\!\left(
\frac{N\log N \, D_{\max}^{2}}{k}
\;+\;
\mathbf{1}_{[d\le \gamma]}\,\frac{N\,3^{d}}{k}
\;+\;
\mathbf{1}_{[d>\gamma]}\,\frac{N\log N \, d}{k}
\right)
\end{equation}

\item Bucketing and approximate caching strategies to generalize the empirical lookup, eventually reducing amortized cost to:
\begin{equation}
T_{\mathrm{total,\,cached}}^{\mathrm{kd}}(N,d,k)
\;=\;
\Theta\!\big(
\mathbf{1}_{[d\le \gamma]}\,\tfrac{N\,3^{d}}{k}
\;+\;
\mathbf{1}_{[d>\gamma]}\,\tfrac{N\log N \, d}{k}
\big)
\end{equation}

\item Note: For high values of d, the method reduces back to brute-force search due to KD-Tree limitations, and has a time complexity of:
\begin{equation}
T_{\mathrm{total,\,cached}}^{\mathrm{pair}}(N,d,k)
\;=\;
\Theta\!\big(
\mathbf{1}_{[d\le \gamma]}\,\tfrac{N\,3^{d}}{k}
\;+\;
\mathbf{1}_{[d>\gamma]}\,\tfrac{N^{2}\,d}{k}
\big)
\end{equation}

\item The value of $\gamma$ can be chosen by comparing $N3^d$ and $N^2d$ and choosing the floor of the value for which they are equal, or a pre-decided constant value can be used.

\item Since most work happens in higher ambient dimensions, usually the brute-force degraded form applies, as does in methods such as TWO-NN and MLE.
\end{itemize}

\vspace{1em}

\subsubsection{Adaptive Target Scaling Heuristic}

To better capture neighborhood statistics in higher ambient dimensions, we adopt an adaptive target percentage heuristic:
\begin{equation}
\text{Target Information Percentage} = \min\left(95.0, \text{base\_target} + 3 \sqrt{n}\right),
\end{equation}

where \( n \) is the ambient dimension and base\_target is a tunable hyperparameter (in practice, we have observed that 50 \% works best for a wide variety of cases).

This empirically improves neighborhood coverage, particularly in higher-dimensional settings. We do note that this is just a heuristic and is not necessary, nor is it optimal and can be further improved.

\subsubsection{Algorithm for \texorpdfstring{$e\mathcal{DCF}$}{eDCF}}

Below is the algorithm for $e\mathcal{DCF}$:

\begin{algorithm}[H]
\caption{$e\mathcal{DCF}$}
\begin{algorithmic}[1]
\Require Dataset $X=\{x_i\in\mathbb{R}^d\}_{i=1}^n$, Information Percentage Range $IP_{range}=[IP_{low}, IP_{high}]$
\Ensure Estimated intrinsic dimension $\widehat{m}\in\mathbb{W}$

\Function{eDCF}{$X,\,IP$}
  \State \textbf{Normalize to unit box:} Affinely rescale each axis so $X\subset[0,1]^d$.
  \State \textbf{Search over spacing $s$:} Find $s^\star$ such that gridding at step $s^\star$ retains an $IP\%$ of points, $IP \in IP_{range}$.
  \State \textbf{Grid and extract representatives:} Map each $x\in X$ to its grid representative at step $s^\star$ and keep the unique set $\mathcal{C}\subset\mathbb{R}^d$.

  \State \textbf{Estimate noise in $X$:} Use local PCA / FGBD / SURE to estimate noise in $X$, and store it in $\sigma$.
  
  \State $\mathcal{R} \gets \Call{GenerateReferenceModel}{n=|X|,\,d,\,D_{max},\,IP,\,\sigma}$ \Comment{$\mathcal{R}=\{\mu_m\}_{m=0}^{D_{max}}$}

  \State \textbf{Calculate neighbor counts:} For each point $u \in \mathcal{C}$, compute its neighbor count $c(u)$ as:
    \[
        c(u) \gets \big|\{v\in\mathcal{C}\setminus\{u\}:\ \|u-v\|_\infty\le s^\star\}\big|.
    \]
    \Comment{Calculated using efficient grid-based neighbour counting}
    
  \State \textbf{Dimensional scoring (summation of influences):} For $m=0,\dots,D_{max}$ set
    \[
        W[m]\ \gets\ \sum_{u \in \mathcal{C}} \kappa_{\triangle}\!\big(c(u);\,\mu_{m-1},\,\mu_m,\,\mu_{m+1}\big),
    \]
    where $c(u)$ is the neighbor count for each point $u$, and $\kappa_{\triangle}$ is the piecewise-linear “hat’’ with peak $1$ at $\mu_m$ and zeros at adjacent $\mu_{m-1},\mu_{m+1}$ (one-sided at the boundaries $m{=}0$ and $m{=}D_{max}$, using the values in $\mathcal{R}$).
  \State \textbf{Normalize weights:} $\pi[m]\gets W[m]\big/\sum_{k=0}^{D_{max}} W[k]$.
  \State \textbf{Weighted mean estimate:} $\widehat{m}\gets \mathrm{round}\!\left(\sum_{m=0}^{D_{max}} m\,\pi[m]\right)$
  \State \Return $\widehat{m}$
\EndFunction

\Statex

\Function{GenerateReferenceModel}{$n,\,d,\,D_{max},\,IP,\,\sigma$}
  \State \Comment{$D_{max}$ = maximum candidate intrinsic dimension $\le$ ambient dimension of $X$}
  \For{$m=0$ \textbf{to} $D_{max}$}
    \State $X^{(m)} \gets$ \textit{generate $m$-dim hypersphere with $n$ points and noise $\sigma$}
    \State \textit{Normalize $X^{(m)}$ to the unit box.}
    \State \textit{Search for spacing $s^\star$ that retains an $IP\%$ of points in $X^{(m)}$, $IP \in IP_{range}$}
    \State \textit{Grid $X^{(m)}$ at step $s^\star$ to get unique set $\mathcal{C}^{(m)}$.}
    \If{$|\mathcal{C}^{(m)}|<2$} 
        \State $\bar c_m \gets 0$
    \Else
        \State \textit{For each $u \in \mathcal{C}^{(m)}$, calculate neighbor count $c(u) = \big|\{v\in\mathcal{C}^{(m)}\setminus\{u\}:\ \|u-v\|_\infty\le s^\star\}\big|$.}
        \State \textit{Aggregate counts: $\bar c_m \gets \frac{1}{|\mathcal{C}^{(m)}|}\sum_{u\in\mathcal{C}^{(m)}} c(u)$.}
    \EndIf
    \State $\mu_m \gets \bar c_m$ \Comment{Store the final average count for this dimension}
  \EndFor
  \State \Return $\{\mu_m\}_{m=0}^{D_{max}}$
\EndFunction
\end{algorithmic}
\end{algorithm}

\vspace{1em}

\section{Results}

Explicit values and extra results are provided in appendix \ref{datarunsextra} and \ref{eDCFrunsvalues}.

\subsection{\texorpdfstring{$e\mathcal{DCF}$}{eDCF} results on Benchmark Manifolds}

We compare TWO-NN and MLE against our method ($e\mathcal{DCF}$) on benchmark manifolds provided in the scikit-dimension (skdim) library \cite{scikit-dimension}, which generates a commonly used benchmark set of synthetic manifolds with known intrinsic dimension described by Hein et al. \cite{HeinAudibert2005} and Campadelli et al. \cite{Campadelli2015}. The TWO-NN and MLE implementations used are also from the skdim library. We compare the three methods using Mean Absolute Error (MAE), Mean Signed Error and Accuracy.

For $e\mathcal{DCF}$, we use a value of 50 \% on the IP scale. We generate benchmark manifolds with 1 \%, 10 \% and 30 \% gaussian noise. For ease of computation, we use a value of $D_{max} = 50$ and $\gamma = 3$ for our experiments. For TWO-NN, we use skdim defaults. For MLE, we use skdim defaults and n\_neighbors = 20.

\begin{figure}[H]
    \centering
    \begin{tabular}{ccc}
        \subcaptionbox{}{\includegraphics[width=0.3\linewidth]{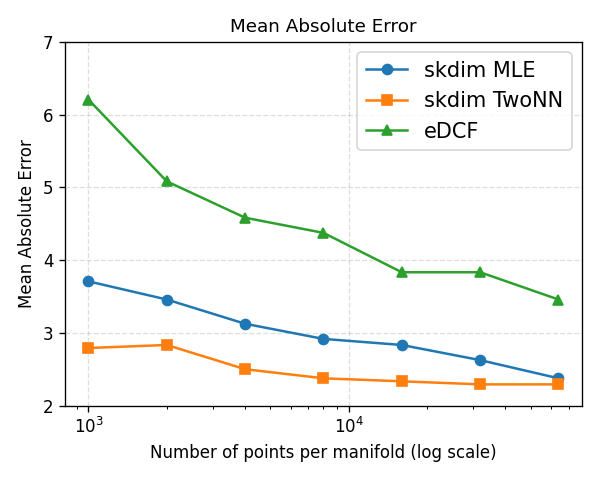}} &
        \subcaptionbox{}{\includegraphics[width=0.3\linewidth]{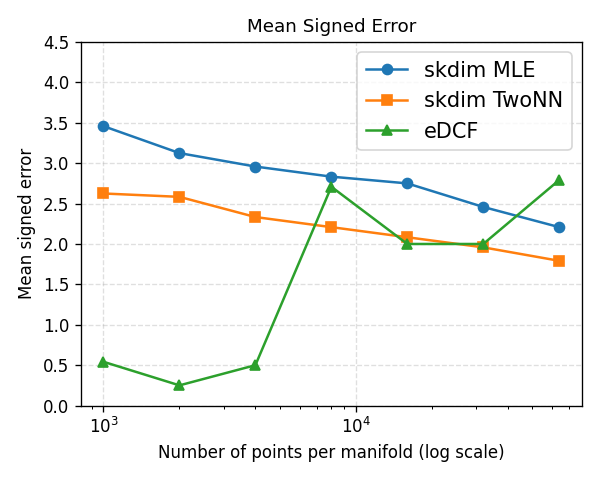}} &
        \subcaptionbox{}{\includegraphics[width=0.3\linewidth]{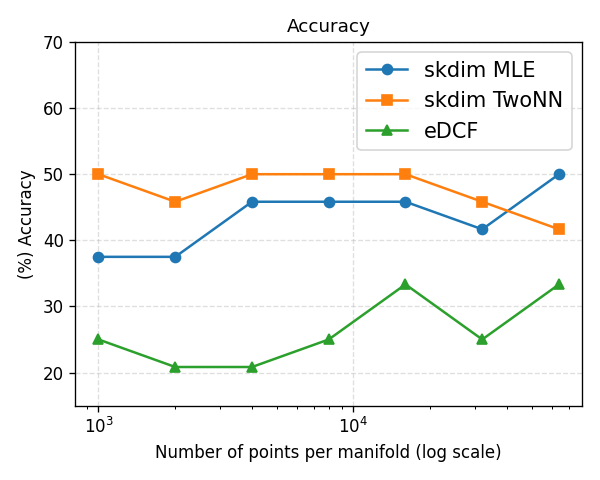}} \\
        \subcaptionbox{}{\includegraphics[width=0.3\linewidth]{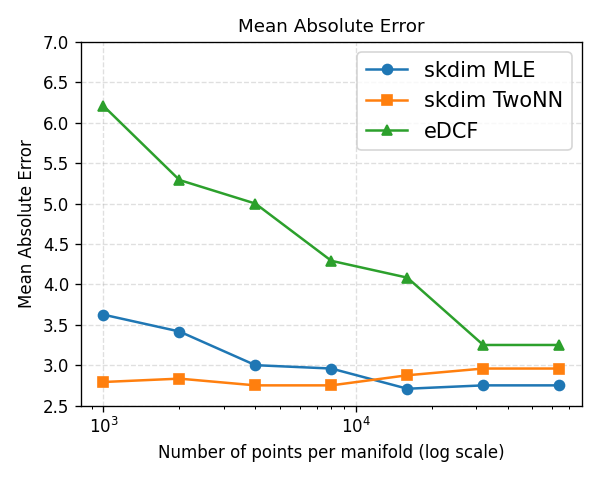}} &
        \subcaptionbox{}{\includegraphics[width=0.3\linewidth]{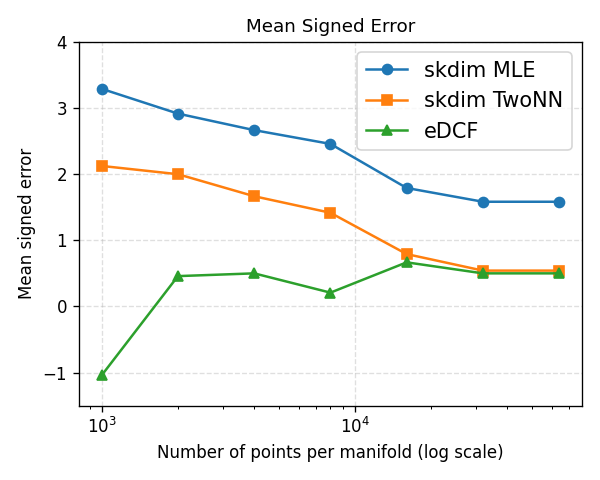}} &
        \subcaptionbox{}{\includegraphics[width=0.3\linewidth]{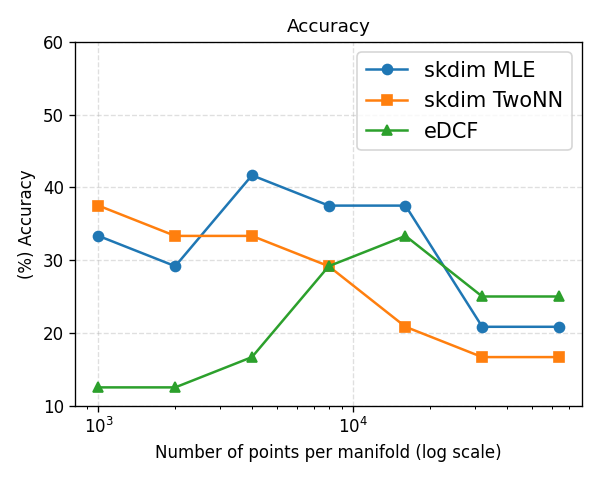}} \\
        \subcaptionbox{}{\includegraphics[width=0.3\linewidth]{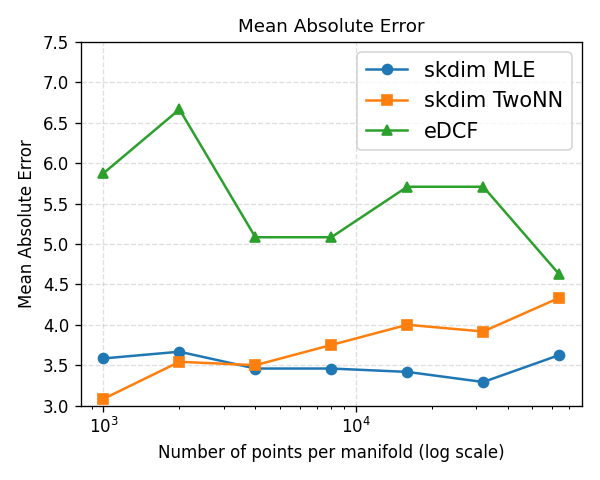}} &
        \subcaptionbox{}{\includegraphics[width=0.3\linewidth]{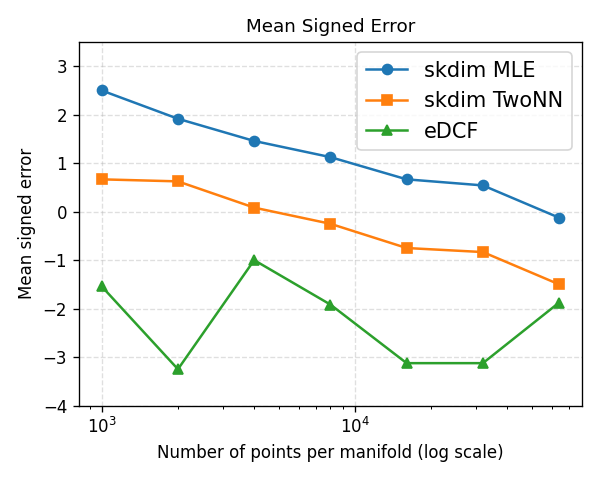}} &
        \subcaptionbox{}{\includegraphics[width=0.3\linewidth]{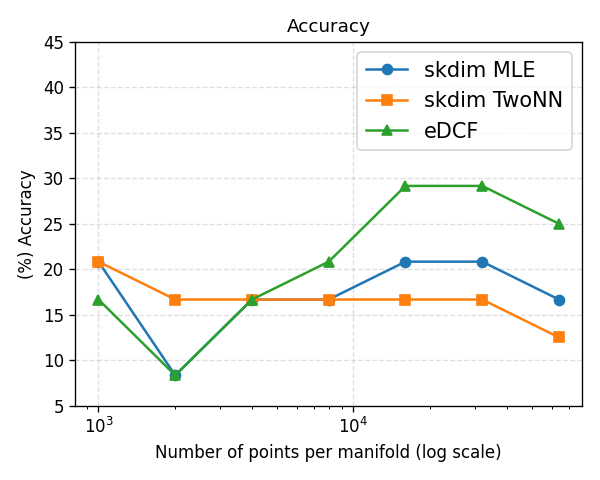}} \\
    \end{tabular}
    \caption{Performance comparison of TWO-NN, MLE, and $e\mathcal{DCF}$ on benchmark manifolds under varying noise conditions. (a) represents Mean Absolute Error for 1\% noise, (b) represents Mean Signed Error for 1\% noise, (c) represents Accuracy for 1\% noise, (d) represents Mean Absolute Error for 10\% noise, (e) represents Mean Signed Error for 10\% noise, (f) represents Accuracy for 10\% noise, (g) represents Mean Absolute Error for 30\% noise, (h) represents Mean Signed Error for 30\% noise, and (i) represents Accuracy for 30\% noise. All metrics are plotted against the number of points per manifold on a logarithmic scale.}
    \label{fig:combined9plots}
\end{figure}

Across all noise levels, $e\mathcal{DCF}$’s performance tightens with increasing data: its error decreases and the percentage of exact dimension estimates rises as sample size grows. At 1\% noise, $e\mathcal{DCF}$’s mean absolute error (MAE) drops markedly from \textbf{6.208} (at 1k samples) to \textbf{3.458} (at 64k), while exact matches increase from \textbf{25.0\%} to \textbf{33.3\%}. Its bias (signed error) slightly increases from \textbf{+0.542} at 1k to \textbf{+2.792} by 64k.

At the intermediate noise setting (10\% noise), $e\mathcal{DCF}$ contracts MAE from \textbf{6.208} to \textbf{3.250} between 1k and 64k samples, and its exact-match rate rises steadily from \textbf{12.5\%} (1–2k) to \textbf{25.0\%} (32k–64k), with a maximum of 
\textbf{33.3\%} at 16k. The signed error remains small 
and moves toward zero ($\bm{-1.042}$ \text{ at 1k to } $\bm{+0.500}$ \text{ at 64k})\text{, indicating well-
balanced estimation in large samples.}

For 30\% noise, $e\mathcal{DCF}$ maintains improvement in MAE with sample size (\textbf{5.875} at 1k to \textbf{4.625} at 64k) and its exact-match rate closely follows the trends seen at lower noise (\textbf{16.7\%} to \textbf{25.0\%}, peaking at \textbf{29.2\%}), though the signed error remains moderately negative.

$e\mathcal{DCF}$ frequently outperforms both in exact dimension recovery for large $N$ in moderate to high noise. For example, at 64k points and 30\% noise, $e\mathcal{DCF}$ achieves \textbf{25.0\%} exact matches, versus MLE’s \textbf{16.7\%} and TWO-NN’s \textbf{12.5\%}. For $e\mathcal{DCF}$, as sample size increases, its precision in identifying the correct dimension improves - even when its average error remains marginally above the baseline methods.

\subsection{\texorpdfstring{$\mathcal{DCF}$}{DCF} results on Synthetic Data} \label{dcfresults}

The following results are for the $\mathcal{DCF}$ framework, with high number of points in the dataset and boundary, thus fitting the criteron for $\mathcal{DCF}$ usage. We run KNN on the Concentric Circles dataset (CCD), Decision Tree on Overlapping Concentric Circles dataset (OCCD), KNN on Barnsley Fern (BF) dataset, and Decision Tree on Sierpinski Carpet (SC) dataset. All dataset generation details and further experiments are provided in the appendix.

\noindent
\begin{minipage}{\textwidth}
    \subsubsection{KNN:}
    % --- Plots Side by Side ---
    \begin{figure}[H]
        \centering
        \begin{subfigure}{0.48\textwidth}
            \centering
            \includegraphics[width=\linewidth,trim=0 0 0 30,clip]{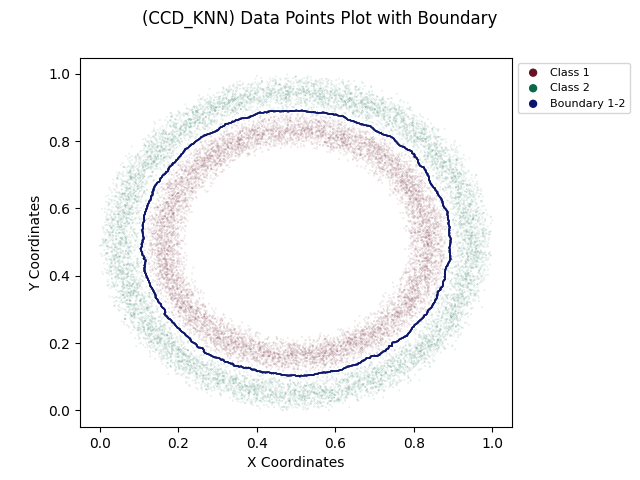} % Updated path
            \caption{Boundary Plot with Datapoints}
            \label{fig:knn_image2}
        \end{subfigure}\hfill
        \begin{subfigure}{0.48\textwidth}
            \centering
            \includegraphics[width=\linewidth,trim=0 0 0 30,clip]{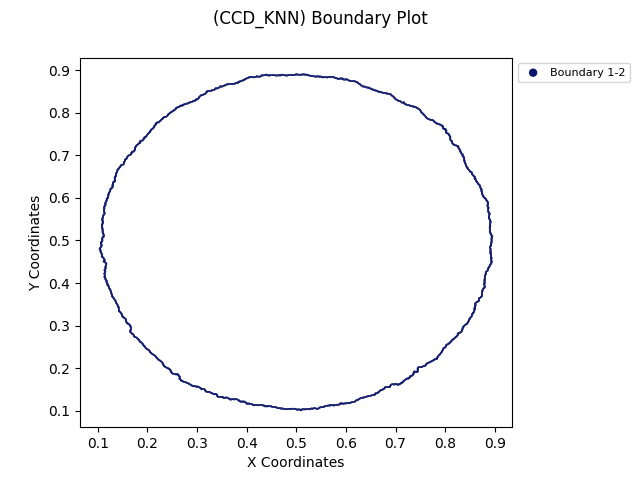} % Updated path
            \caption{Boundary Plot}
            \label{fig:knn_image3}
        \end{subfigure}
    
        \caption{Plots of KNN on CCD Dataset}
    \end{figure}
    \label{fig:knn_small_images} % Updated label

    % --- Tables Side by Side ---
    \vspace{1em}
    \noindent
    \begin{minipage}{0.48\textwidth}
        \centering
        \renewcommand{\arraystretch}{1.5}
        \begin{tabular}{lc}
            \toprule
            Fractal Dimension (Boundary)          & 1.0306  \\
            Connectivity Factor (Boundary)        & 0.3968  \\
            Topological Dimension (${}^L\mathcal{CF}$ Based)      & 1       \\
            Topological Dimension ($\mathcal{DCF}$ Based)  & 1       \\
            \bottomrule
        \end{tabular}
    \end{minipage}%
    \hfill
    \begin{minipage}{0.48\textwidth}
        \centering
        \renewcommand{\arraystretch}{1.5}
        \begin{tabular}{lc}
            \toprule
            Weight (Topology 0, Boundary)         & 0.0000 \\
            Weight (Topology 1, Boundary)         & 0.8042 \\
            Weight (Topology 2, Boundary)         & 0.1957 \\
            Fractal Dimension (Object 1)          & 1.7473 \\
            Fractal Dimension (Object 2)          & 1.7120 \\
            \bottomrule
        \end{tabular}
    \end{minipage}

    \captionsetup{skip=2pt}
    \captionof{table}{KNN boundary characteristics results} % Updated caption
    \label{tab:knn_boundary_characteristics} % Updated label
\end{minipage}

\noindent
\begin{minipage}{\textwidth}
    \subsubsection{Decision Tree:}

    % --- Plots Side by Side ---
    \centering
    \begin{figure}[H]
        \begin{subfigure}{0.48\textwidth}
            \centering
            \includegraphics[width=\linewidth,trim=0 0 0 30,clip]{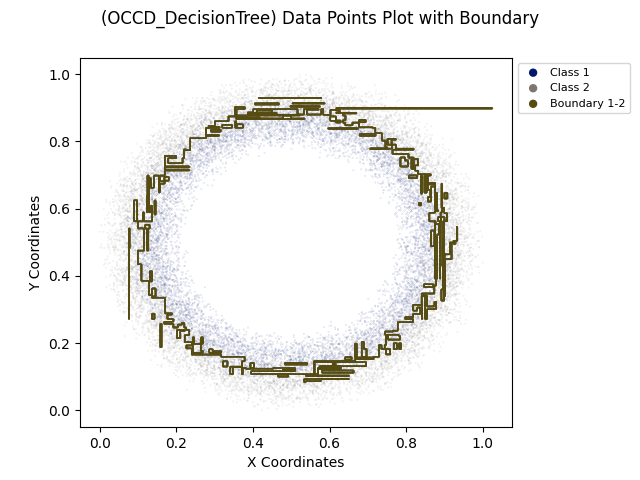}
            \caption{Boundary Plot with Datapoints}
            \label{fig:image2}
        \end{subfigure}\hfill
        \begin{subfigure}{0.48\textwidth}
            \centering
            \includegraphics[width=\linewidth,trim=0 0 0 30,clip]{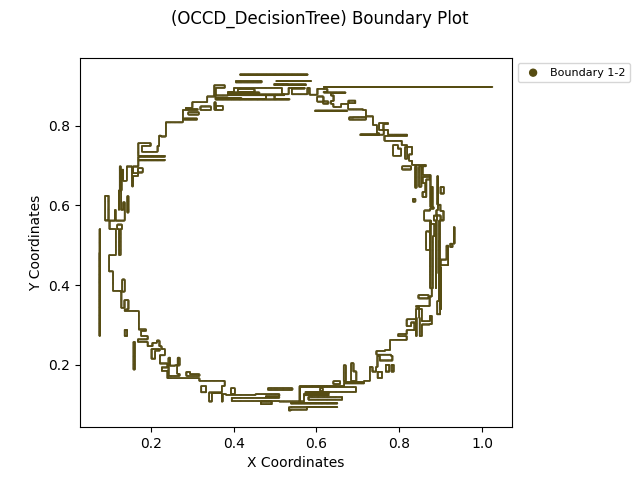}
            \caption{Boundary Plot}
            \label{fig:image3}
        \end{subfigure}
    
        \caption{Plots of Decision Tree on OCCD Dataset}
    \end{figure}
    \label{fig:four_small_images}

    % --- Tables Side by Side ---
    \vspace{1em}
    \noindent
    \begin{minipage}{0.48\textwidth}
        \centering
        \renewcommand{\arraystretch}{1.5}
        \begin{tabular}{lc}
            \toprule
            Fractal Dimension (Boundary)          & 1.4441  \\
            Connectivity Factor (Boundary)        & 0.2713  \\
            Topological Dimension (${}^L\mathcal{CF}$ Based)      & 1       \\
            Topological Dimension ($\mathcal{DCF}$ Based)  & 1       \\
            \bottomrule
        \end{tabular}
    \end{minipage}%
    \hfill
    \begin{minipage}{0.48\textwidth}
        \centering
        \renewcommand{\arraystretch}{1.5}
        \begin{tabular}{lc}
            \toprule
            Weight (Topology 0, Boundary)         & 0.0000 \\
            Weight (Topology 1, Boundary)         & 0.9715 \\
            Weight (Topology 2, Boundary)         & 0.0284 \\
            Fractal Dimension (Object 1)          & 1.7625 \\
            Fractal Dimension (Object 2)          & 1.7662 \\
            \bottomrule
        \end{tabular}
    \end{minipage}

    \captionsetup{skip=2pt}
    \captionof{table}{Boundary characteristics results}
    \label{tab:boundary_characteristics}
\end{minipage}

\noindent
\begin{minipage}{\textwidth}
    \subsubsection{KNN:}
    % --- Plots Side by Side ---
    \centering
    \begin{figure}[H]
        \begin{subfigure}{0.48\textwidth}
            \centering
            \includegraphics[width=\linewidth,trim=0 0 0 30,clip]{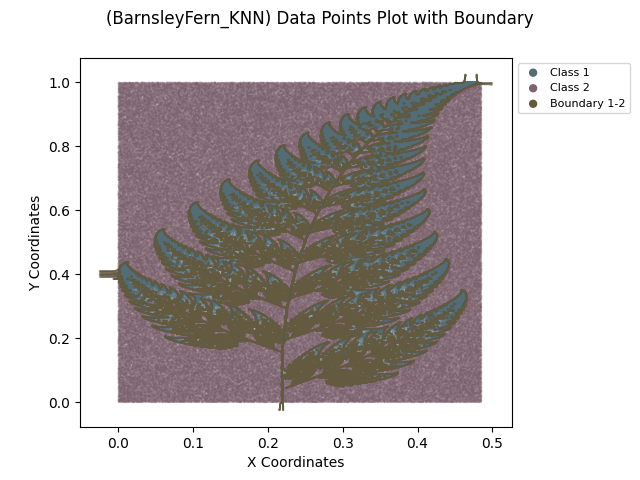}
            \caption{Boundary Plot with Datapoints}
            \label{fig:knn_bf_image2}
        \end{subfigure}\hfill
        \begin{subfigure}{0.48\textwidth}
            \centering
            \includegraphics[width=\linewidth,trim=0 0 0 30,clip]{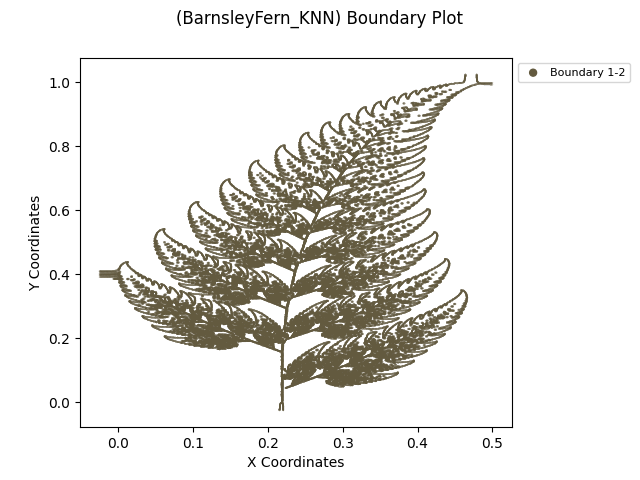}
            \caption{Boundary Plot}
            \label{fig:knn_bf_image3}
        \end{subfigure}
    
        \caption{Plots of KNN on BF Dataset}
    \end{figure}
    \label{fig:knn_bf_small_images}

    % --- Tables Side by Side ---
    \vspace{1em}
    \noindent
    \begin{minipage}{0.48\textwidth}
        \centering
        \renewcommand{\arraystretch}{1.5}
        \begin{tabular}{lc}
            \toprule
            Fractal Dimension (Boundary)          & 1.7930  \\
            Connectivity Factor (Boundary)        & 0.4325  \\
            Topological Dimension (${}^L\mathcal{CF}$ Based)      & 1       \\
            Topological Dimension ($\mathcal{DCF}$ Based)  & 1       \\
            \bottomrule
        \end{tabular}
    \end{minipage}%
    \hfill
    \begin{minipage}{0.48\textwidth}
        \centering
        \renewcommand{\arraystretch}{1.5}
        \begin{tabular}{lc}
            \toprule
            Weight (Topology 0, Boundary)         & 0.0000 \\
            Weight (Topology 1, Boundary)         & 0.7565 \\
            Weight (Topology 2, Boundary)         & 0.2434 \\
            Fractal Dimension (Object 1)          & 1.8340 \\
            \bottomrule
        \end{tabular}
    \end{minipage}

    \captionsetup{skip=2pt}
    \captionof{table}{KNN boundary characteristics results}
    \label{tab:knn_bf_boundary_characteristics}
\end{minipage}

\noindent
\begin{minipage}{\textwidth}
    \subsubsection{Decision Tree:}

    % --- Plots Side by Side ---
    \centering
    \begin{figure}[H]
        \begin{subfigure}{0.48\textwidth}
            \centering
            \includegraphics[width=\linewidth,trim=0 0 0 30,clip]{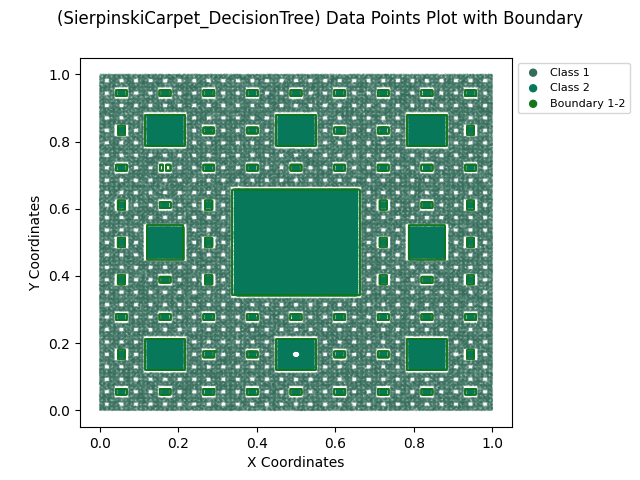}
            \caption{Boundary Plot with Datapoints}
            \label{fig:sc_image2}
        \end{subfigure}\hfill
        \begin{subfigure}{0.48\textwidth}
            \centering
            \includegraphics[width=\linewidth,trim=0 0 0 30,clip]{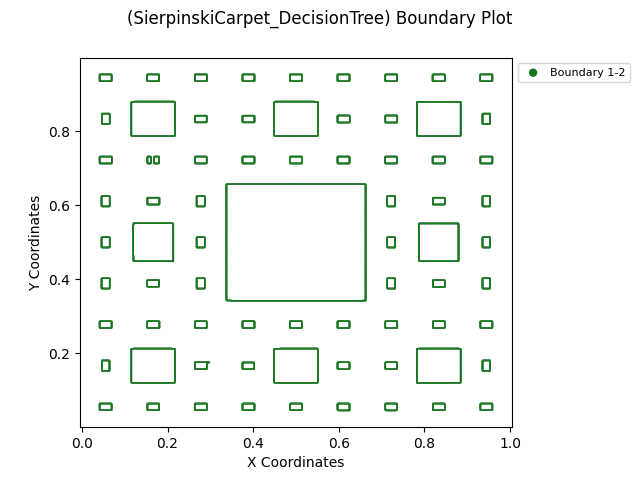}
            \caption{Boundary Plot}
            \label{fig:sc_image3}
        \end{subfigure}
    
        \caption{Plots of Decision Tree on SCD Dataset}
    \end{figure}
    \label{fig:sc_small_images}

    % --- Tables Side by Side ---
    \vspace{1em}
    \noindent
    \begin{minipage}{0.48\textwidth}
        \centering
        \renewcommand{\arraystretch}{1.5}
        \begin{tabular}{lc}
            \toprule
            Fractal Dimension (Boundary)          & 1.0188  \\
            Connectivity Factor (Boundary)        & 0.2543  \\
            Topological Dimension (${}^L\mathcal{CF}$ Based)      & 1       \\
            Topological Dimension ($\mathcal{DCF}$ Based)  & 1       \\
            \bottomrule
        \end{tabular}
    \end{minipage}%
    \hfill
    \begin{minipage}{0.48\textwidth}
        \centering
        \renewcommand{\arraystretch}{1.5}
        \begin{tabular}{lc}
            \toprule
            Weight (Topology 0, Boundary)         & 0.0000 \\
            Weight (Topology 1, Boundary)         & 0.9941 \\
            Weight (Topology 2, Boundary)         & 0.0058 \\
            Fractal Dimension (Object 1)          & 1.8967 \\
            \bottomrule
        \end{tabular}
    \end{minipage}

    \captionsetup{skip=2pt}
    \captionof{table}{Boundary characteristics results}
    \label{tab:boundary_characteristics}
\end{minipage}

\vspace{1em}

Across datasets like Concentric Circle Data, Overlapping Circles, Barnsley Fern, and Sierpinski Carpet, the estimated boundary topological dimension was consistently correctly estimated as 1. Fractal dimensions of boundaries ranged from about 1.0 to 1.79, with connectivity factors between roughly 0.25 and 0.43. Weighted topology contributions showed a dominant one-dimensional membership with smaller fractions attributed to zero- and two-dimensional topologies. For fractal data, fractal dimensions aligned with known fractal characteristics, mostly between 1.6 and 1.9. Boundary plots confirmed effective identification of data boundaries consistent with the underlying dataset geometry. KNN CCD boundary was correctly identified as non-fractal, Decision Tree OCCD was identified as having some fractal nature due to inherent complexity and self similarity across sections of the boundary, KNN BF boundary was correctly identified as having fractal nature, and Decision Tree SC boundary was correctly identified and being non-fractal (while SC is a fractal, the decision boundary was too simple and not self similar to a great enough depth to truly be considered a fractal).

\section{Discussion}

The results demonstrate that the Empirically-weighted Distributed Connectivity Factor ($e\mathcal{DCF}$) method provides a robust, scalable, and noise-resilient approach for intrinsic dimension estimation on diverse synthetic benchmark manifolds. Its design, centered on a grid-based neighbor framework and a novel Connectivity Factor formulation, enables efficient parallelization and scalability, distinguishing it from traditional distance-metric-dependent methods that face challenges in high-dimensional spaces.

A key factor underlying $e\mathcal{DCF}$’s robustness to noise is its reliance on local connectivity patterns rather than purely distance-based metrics. By discretizing space into uniform grid cells and defining neighbors through a fixed neighborhood structure, $e\mathcal{DCF}$ mitigates sensitivity to small perturbations in point locations caused by noise. Additionally, the empirical weighting mechanism calibrates neighbor contributions using reference models from synthetic noisy manifolds, allowing accurate fractional membership assignment under sampling variability and outliers. This contrasts with classical methods, such as TWO-NN and MLE, whose dependence on nearest-neighbor distances can be disproportionately impacted by noise.

Quantitatively, $e\mathcal{DCF}$ exhibits consistent improvement with increasing sample size across all noise levels: mean absolute error (MAE) decreases and the proportion of exact dimension matches rises. At low noise (1\%), the method achieves substantial reduction in error alongside increasing exact-match rates, indicating reliable convergence to true intrinsic dimensionality as data density grows. The signed error shows a mild positive bias at small sample sizes but stabilizes with larger samples.

At moderate noise (10\%), the signed error approaches zero for large samples, signifying balanced and unbiased dimension estimation. The highest exact-match rate at intermediate sample sizes (around 16k) suggests an optimal scale where the local neighborhood structure is best captured, with diminishing returns beyond this point likely due to saturation or finite sampling effects.

Under high noise conditions (30\%), $e\mathcal{DCF}$ maintains error reduction and consistent exact-match trends, though gains are more modest. The observed negative bias in signed error indicates a slight underestimation tendency, plausibly resulting from noise-induced disruption of local connectivity.

Comparisons with TWO-NN and MLE reveal that despite occasionally higher MAE, $e\mathcal{DCF}$ consistently attains superior exact dimension recovery in large-sample, moderate-to-high noise regimes. This highlights $e\mathcal{DCF}$’s focus on precise identification of discrete dimensionality rather than minimizing average error alone, making it particularly suited for applications demanding reliable, topology-aware dimension inference.

Further, $e\mathcal{DCF}$’s dynamic neighborhood scaling via \emph{Information Percentage} allows adaptive tuning to data scale, enhancing flexibility in heterogeneous, noisy environments.

For fractal analysis using $\mathcal{DCF}$, as is seen in \ref{dcfresults}, it is accurate and reliable for low dimensional fractals.

\subsection{Limitations}
While the empirical results are encouraging, our study has several limitations.
\begin{enumerate}[leftmargin=*, itemsep=2pt]
    \item \textbf{Discretization choices:} The grid discretization and the target Information Percentage ($IP$) act as structural hyperparameters. We did not present a full ablation over spacing schedules/IP targets, so sensitivity to those choices remains partially characterized.
    \item \textbf{Empirical calibration:} $e\mathcal{DCF}$ relies on an empirical reference table (neighbor-count anchors across point counts/noise). Its fidelity depends on how well the synthetic calibration families match the data under study; domain shift or misestimated noise can bias assignments.
    \item \textbf{Computation at high ambient dimension:} Although the workflow parallelizes well, neighbor counting can revert to \(O(N^2 d)\) behavior in high \(d\), with nontrivial memory pressure. We did not benchmark GPU/approximate variants, nor wall-clock vs. baselines.
    \item \textbf{Scope of datasets/baselines:} Benchmark manifolds are synthetic; real-world evaluations are limited to illustrative boundary analyses (CCD/OCCD/BF/SC). Baselines focus on TWO-NN and MLE; broader comparisons (e.g., DANCo, MiND/ESS, kNN-graph estimators) were not included.
    \item \textbf{Bias behavior under noise:} While exact-hit rates generally increase with sample size across noise levels, signed-error trends can drift with noise and dataset scale. We did not analyze causes (e.g., density nonuniformity, curvature, class imbalance) in depth.
\end{enumerate}

\subsection{Future Work}
We outline several directions that address the above limitations and extend applicability.
\begin{enumerate}[leftmargin=*, itemsep=2pt]
    \item \textbf{LMU-CF:} Developing an appropriate empirical estimate / proxy for the $\mathcal{LMU-CF}$ method; running benchmarks for it.
    \item \textbf{Fraud detection:} Using $\mathcal{CF}$ for applications in fraud detection, using a methodology similar to that proposed in \cite{mzoom}.
    \item \textbf{Object Degradation Analysis:} Using $\mathcal{CF}$ to track changes in structure of point cloud objects, thus estimating rate of degradation.
    \item \textbf{Ablations and multi-resolution schemes:} Systematic sweeps over grid spacing/IP targets, plus multi-resolution ensembling (e.g., voting or stacking across IP scales) to reduce discretization bias.
    \item \textbf{Adaptive, data-driven calibration:} More accurate / robust methods to learn the membership caps and anchor counts directly from data via density-aware or noise-aware models; development of a better suited Adaptive Target Scaling Heuristic in $e\mathcal{DCF}$.
    \item \textbf{Scalability:} GPU implementations and approximate neighbor counting (e.g., LSH/IVF-PQ); streaming/online $e\mathcal{DCF}$ for evolving datasets; compressed caching for the calibration table.
    \item \textbf{Broader benchmarks and baselines:} Real-world high-dimensional corpora (vision, audio, graphs) and additional ID estimators beyond MLE/TWO-NN; stress tests on anisotropy, heavy-tailed noise, and strong density gradients.
    \item \textbf{Calibration diagnostics:} Per-dataset reliability diagrams for signed error, bias–variance decompositions, and sample-size curves that contextualize when exact-hits overtake baselines.
    \item \textbf{Boundary and application studies:} Extend boundary fractality analysis across modern classifiers (e.g., CNNs/transformers) and tasks; evaluate ties to generalization, drift/degradation monitoring, and fraud/outlier detection pipelines.
\end{enumerate}

\section{Conclusion}
Across synthetic benchmark manifolds with 1–30\% noise, $e\mathcal{DCF}$ exhibits a consistent empirical pattern: as sample size grows, mean absolute error decreases and the fraction of \emph{exact} intrinsic-dimension hits rises, often rivaling or surpassing baselines at medium–large \(N\), while typically posting slightly higher MAE than MLE/TWO-NN overall. On a suite of constructed datasets (CCD/OCCD/BF/SC), the framework also supports boundary-centric analyses. Taken together, these findings position $e\mathcal{DCF}$ as a practical, scalable default when reliability and exact recovery at realistic data volumes are prioritized. The identified limitations, most notably discretization sensitivity, empirical calibration dependence, and high-\(d\) compute, suggest clear next steps: multi-resolution and adaptive calibration strategies, uncertainty quantification, real-world validation, and engineering for large-scale deployment.

\section*{Acknowledgments}
This was was supported in part by the CSIS Department at BITS Pilani, K. K. Birla Goa Campus. We extend our thanks to Dr. Harikrishnan N B (CSIS Department at BITS Pilani, K. K. Birla Goa Campus) and Dr. Nithin Nagaraj (Head of Complex Systems Programme, NIAS, IISC Bangalore) for their guidance.

%Bibliography
\bibliographystyle{unsrt}  % or another style like plain, alpha, ieeetr, acm, etc.
\bibliography{references}  % without the .bib extension

\newpage

\appendix
\section{Appendix}

\subsection{Solving Recurrence Relation for Lower Bound} \label{proofofat}

The recurrence relation for $a_i$ is given by:
\[
a_i = a_{i-1} - \frac{1}{3}(a_{i-1} + 1) = \frac{2}{3}a_{i-1} - \frac{1}{3}
\]
with the initial condition $a_0 = 3^n - 1$. This can be written in a linear algebraic form:
\[
\begin{bmatrix} a_i \\ 1 \end{bmatrix} =
\begin{bmatrix} 2/3 & -1/3 \\ 0 & 1 \end{bmatrix}
\begin{bmatrix} a_{i-1} \\ 1 \end{bmatrix}
\]
Iterating this relation $i$ times gives:
\[
\begin{bmatrix} a_i \\ 1 \end{bmatrix} =
\begin{bmatrix} 2/3 & -1/3 \\ 0 & 1 \end{bmatrix}^i
\begin{bmatrix} a_0 \\ 1 \end{bmatrix}
\]
Let $A = \begin{bmatrix} 2/3 & -1/3 \\ 0 & 1 \end{bmatrix}$. We solve this by diagonalizing the matrix $A$, such that $A^i = P D^i P^{-1}$.

The matrix $A$ has eigenvalues $\lambda_1 = 2/3$ and $\lambda_2 = 1$. The corresponding diagonalization is:
\[
\begin{bmatrix} a_i \\ 1 \end{bmatrix} =
\begin{bmatrix} 1 & 1 \\ -1 & 0 \end{bmatrix}
\begin{bmatrix} (2/3)^i & 0 \\ 0 & 1^i \end{bmatrix}
\begin{bmatrix} 0 & -1 \\ 1 & 1 \end{bmatrix}
\begin{bmatrix} a_0 \\ 1 \end{bmatrix}
\]
Multiplying the matrices $P D^i P^{-1}$ first yields:
\[
\begin{bmatrix} a_i \\ 1 \end{bmatrix} =
\left(
\begin{bmatrix} 1 & 1 \\ -1 & 0 \end{bmatrix}
\begin{bmatrix} (2/3)^i & 0 \\ 0 & 1 \end{bmatrix}
\begin{bmatrix} 0 & -1 \\ 1 & 1 \end{bmatrix}
\right)
\begin{bmatrix} a_0 \\ 1 \end{bmatrix}
\]
\[
\implies
\begin{bmatrix} a_i \\ 1 \end{bmatrix} =
\begin{bmatrix} (2/3)^i & 1 - (2/3)^i \\ 0 & 1 \end{bmatrix}
\begin{bmatrix} a_0 \\ 1 \end{bmatrix}
\]
This gives the equation for $a_i$:
\[
a_i = (2/3)^i a_0 + (1 - (2/3)^i)(1) = (2/3)^i (a_0 - 1) + 1
\]
\[
a_i = (2/3)^i (a_0 + 1) - 1
\]
Substituting the initial condition $a_0 = 3^n - 1$, which means $a_0 + 1 = 3^n$:
\[
a_i = (2/3)^i (3^n) - 1 = \frac{2^i}{3^i} 3^n - 1
\]
\[
\implies \mathbf{a_i = 2^i 3^{n-i} - 1}
\]

\subsection{A Complete End-to-End Example: 1D Topology in 2D Space}

Let's calculate the upper bound for a 1D structure in a 2D space: $\boldsymbol{{}^U\mathcal{CF}_{1}^{2}}$.
\begin{itemize}
    \item Target Topology: $m=1$
    \item Ambient Space: $n=2$
\end{itemize}

\textbf{1. Elimination:}\\
The rule is to eliminate types from $a_0$ up to $a_{n-m-1}$. Here, $n-m-1 = 2-1-1 = 0$. So, we must eliminate type $\boldsymbol{a_0}$. The \textbf{remaining point types} are $\boldsymbol{a_1}$ and $\boldsymbol{a_2}$.

\textbf{2. Interactions:}\\
We will now use the interaction formula to calculate the total number of connections each remaining point type has with the other remaining point types.

\textbf{Interactions from an $a_1$ perspective ($x=1$):}
The total connections for an $a_1$ point is the sum of its connections to other $a_1$ points and to $a_2$ points: $\prescript{1}{}{\alpha}_{1}^{2} + \prescript{1}{}{\alpha}_{2}^{2}$.
\begin{itemize}
    \item \textbf{Connections to $a_1$ points ($m=1$):}
    \begin{align*}
        \prescript{1}{}{\alpha}_{1}^{2} &= \left[ 2^0 \cdot {}^1C_0 \cdot (2^{1-1} \cdot {}^{1}C_{1-1}) \right] + \left[ 2^1 \cdot {}^1C_1 \cdot (2^{1-0} \cdot {}^{1}C_{1-0}) \right] - {}^0C_0 \\
        &= [1 \cdot 1 \cdot (1 \cdot 1)] + [2 \cdot 1 \cdot (2 \cdot 1)] - 1 = 1 + 4 - 1 = 4
    \end{align*}
    \item \textbf{Connections to $a_2$ points ($m=2$):}
    \begin{align*}
        \prescript{1}{}{\alpha}_{2}^{2} &= \left[ 2^0 \cdot {}^1C_0 \cdot (2^{2-1} \cdot {}^{1}C_{2-1}) \right] + \left[ 2^1 \cdot {}^1C_1 \cdot (2^{2-0} \cdot {}^{1}C_{2-0}) \right] - {}^0C_{-1} \\
        &= [1 \cdot 1 \cdot (2 \cdot 1)] + [2 \cdot 1 \cdot (4 \cdot 0)] - 0 = 2 + 0 - 0 = 2
    \end{align*}
\end{itemize}
Total connections for an $a_1$ point = $4 + 2 = \textbf{6}$.

\textbf{Interactions from an $a_2$ perspective ($x=2$):}
The total connections for an $a_2$ point is the sum of its connections to $a_1$ points and to other $a_2$ points: $\prescript{2}{}{\alpha}_{1}^{2} + \prescript{2}{}{\alpha}_{2}^{2}$.
\begin{itemize}
    \item \textbf{Connections to $a_1$ points ($m=1$):}
    \begin{align*}
        \prescript{2}{}{\alpha}_{1}^{2} &= \left[ \dots \right]_{\text{i=0}} + \left[ 2^1 \cdot {}^2C_1 \cdot (2^{1-1} \cdot {}^{0}C_{1-1}) \right]_{\text{i=1}} + \left[ \dots \right]_{\text{i=2}} - {}^0C_{1} \\
        &= 0 + [2 \cdot 2 \cdot (1 \cdot 1)] + 0 - 0 = 4
    \end{align*}
    \item \textbf{Connections to $a_2$ points ($m=2$):}
    \begin{align*}
        \prescript{2}{}{\alpha}_{2}^{2} &= \left[ 2^0 \cdot {}^2C_0 \cdot (2^{2-2} \cdot {}^{0}C_{2-2}) \right] + \left[ \dots \right]_{\text{i=1}} + \left[ \dots \right]_{\text{i=2}} - {}^0C_0 \\
        &= [1 \cdot 1 \cdot (1 \cdot 1)] + 0 + 0 - 1 = 0
    \end{align*}
\end{itemize}
Total connections for an $a_2$ point = $4 + 0 = \textbf{4}$.

\textbf{3. Contributions: }\\
Now we calculate the CF contribution of each remaining point type. The total number of neighbors in 2D space is $3^2-1=8$.
\begin{itemize}
    \item Contribution of an $a_1$ point: $\prescript{1}{}{\chi}_{1}^{2} = \frac{6}{8}$
    \item Contribution of an $a_2$ point: $\prescript{1}{}{\chi}_{2}^{2} = \frac{4}{8}$
\end{itemize}

\textbf{4. Frequencies: }\\
We calculate the frequency of the remaining points ($a_1, a_2$) in the idealized structure.
\begin{itemize}
    \item Total non-eliminated point types for frequency calculation: ${}^2C_1 + {}^2C_2 = 2 + 1 = 3$.
    \item Frequency of $a_1$ points: $f_1 = \frac{{}^2C_1}{3} = 2/3$
    \item Frequency of $a_2$ points: $f_2 = \frac{{}^2C_2}{3} = 1/3$
\end{itemize}

\textbf{5. Final Calculation: }\\
Finally, we calculate the weighted average.
\begin{equation}
    {}^U\mathcal{CF}_{1}^{2} = (f_1 \cdot \prescript{1}{}{\chi}_{1}^{2}) + (f_2 \cdot \prescript{1}{}{\chi}_{2}^{2}) = \left(\frac{2}{3} \cdot \frac{6}{8}\right) + \left(\frac{1}{3} \cdot \frac{4}{8}\right) = \frac{12}{24} + \frac{4}{24} = \frac{16}{24} = \frac{2}{3} \approx 0.667
\end{equation}

\newpage

\section{Dataset Generation Details and Extra Runs}\label{datarunsextra}

\subsection{Concentric Circle Data (CCD)}
\textbf{Dataset Creation Details:}
The Circle Dataset (CCD) generates data points arranged in concentric circular patterns by sampling angles uniformly around each circle and converting them to Cartesian coordinates. Each point’s location is determined by its radius and center, and independent noise—controlled by a noise rate parameter—is added to both coordinates.

\begin{table}[H]
\centering
\caption{CCD Dataset Parameters, Train-Test Distribution, and Formulas}
\label{tab:CCD_combined}
\begin{tabular}{@{}lp{6.5cm}p{5.5cm}@{}}
\toprule
\textbf{Parameter / Subset / Formula} & \textbf{Value / Class / Formula} & \textbf{Description} \\ \midrule
$\theta_{\text{start}}$ & 0 radians & Start angle \\
$\theta_{\text{end}}$ & $2\pi$ radians & End angle \\
$\theta$ (samples) & 360 samples & Points per circle \\
radius & 3.0 (Circle 1), 4.0 (Circle 2) & Circle radii \\
$x_{\text{center}}$ & 0.0 & X-center \\
$y_{\text{center}}$ & 0.0 & Y-center \\
noise\_rate & 0.5 & Noise magnitude \\
Training & Circle 1 (1): 8,000 (40\%) \newline Circle 2 (2): 8,000 (40\%) & Training data \\
Testing & Circle 1 (1): 2,000 (10\%) \newline Circle 2 (2): 2,000 (10\%) & Test data \\
Point Generation & 
$\begin{pmatrix} x \\ y \end{pmatrix} = 
\begin{pmatrix} r \cos\theta \\ r \sin\theta \end{pmatrix} + 
\begin{pmatrix} x_c \\ y_c \end{pmatrix} + 
\begin{pmatrix} n_x \\ n_y \end{pmatrix}$ & Coordinate formula \\
Noise ($x$) & $n_x = \text{rand}() \cdot \text{noise\_rate} - \text{rand}() \cdot \text{noise\_rate}$ & X-noise \\
Noise ($y$) & $n_y = \text{rand}() \cdot \text{noise\_rate} - \text{rand}() \cdot \text{noise\_rate}$ & Y-noise \\
\bottomrule
\end{tabular}
\vspace{-1mm}
\footnotesize{\textit{Note: rand() uses uniform distribution in $[0, 1)$}}
\end{table}

\noindent
\begin{minipage}{\textwidth}
    \subsubsection{Decision Tree:}

    % --- Plots Side by Side ---
    \centering
    \begin{figure}[H]
    \begin{subfigure}{0.48\textwidth}
        \centering
        \includegraphics[width=\linewidth,trim=0 0 0 30,clip]{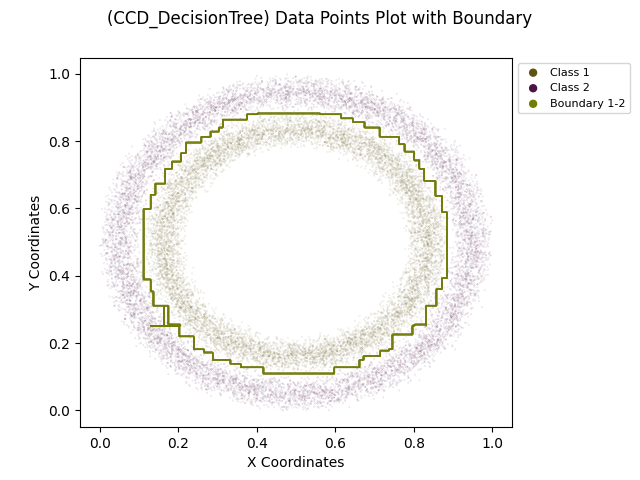}
        \caption{Boundary Plot with Datapoints}
        \label{fig:image2}
    \end{subfigure}\hfill
    \begin{subfigure}{0.48\textwidth}
        \centering
        \includegraphics[width=\linewidth,trim=0 0 0 30,clip]{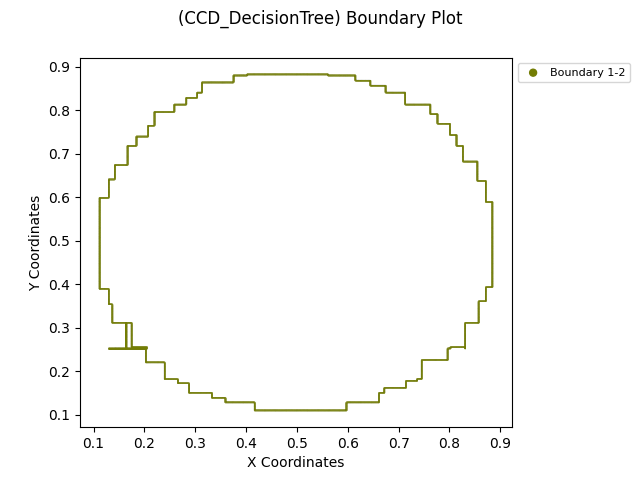}
        \caption{Boundary Plot}
        \label{fig:image3}
    \end{subfigure}

    \caption{Plots of Decision Tree on CCD Dataset corresponding to Table~\ref{tab:CCD_combined}}
    \end{figure}
    \label{fig:four_small_images}

    % --- Tables Side by Side ---
    \vspace{1em}
    \noindent
    \begin{minipage}{0.48\textwidth}
        \centering
        \renewcommand{\arraystretch}{1.5}
        \begin{tabular}{lc}
            \toprule
            Fractal Dimension (Boundary)          & 1.0064  \\
            Connectivity Factor (Boundary)        & 0.2540  \\
            Topological Dimension (${}^L\mathcal{CF}$ Based)      & 1       \\
            Topological Dimension (Weight Based)  & 1       \\
            \bottomrule
        \end{tabular}
    \end{minipage}%
    \hfill
    \begin{minipage}{0.48\textwidth}
        \centering
        \renewcommand{\arraystretch}{1.5}
        \begin{tabular}{lc}
            \toprule
            Weight (Topology 0, Boundary)         & 0.0001 \\
            Weight (Topology 1, Boundary)         & 0.9944 \\
            Weight (Topology 2, Boundary)         & 0.0053 \\
            Fractal Dimension (Object 1)          & 1.7473 \\
            Fractal Dimension (Object 2)          & 1.7120 \\
            \bottomrule
        \end{tabular}
    \end{minipage}

    \captionsetup{skip=2pt}
    \captionof{table}{Boundary characteristics results}
    \label{tab:boundary_characteristics}
\end{minipage}

\noindent
\begin{minipage}{\textwidth}
    \subsubsection{MLP:} % Changed to MLP

    % --- Plots Side by Side ---
    \centering
    \begin{figure}[H]
        \begin{subfigure}{0.48\textwidth}
            \centering
            \includegraphics[width=\linewidth,trim=0 0 0 30,clip]{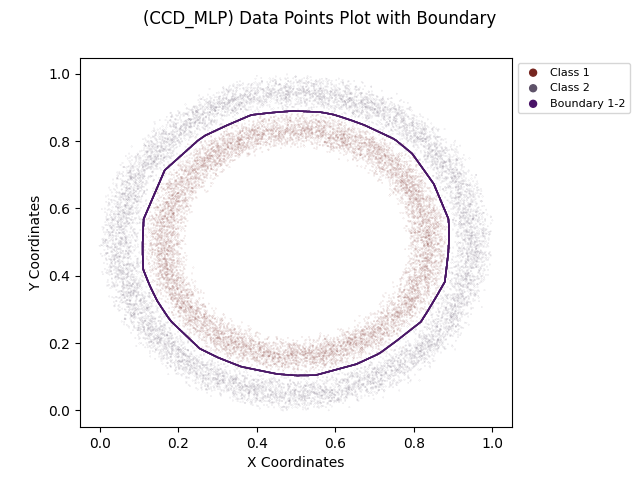}
            \caption{Boundary Plot with Datapoints}
            \label{fig:mlp_image2}
        \end{subfigure}\hfill
        \begin{subfigure}{0.48\textwidth}
            \centering
            \includegraphics[width=\linewidth,trim=0 0 0 30,clip]{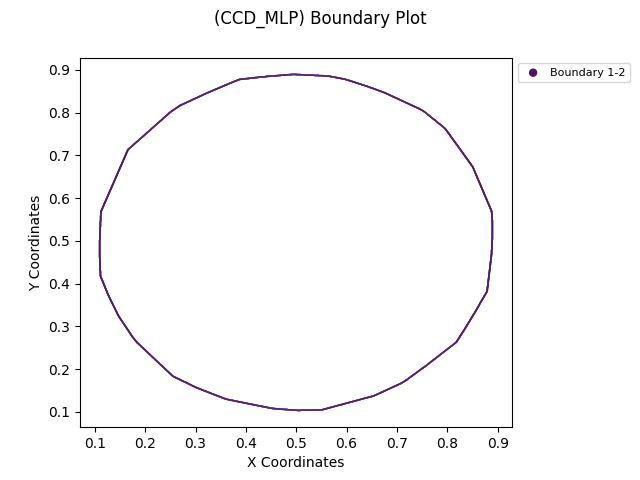}
            \caption{Boundary Plot}
            \label{fig:mlp_image3}
        \end{subfigure}
    
        \caption{Plots of MLP on CCD Dataset corresponding to Table~\ref{tab:CCD_combined}}
    \end{figure}
    \label{fig:mlp_small_images}

    % --- Tables Side by Side ---
    \vspace{1em}
    \noindent
    \begin{minipage}{0.48\textwidth}
        \centering
        \renewcommand{\arraystretch}{1.5}
        \begin{tabular}{lc}
            \toprule
            Fractal Dimension (Boundary)          & 1.0332  \\
            Connectivity Factor (Boundary)        & 0.3949  \\
            Topological Dimension (${}^L\mathcal{CF}$ Based)      & 1       \\
            Topological Dimension (Weight Based)  & 1       \\
            \bottomrule
        \end{tabular}
    \end{minipage}%
    \hfill
    \begin{minipage}{0.48\textwidth}
        \centering
        \renewcommand{\arraystretch}{1.5}
        \begin{tabular}{lc}
            \toprule
            Weight (Topology 0, Boundary)         & 0.0000 \\
            Weight (Topology 1, Boundary)         & 0.8068 \\
            Weight (Topology 2, Boundary)         & 0.1932 \\
            Fractal Dimension (Object 1)          & 1.7473 \\
            Fractal Dimension (Object 2)          & 1.7120 \\
            \bottomrule
        \end{tabular}
    \end{minipage}

    \captionsetup{skip=2pt}
    \captionof{table}{MLP boundary characteristics results}
    \label{tab:mlp_boundary_characteristics}
\end{minipage}

\subsection{Overlapping Concentric Circle Data (OCCD)}
\textbf{Dataset Creation Details:}
The Overlapping Concentric Circle Dataset (OCCD) generates data points arranged in concentric circular patterns with intentional overlap and added noise. Points are sampled by selecting angles uniformly around each circle and converting them to Cartesian coordinates. Each point’s position is determined by the specific radius and center of its circle. Independent noise—controlled by a noise rate parameter—is added to both the $x$ and $y$ coordinates, increasing the variability and overlap between circles.

\begin{table}[H]
\centering
\caption{OCCD Dataset Parameters, Train-Test Distribution, and Formulas}
\label{tab:OCCD_combined}
\begin{tabular}{@{}lp{6.5cm}p{5.5cm}@{}}
\toprule
\textbf{Parameter / Subset / Formula} & \textbf{Value / Class / Formula} & \textbf{Description} \\ \midrule
$\theta_{\text{start}}$ & 0 radians & Start angle \\
$\theta_{\text{end}}$ & $2\pi$ radians & End angle \\
$\theta$ (samples) & 360 samples & Points per circle \\
radius & 3.0 (Circle 1), 3.5 (Circle 2) & Circle radii \\
$x_{\text{center}}$ & 0.0 & X-center \\
$y_{\text{center}}$ & 0.0 & Y-center \\
noise\_rate & 0.7 & Noise magnitude \\
Training & Circle 1 (1): 8,000 (40\%) \newline Circle 2 (2): 8,000 (40\%) & Training data \\
Testing & Circle 1 (1): 2,000 (10\%) \newline Circle 2 (2): 2,000 (10\%) & Test data \\
Point Generation & 
$\begin{pmatrix} x \\ y \end{pmatrix} = 
\begin{pmatrix} r \cos\theta \\ r \sin\theta \end{pmatrix} + 
\begin{pmatrix} x_c \\ y_c \end{pmatrix} + 
\begin{pmatrix} n_x \\ n_y \end{pmatrix}$ & Coordinate formula \\
Noise ($x$) & $n_x = \text{rand}() \cdot \text{noise\_rate} - \text{rand}() \cdot \text{noise\_rate}$ & X-noise \\
Noise ($y$) & $n_y = \text{rand}() \cdot \text{noise\_rate} - \text{rand}() \cdot \text{noise\_rate}$ & Y-noise \\
\bottomrule
\end{tabular}
\vspace{-1mm}
\footnotesize{\textit{Note: rand() uses uniform distribution in $[0, 1)$}}
\end{table}

\noindent
\begin{minipage}{\textwidth}
    \subsubsection{KNN:}
    % --- Plots Side by Side ---
    \begin{figure}[H]
        \centering
        \begin{subfigure}{0.48\textwidth}
            \centering
            \includegraphics[width=\linewidth,trim=0 0 0 30,clip]{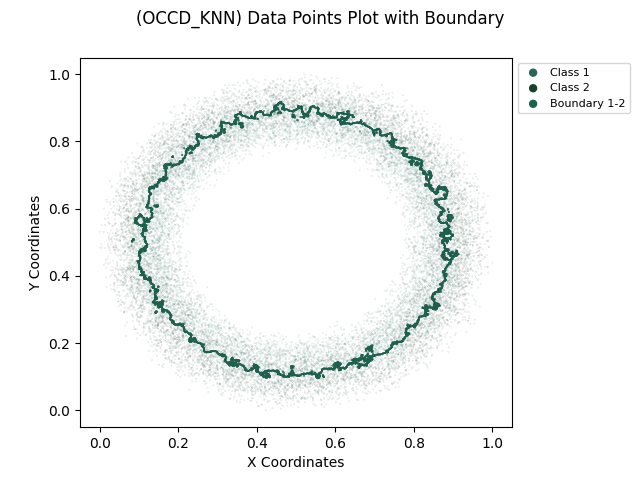} % Updated path
            \caption{Boundary Plot with Datapoints}
            \label{fig:knn_image2}
        \end{subfigure}\hfill
        \begin{subfigure}{0.48\textwidth}
            \centering
            \includegraphics[width=\linewidth,trim=0 0 0 30,clip]{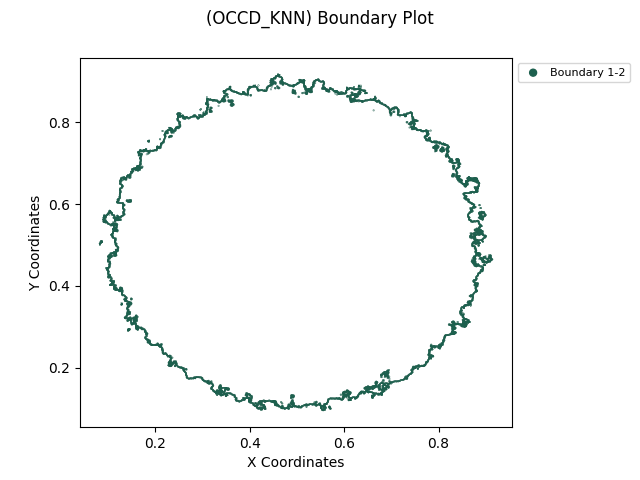} % Updated path
            \caption{Boundary Plot}
            \label{fig:knn_image3}
        \end{subfigure}
    
        \caption{Plots of KNN on CCD Dataset corresponding to Table~\ref{tab:OCCD_combined}}
    \end{figure}
    \label{fig:knn_small_images} % Updated label

    % --- Tables Side by Side ---
    \vspace{1em}
    \noindent
    \begin{minipage}{0.48\textwidth}
        \centering
        \renewcommand{\arraystretch}{1.5}
        \begin{tabular}{lc}
            \toprule
            Fractal Dimension (Boundary)          & 1.3106  \\
            Connectivity Factor (Boundary)        & 0.4254  \\
            Topological Dimension (${}^L\mathcal{CF}$ Based)      & 1       \\
            Topological Dimension (Weight Based)  & 1       \\
            \bottomrule
        \end{tabular}
    \end{minipage}%
    \hfill
    \begin{minipage}{0.48\textwidth}
        \centering
        \renewcommand{\arraystretch}{1.5}
        \begin{tabular}{lc}
            \toprule
            Weight (Topology 0, Boundary)         & 0.0106 \\
            Weight (Topology 1, Boundary)         & 0.7518 \\
            Weight (Topology 2, Boundary)         & 0.2374 \\
            Fractal Dimension (Object 1)          & 1.7625 \\
            Fractal Dimension (Object 2)          & 1.7662 \\
            \bottomrule
        \end{tabular}
    \end{minipage}

    \captionsetup{skip=2pt}
    \captionof{table}{KNN boundary characteristics results} % Updated caption
    \label{tab:knn_boundary_characteristics} % Updated label
\end{minipage}

\noindent
\begin{minipage}{\textwidth}
    \subsubsection{MLP:} % Changed to MLP

    % --- Plots Side by Side ---
    \begin{figure}[H]
        \centering
        \begin{subfigure}{0.48\textwidth}
            \centering
            \includegraphics[width=\linewidth,trim=0 0 0 30,clip]{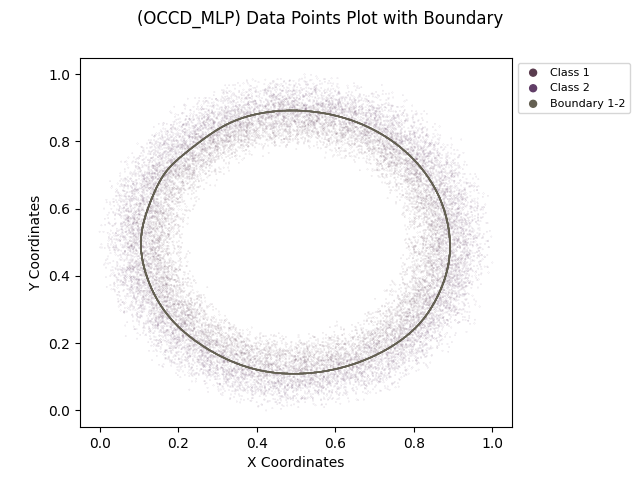}
            \caption{Boundary Plot with Datapoints}
            \label{fig:mlp_image2}
        \end{subfigure}\hfill
        \begin{subfigure}{0.48\textwidth}
            \centering
            \includegraphics[width=\linewidth,trim=0 0 0 30,clip]{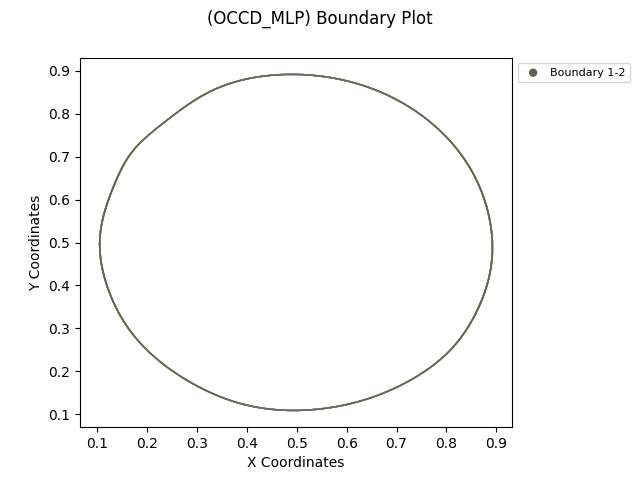}
            \caption{Boundary Plot}
            \label{fig:mlp_image3}
        \end{subfigure}
    
        \caption{Plots of MLP on OCCD Dataset corresponding to Table~\ref{tab:OCCD_combined}}
    \end{figure}
    \label{fig:mlp_small_images}

    % --- Tables Side by Side ---
    \vspace{1em}
    \noindent
    \begin{minipage}{0.48\textwidth}
        \centering
        \renewcommand{\arraystretch}{1.5}
        \begin{tabular}{lc}
            \toprule
            Fractal Dimension (Boundary)          & 1.0258  \\
            Connectivity Factor (Boundary)        & 0.3985  \\
            Topological Dimension (${}^L\mathcal{CF}$ Based)      & 1       \\
            Topological Dimension (Weight Based)  & 1       \\
            \bottomrule
        \end{tabular}
    \end{minipage}%
    \hfill
    \begin{minipage}{0.48\textwidth}
        \centering
        \renewcommand{\arraystretch}{1.5}
        \begin{tabular}{lc}
            \toprule
            Weight (Topology 0, Boundary)         & 0.0000 \\
            Weight (Topology 1, Boundary)         & 0.8018 \\
            Weight (Topology 2, Boundary)         & 0.1981 \\
            Fractal Dimension (Object 1)          & 1.7625 \\
            Fractal Dimension (Object 2)          & 1.7662 \\
            \bottomrule
        \end{tabular}
    \end{minipage}

    \captionsetup{skip=2pt}
    \captionof{table}{MLP boundary characteristics results}
    \label{tab:mlp_boundary_characteristics}
\end{minipage}

\subsection{Sinusoidal Curve Data (SD):}
\textbf{Dataset Creation Details:}
The Sinusoidal Curve Dataset (SD) with Mean and Phase Difference generates data points arranged along sinusoidal curves with identical amplitude but distinct phase differences and vertical offsets. Each curve has a unique class identity, a specified phase difference, a y-axis offset, and added noise. The sinusoidal curve is represented by the equation: $y = \text{amplitude} \cdot \sin(x + \text{phase\_difference}) + y_{\text{center}} + \text{noise}_y$
\begin{table}[H]
\centering
\caption{SD Dataset Parameters, Train-Test Distribution, and Formulas}
\label{tab:SD_combined}
\begin{tabular}{@{}lp{6.5cm}p{5.5cm}@{}}
\toprule
\textbf{Parameter / Subset / Formula} & \textbf{Value / Class / Formula} & \textbf{Description} \\ \midrule
$x_{\text{min}}$ & 0.0 & Start of $x$-sampling interval \\
$x_{\text{max}}$ & $2\pi$ & End of $x$-sampling interval \\
$x$ distribution & evenly spaced & Horizontal distribution along sinusoidal wave \\
amplitude & 1.0 (both curves) & Peak vertical distance from center line \\
phase\_difference & 0.0 radians (Curve 1), $\pi$ radians (Curve 2) & Horizontal shift of sinusoidal wave \\
$y_{\text{center}}$ & 0.0 (Curve 1), 0.5 (Curve 2) & Vertical offset along y-axis \\
noise\_rate & 0.5 & Level of noise around each point's mean position \\
Training & Curve 1: 80\%, Curve 2: 80\% & Training samples per curve \\
Testing & Curve 1: 20\%, Curve 2: 20\% & Test samples per curve \\
Point Generation & 
$y = \text{amplitude} \cdot \sin(x + \text{phase\_difference}) + y_{\text{center}} + \text{noise}_y$ & Sinusoidal curve equation \\
Noise ($y$) & $\text{noise}_y = \text{rand}() \cdot \text{noise\_rate} - \frac{\text{noise\_rate}}{2}$ & Random noise added to $y$-coordinate \\
\bottomrule
\end{tabular}
\vspace{-1mm}
\footnotesize{\textit{Note: rand() uses a uniform distribution in $[0, 1)$.}}
\end{table}

\vspace{-2mm}

\noindent
\begin{minipage}{\textwidth}
    \subsubsection{KNN:}
    % --- Plots Side by Side ---
    \centering
    \begin{figure}[H]
        \begin{subfigure}{0.48\textwidth}
            \centering
            \includegraphics[width=\linewidth,trim=0 0 0 30,clip]{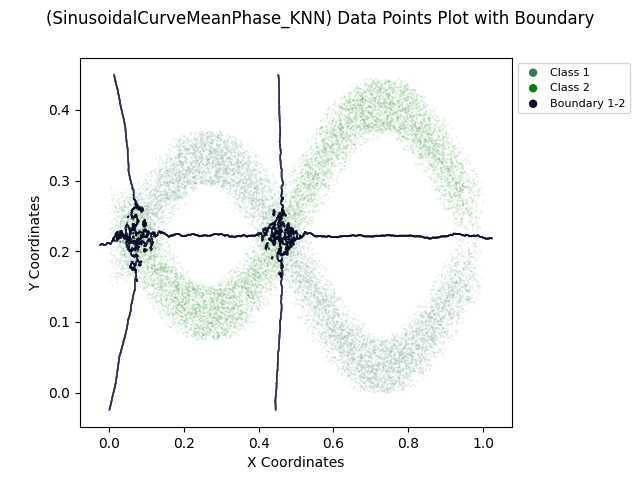} % Updated path
            \caption{Boundary Plot with Datapoints}
            \label{fig:knn_image2}
        \end{subfigure}\hfill
        \begin{subfigure}{0.48\textwidth}
            \centering
            \includegraphics[width=\linewidth,trim=0 0 0 30,clip]{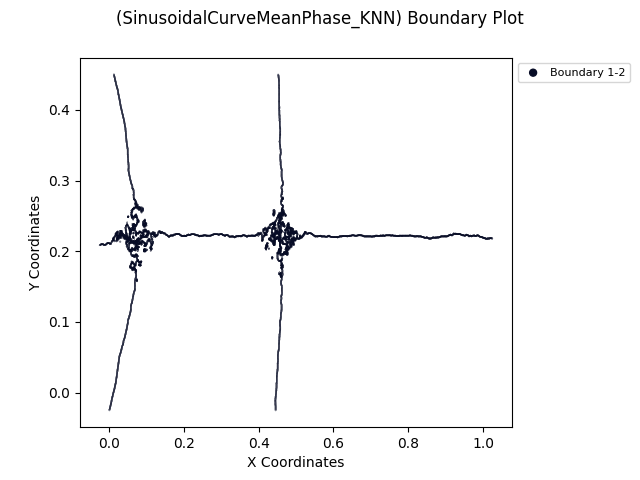} % Updated path
            \caption{Boundary Plot}
            \label{fig:knn_image3}
        \end{subfigure}
    
        \caption{Plots of KNN on SD Dataset corresponding to Table~\ref{tab:SD_combined}}
    \end{figure}
    \label{fig:knn_small_images} % Updated label

    % --- Tables Side by Side ---
    \vspace{1em}
    \noindent
    \begin{minipage}{0.48\textwidth}
        \centering
        \renewcommand{\arraystretch}{1.5}
        \begin{tabular}{lc}
            \toprule
            Fractal Dimension (Boundary)          & 1.2515  \\
            Connectivity Factor (Boundary)        & 0.4076  \\
            Topological Dimension (${}^L\mathcal{CF}$ Based)      & 1       \\
            Topological Dimension (Weight Based)  & 1       \\
            \bottomrule
        \end{tabular}
    \end{minipage}%
    \hfill
    \begin{minipage}{0.48\textwidth}
        \centering
        \renewcommand{\arraystretch}{1.5}
        \begin{tabular}{lc}
            \toprule
            Weight (Topology 0, Boundary)         & 0.0087 \\
            Weight (Topology 1, Boundary)         & 0.7780 \\
            Weight (Topology 2, Boundary)         & 0.2131 \\
            Fractal Dimension (Object 1)          & 1.7631 \\
            Fractal Dimension (Object 2)          & 1.7348 \\
            \bottomrule
        \end{tabular}
    \end{minipage}

    \captionsetup{skip=2pt}
    \captionof{table}{KNN boundary characteristics results} 
    \label{tab:knn_boundary_characteristics} % Updated label
\end{minipage}

\newpage

\subsection{Barnsley Fern Data (BF)}
Dataset Creation Details:
The Barnsley Fern Dataset (BF) generates data points that collectively form the fractal structure known as the Barnsley fern. Each point is produced iteratively using the Chaos Game method, which applies one of four affine transformations at each step. The transformation is chosen probabilistically, based on predefined probabilities, and the process starts from the initial point 
(
0.0
,
0.0
)
(0.0,0.0). The dataset includes both fern points and randomly sampled non-fern points within the fern's bounding box, resulting in two classes.

\begin{table}[H]
\centering
\caption{BF Dataset Parameters, Train-Test Distribution, and Formulas}
\label{tab:BF_combined}
\begin{tabular}{@{}lp{6.5cm}p{5.5cm}@{}}
\toprule
\textbf{Parameter / Subset / Formula} & \textbf{Value / Class / Formula} & \textbf{Description} \\ \midrule
Initial Point & $(0.0, 0.0)$ & Starting coordinates for the iterative process \\
Number of Fern Points & 5,000,000 & Total points generated for the fern structure (Class 1) \\
Number of Non-Fern Points & 5,000,000 & Random points within the fern's bounding box (Class 0) \\
Affine Transformations & Four, with coefficients: \newline
\begin{tabular}{l}
1: $a=0.00$, $b=0.00$, $c=0.00$, $d=0.16$, $e=0.00$, $f=0.00$ \\
2: $a=0.85$, $b=0.04$, $c=-0.04$, $d=0.85$, $e=0.00$, $f=1.6$ \\
3: $a=0.20$, $b=-0.26$, $c=0.23$, $d=0.22$, $e=0.00$, $f=1.6$ \\
4: $a=-0.15$, $b=0.28$, $c=0.26$, $d=0.24$, $e=0.00$, $f=0.44$ \\
\end{tabular}
& shape and branching of the fern \\
Transformation Probabilities & $p_1=0.01$, $p_2=0.85$, $p_3=0.07$, $p_4=0.07$ & Probability of selecting each transformation \\
Training & Fern (1): 4,000,000 samples (40\%) \newline Non-Fern (0): 4,000,000 samples (40\%) & Number and fraction of training samples for each class \\
Testing & Fern (1): 1,000,000 samples (10\%) \newline Non-Fern (0): 1,000,000 samples (10\%) & Number and fraction of test samples for each class \\
Point Generation & 
$\displaystyle
\begin{pmatrix} x_{n+1} \\ y_{n+1} \end{pmatrix} =
\begin{pmatrix} a & b \\ c & d \end{pmatrix}
\begin{pmatrix} x_n \\ y_n \end{pmatrix} +
\begin{pmatrix} e \\ f \end{pmatrix}
$ & Affine transformation applied at each iteration \\
\bottomrule
\end{tabular}
\vspace{-1mm}
\footnotesize{\textit{Note: At each iteration, a random number in $[0,1)$ determines which transformation is applied, according to the specified probabilities. The dataset comprises two classes: Class 1 (fern points) and Class 0 (random non-fern points).}}
\end{table}

\vspace{-2mm}

\noindent
\begin{minipage}{\textwidth}
    \subsubsection{Decision Tree:}

    % --- Plots Side by Side ---
    \begin{figure}[H]
        \centering
        \begin{subfigure}{0.48\textwidth}
            \centering
            \includegraphics[width=\linewidth,trim=0 0 0 30,clip]{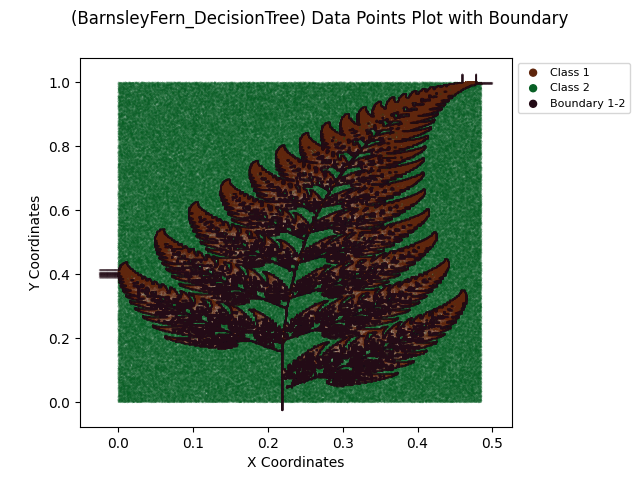}
            \caption{Boundary Plot with Datapoints}
            \label{fig:bf_image2}
        \end{subfigure}\hfill
        \begin{subfigure}{0.48\textwidth}
            \centering
            \includegraphics[width=\linewidth,trim=0 0 0 30,clip]{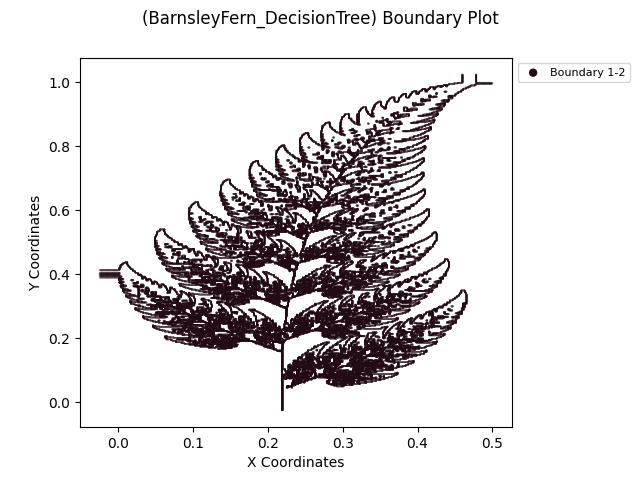}
            \caption{Boundary Plot}
            \label{fig:bf_image3}
        \end{subfigure}
    
        \caption{Plots of Decision Tree on BF Dataset corresponding to Table~\ref{tab:BF_combined}}
    \end{figure}
    \label{fig:bf_small_images}

    % --- Tables Side by Side ---
    \vspace{1em}
    \noindent
    \begin{minipage}{0.48\textwidth}
        \centering
        \renewcommand{\arraystretch}{1.5}
        \begin{tabular}{lc}
            \toprule
            Fractal Dimension (Boundary)          & 1.7415  \\
            Connectivity Factor (Boundary)        & 00.3289  \\
            Topological Dimension (${}^L\mathcal{CF}$ Based)      & 1       \\
            Topological Dimension (Weight Based)  & 1       \\
            \bottomrule
        \end{tabular}
    \end{minipage}%
    \hfill
    \begin{minipage}{0.48\textwidth}
        \centering
        \renewcommand{\arraystretch}{1.5}
        \begin{tabular}{lc}
            \toprule
            Weight (Topology 0, Boundary)         & 0.1450 \\
            Weight (Topology 1, Boundary)         & 0.5764 \\
            Weight (Topology 2, Boundary)         & 0.2785 \\
            Fractal Dimension (Object 1)          & 1.8340 \\
            \bottomrule
        \end{tabular}
    \end{minipage}

    \captionsetup{skip=2pt}
    \captionof{table}{Boundary characteristics results}
    \label{tab:boundary_characteristics}
\end{minipage}

\newpage
\subsection{Sierpinski Carpet Data (SC)}

\textbf{Dataset Creation Details:}
The Sierpinski Carpet Dataset (SC) is generated using the Chaos Game method, which employs iterative affine transformations. Each new point is computed from the previous one using the following equation:

\begin{table}[H]
\centering
\caption{SCD Dataset Parameters, Train-Test Distribution, and Formulas}
\label{tab:SCD_combined}
\begin{tabular}{@{}lp{6.5cm}p{5.5cm}@{}}
\toprule
\textbf{Parameter / Subset / Formula} & \textbf{Value / Class / Formula} & \textbf{Description} \\ \midrule
Initial Point & $(0.5, 0.5)$ & Starting coordinates for the iterative process \\
Number of Carpet Points & 1,000,000 & Total points generated for the carpet structure (Class 1) \\
Number of Non-Carpet Points & 1,000,000 & Random points within the carpet's bounding box (Class 0) \\
Affine Transformations & Eight, with coefficients: \newline
\begin{tabular}{l}
1: $a=0.333$, $b=0.0$, $c=0.0$, $d=0.333$, $e=0.0$, $f=0.0$ \\
2: $a=0.333$, $b=0.0$, $c=0.0$, $d=0.333$, $e=0.333$, $f=0.0$ \\
3: $a=0.333$, $b=0.0$, $c=0.0$, $d=0.333$, $e=0.666$, $f=0.0$ \\
4: $a=0.333$, $b=0.0$, $c=0.0$, $d=0.333$, $e=0.0$, $f=0.333$ \\
5: $a=0.333$, $b=0.0$, $c=0.0$, $d=0.333$, $e=0.666$, $f=0.333$ \\
6: $a=0.333$, $b=0.0$, $c=0.0$, $d=0.333$, $e=0.0$, $f=0.666$ \\
7: $a=0.333$, $b=0.0$, $c=0.0$, $d=0.333$, $e=0.333$, $f=0.666$ \\
8: $a=0.333$, $b=0.0$, $c=0.0$, $d=0.333$, $e=0.666$, $f=0.666$ \\
\end{tabular}
& Recursive structure of the carpet \\
Transformation Probabilities & $p_1$--$p_8 = 0.125$ each & Probability of selecting each transformation \\
Training & Carpet (1): 800,000 samples (40\%) \newline Non-Carpet (0): 800,000 samples (40\%) & Number and fraction of training samples for each class \\
Testing & Carpet (1): 200,000 samples (10\%) \newline Non-Carpet (0): 200,000 samples (10\%) & Number and fraction of test samples for each class \\
Point Generation & 
$\displaystyle
\begin{pmatrix} x_{n+1} \\ y_{n+1} \end{pmatrix} =
\begin{pmatrix} a & b \\ c & d \end{pmatrix}
\begin{pmatrix} x_n \\ y_n \end{pmatrix} +
\begin{pmatrix} e \\ f \end{pmatrix}
$ & Affine transformation applied at each iteration \\
\bottomrule
\end{tabular}
\vspace{-1mm}
\footnotesize{\textit{Note: At each iteration, a random number in $[0,1)$ determines which transformation is applied, according to the specified probabilities. The dataset comprises two classes: Class 1 (carpet points) and Class 0 (random non-carpet points).}}
\end{table}

\vspace{-2mm}

\noindent
\begin{minipage}{\textwidth}
    \subsubsection{KNN:}
    % --- Plots Side by Side ---
    \begin{figure}[H]
        \centering
        \begin{subfigure}{0.48\textwidth}
            \centering
            \includegraphics[width=\linewidth,trim=0 0 0 30,clip]{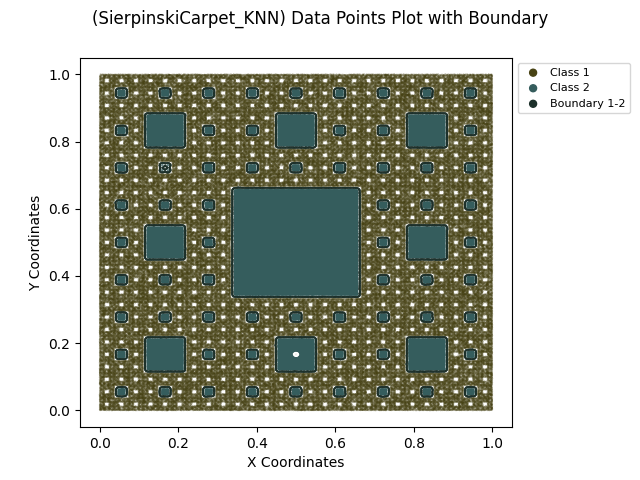}
            \caption{Boundary Plot with Datapoints}
            \label{fig:knn_sc_image2}
        \end{subfigure}\hfill
        \begin{subfigure}{0.48\textwidth}
            \centering
            \includegraphics[width=\linewidth,trim=0 0 0 30,clip]{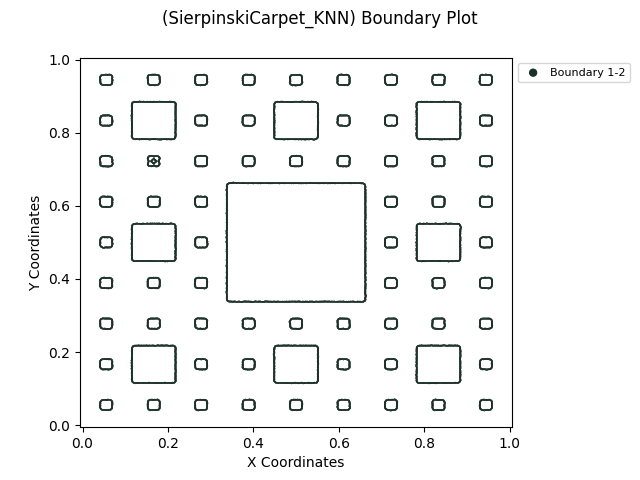}
            \caption{Boundary Plot}
            \label{fig:knn_sc_image3}
        \end{subfigure}
    
        \caption{Plots of KNN on SCD Dataset corresponding to Table~\ref{tab:SCD_combined}}
    \end{figure}
    \label{fig:knn_sc_small_images}

    % --- Tables Side by Side ---
    \vspace{1em}
    \noindent
    \begin{minipage}{0.48\textwidth}
        \centering
        \renewcommand{\arraystretch}{1.5}
        \begin{tabular}{lc}
            \toprule
            Fractal Dimension (Boundary)          & 1.0768  \\
            Connectivity Factor (Boundary)        & 0.3962  \\
            Topological Dimension (${}^L\mathcal{CF}$ Based)      & 1       \\
            Topological Dimension (Weight Based)  & 1       \\
            \bottomrule
        \end{tabular}
    \end{minipage}%
    \hfill
    \begin{minipage}{0.48\textwidth}
        \centering
        \renewcommand{\arraystretch}{1.5}
        \begin{tabular}{lc}
            \toprule
            Weight (Topology 0, Boundary)         & 0.0000 \\
            Weight (Topology 1, Boundary)         & 0.8049 \\
            Weight (Topology 2, Boundary)         & 0.1950 \\
            Fractal Dimension (Object 1)          & 1.8967 \\
            \bottomrule
        \end{tabular}
    \end{minipage}

    \captionsetup{skip=2pt}
    \captionof{table}{KNN boundary characteristics results}
    \label{tab:knn_sc_boundary_characteristics}
\end{minipage}

\newpage
\subsection{Sierpinski Triangle (ST)}
The Sierpinski Triangle dataset is generated using the Chaos Game method, which employs iterative affine transformations.

\begin{table}[H]
\centering
\caption{STD Dataset Parameters, Train-Test Distribution, and Formulas}
\label{tab:STD_combined}
\begin{tabular}{@{}lp{6.5cm}p{5.5cm}@{}}
\toprule
\textbf{Parameter / Subset / Formula} & \textbf{Value / Class / Formula} & \textbf{Description} \\ \midrule
Initial Point & $(0.0, 0.0)$ & Starting coordinates for the iterative process \\
Number of Triangle Points & 1,000,000 & Total points generated for the triangle structure (Class 1) \\
Number of Non-Triangle Points & 1,000,000 & Random points within the triangle's bounding box (Class 0) \\
Affine Transformations & Three, with coefficients: \newline
\begin{tabular}{l}
\\1: $a=0.5$, $b=0.0$, $c=0.0$, $d=0.5$, $e=0.0$, $f=0.0$ \\
2: $a=0.5$, $b=0.0$, $c=0.0$, $d=0.5$, $e=0.5$, $f=0.0$ \\
3: $a=0.5$, $b=0.0$, $c=0.0$, $d=0.5$, $e=0.25$, $f=0.433$ \\
\end{tabular}
& Defines the recursive structure of the triangle\\
Transformation Probabilities & $p_1 = p_2 = p_3 = 0.333$ & Probability of selecting each transformation \\
Training & Triangle (1): 800,000 samples (40\%) \newline Non-Triangle (0): 800,000 samples (40\%) & Number and fraction of training samples for each class \\
Testing & Triangle (1): 200,000 samples (10\%) \newline Non-Triangle (0): 200,000 samples (10\%) & Number and fraction of test samples for each class \\
Point Generation & 
$\displaystyle
\begin{pmatrix} x_{n+1} \\ y_{n+1} \end{pmatrix} =
\begin{pmatrix} a & b \\ c & d \end{pmatrix}
\begin{pmatrix} x_n \\ y_n \end{pmatrix} +
\begin{pmatrix} e \\ f \end{pmatrix}
$ & Affine transformation applied at each iteration \\
\bottomrule
\end{tabular}
\vspace{-1mm}
\footnotesize{\textit{Note: At each iteration, a random number in $[0,1)$ determines which transformation is applied, according to the specified probabilities. The dataset comprises two classes: Class 1 (triangle points) and Class 0 (random non-triangle points).}}
\end{table}

\vspace{-2mm}

\noindent
\begin{minipage}{\textwidth}
    \subsubsection{Decision Tree:}

    % --- Plots Side by Side ---
    \begin{figure}[H]
        \centering
        \begin{subfigure}{0.48\textwidth}
            \centering
            \includegraphics[width=\linewidth,trim=0 0 0 30,clip]{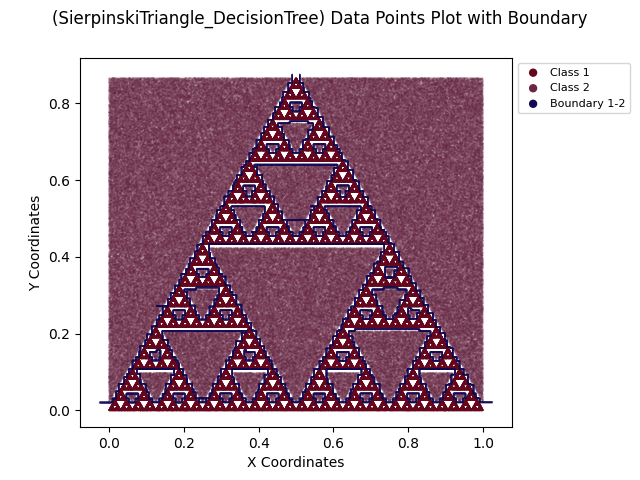}
            \caption{Boundary Plot with Datapoints}
            \label{fig:st_image2}
        \end{subfigure}\hfill
        \begin{subfigure}{0.48\textwidth}
            \centering
            \includegraphics[width=\linewidth,trim=0 0 0 30,clip]{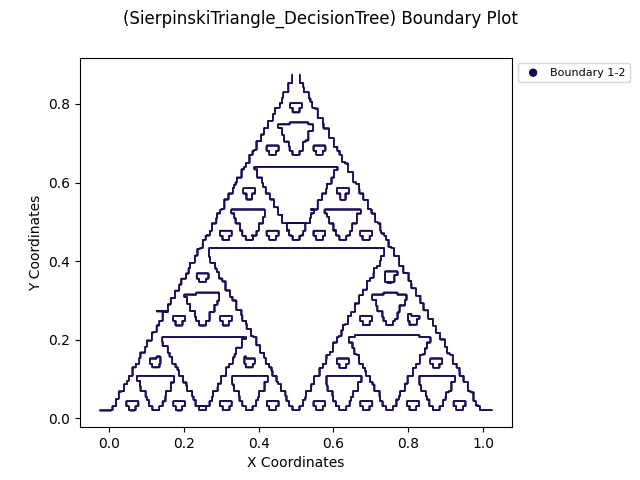}
            \caption{Boundary Plot}
            \label{fig:st_image3}
        \end{subfigure}
    
        \caption{Plots of Decision Tree on STD Dataset corresponding to Table~\ref{tab:STD_combined}}
    \end{figure}
    \label{fig:st_small_images}

    % --- Tables Side by Side ---
    \vspace{1em}
    \noindent
    \begin{minipage}{0.48\textwidth}
        \centering
        \renewcommand{\arraystretch}{1.5}
        \begin{tabular}{lc}
            \toprule
            Fractal Dimension (Boundary)          & 1.0176  \\
            Connectivity Factor (Boundary)        & 0.2599  \\
            Topological Dimension (${}^L\mathcal{CF}$ Based)      & 1       \\
            Topological Dimension (Weight Based)  & 1       \\
            \bottomrule
        \end{tabular}
    \end{minipage}%
    \hfill
    \begin{minipage}{0.48\textwidth}
        \centering
        \renewcommand{\arraystretch}{1.5}
        \begin{tabular}{lc}
            \toprule
            Weight (Topology 0, Boundary)         & 0.0001 \\
            Weight (Topology 1, Boundary)         & 0.9864 \\
            Weight (Topology 2, Boundary)         & 0.0133 \\
            Fractal Dimension (Object 1)          & 1.6232 \\
            \bottomrule
        \end{tabular}
    \end{minipage}

    \captionsetup{skip=2pt}
    \captionof{table}{Boundary characteristics results}
    \label{tab:boundary_characteristics}
\end{minipage}

\noindent
\begin{minipage}{\textwidth}
    \subsubsection{KNN:}
    % --- Plots Side by Side ---
    \begin{figure}[H]
        \centering
        \begin{subfigure}{0.48\textwidth}
            \centering
            \includegraphics[width=\linewidth,trim=0 0 0 30,clip]{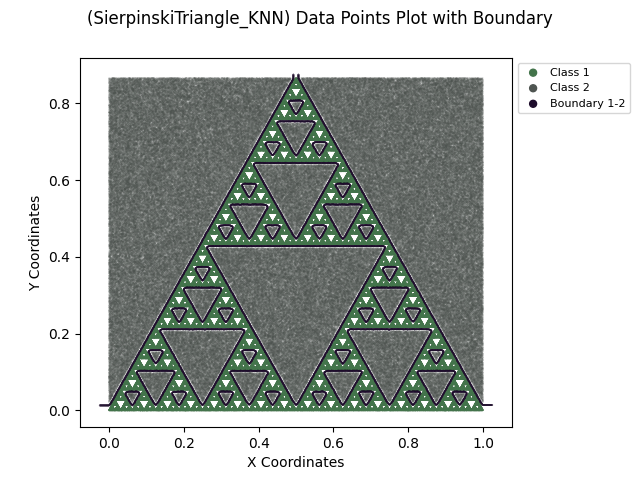}
            \caption{Boundary Plot with Datapoints}
            \label{fig:knn_st_image2}
        \end{subfigure}\hfill
        \begin{subfigure}{0.48\textwidth}
            \centering
            \includegraphics[width=\linewidth,trim=0 0 0 30,clip]{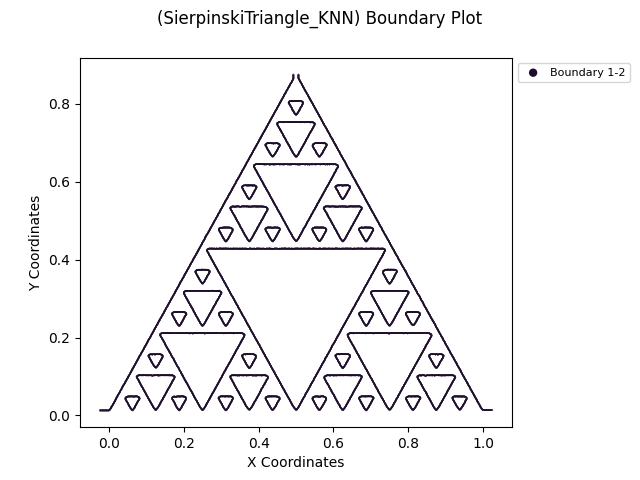}
            \caption{Boundary Plot}
            \label{fig:knn_st_image3}
        \end{subfigure}
    
        \caption{Plots of KNN on STD Dataset corresponding to Table~\ref{tab:STD_combined}}
    \end{figure}
    \label{fig:knn_st_small_images}

    % --- Tables Side by Side ---
    \vspace{1em}
    \noindent
    \begin{minipage}{0.48\textwidth}
        \centering
        \renewcommand{\arraystretch}{1.5}
        \begin{tabular}{lc}
            \toprule
            Fractal Dimension (Boundary)          & 1.0292  \\
            Connectivity Factor (Boundary)        & 0.4222  \\
            Topological Dimension (${}^L\mathcal{CF}$ Based)      & 1       \\
            Topological Dimension (Weight Based)  & 1       \\
            \bottomrule
        \end{tabular}
    \end{minipage}%
    \hfill
    \begin{minipage}{0.48\textwidth}
        \centering
        \renewcommand{\arraystretch}{1.5}
        \begin{tabular}{lc}
            \toprule
            Weight (Topology 0, Boundary)         & 6.3985 \\
            Weight (Topology 1, Boundary)         & 0.7702 \\
            Weight (Topology 2, Boundary)         & 0.2297 \\
            Fractal Dimension (Object 1)          & 1.6232 \\
            \bottomrule
        \end{tabular}
    \end{minipage}

    \captionsetup{skip=2pt}
    \captionof{table}{KNN boundary characteristics results}
    \label{tab:knn_st_boundary_characteristics}
\end{minipage}

\newpage

\section{Discussion on the LMU Bounds}\label{lmudiscussion}

\begin{figure}[htbp]
    \centering
    % Row 1: 2 images
    \begin{subfigure}{0.45\textwidth}
        \includegraphics[width=\linewidth]{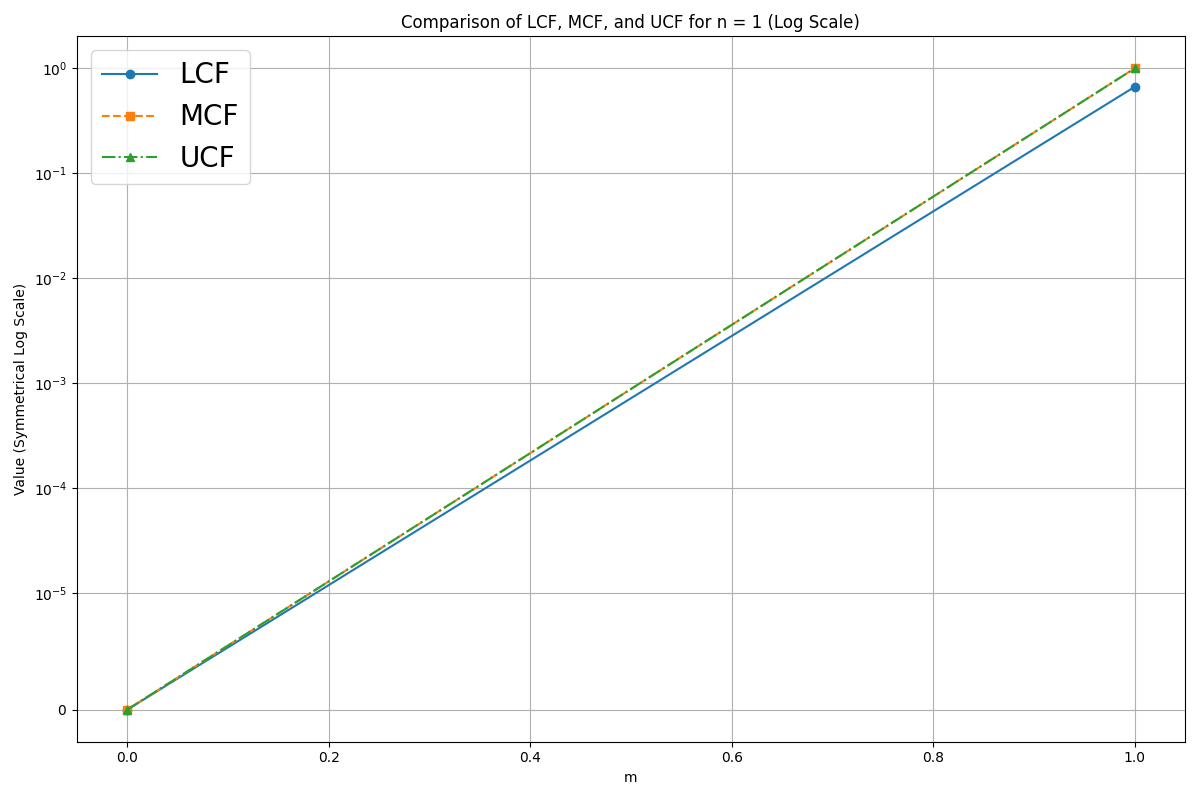}
        \caption{}
    \end{subfigure}\hfill
    \begin{subfigure}{0.45\textwidth}
        \includegraphics[width=\linewidth]{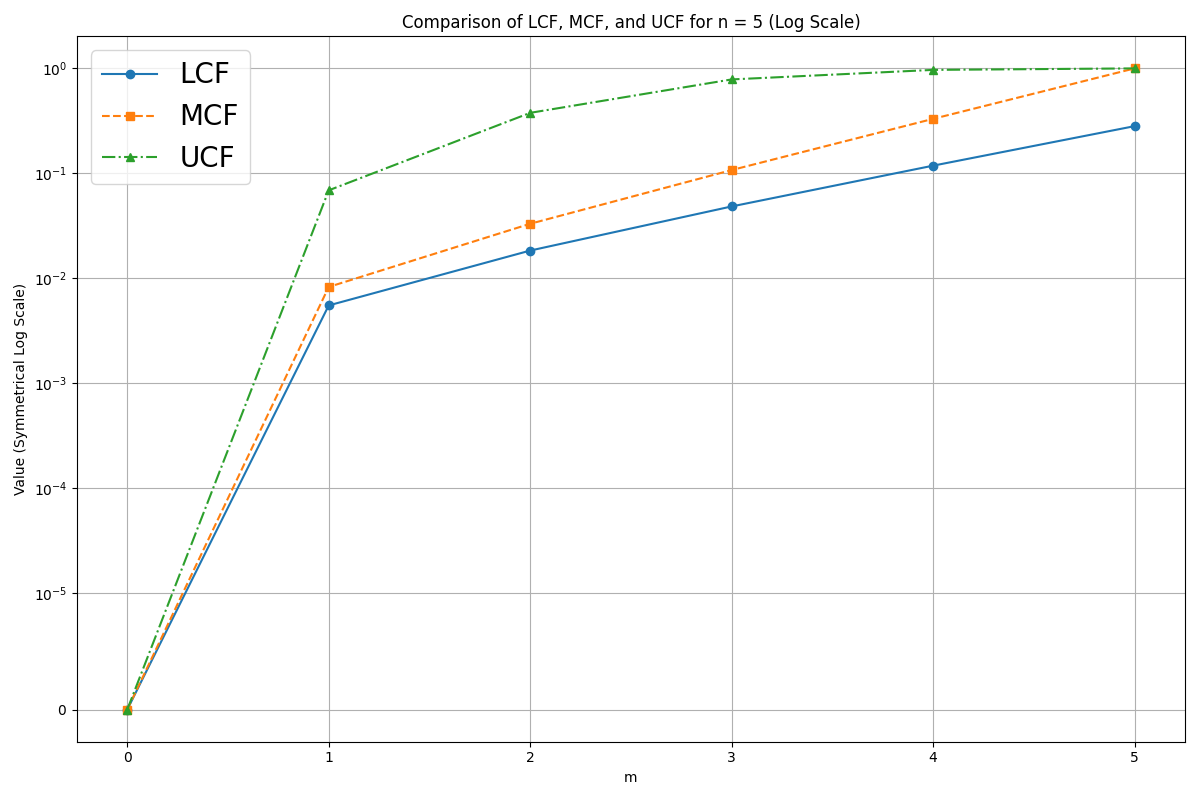}
        \caption{}
    \end{subfigure}

    \vspace{1em}

    % Row 2: 2 images
    \begin{subfigure}{0.45\textwidth}
        \includegraphics[width=\linewidth]{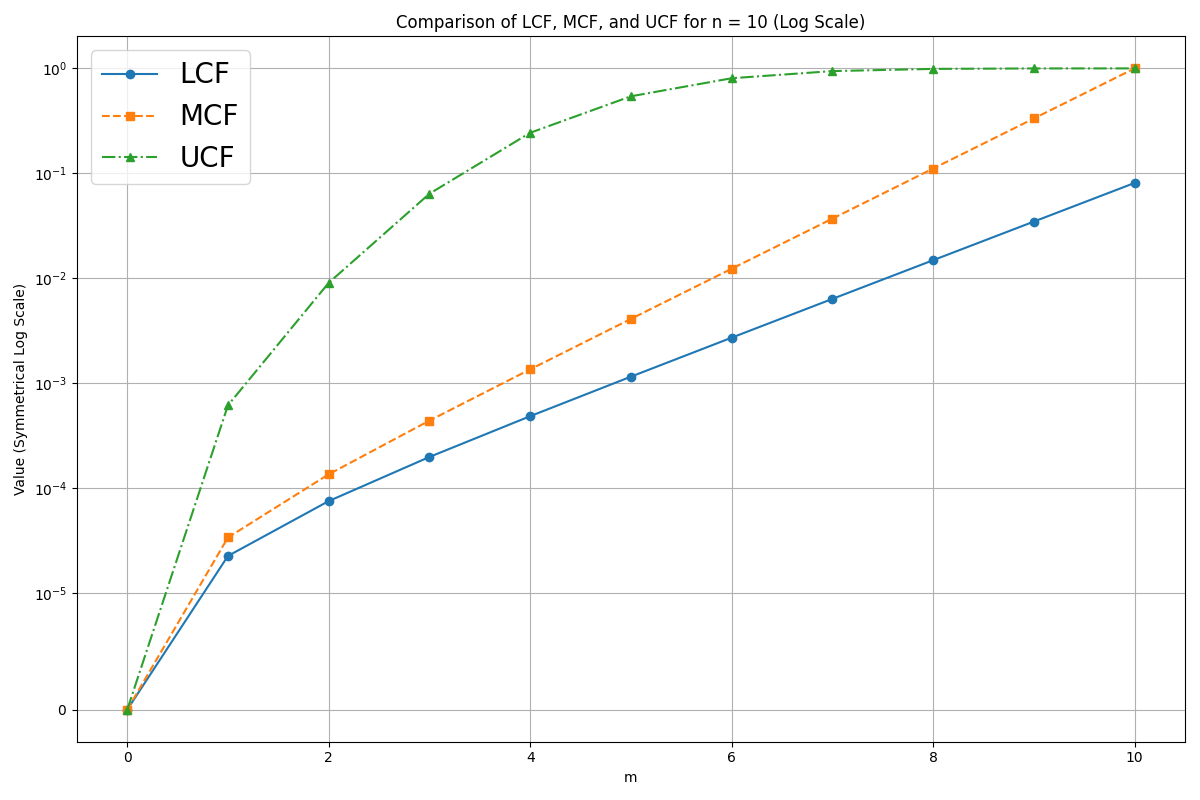}
        \caption{}
    \end{subfigure}\hfill
    \begin{subfigure}{0.45\textwidth}
        \includegraphics[width=\linewidth]{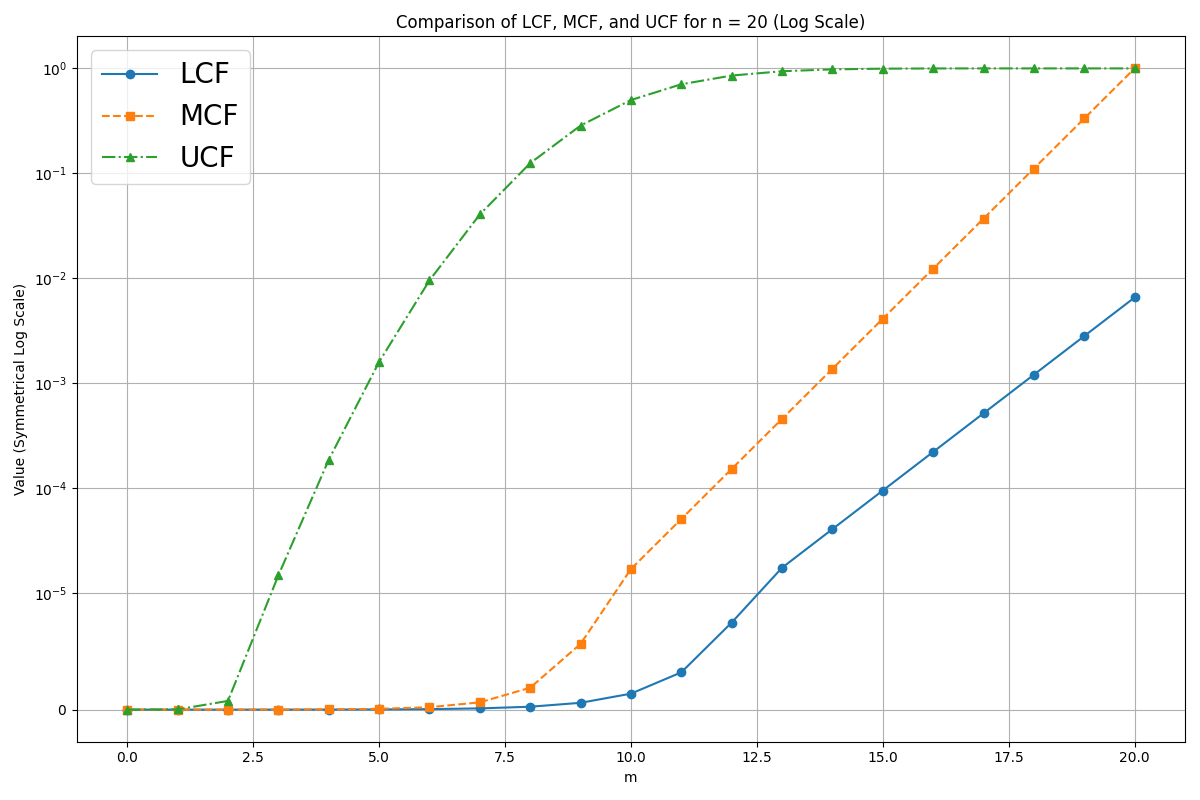}
        \caption{}
    \end{subfigure}

    \vspace{1em}

    % Row 3: 2 images
    \begin{subfigure}{0.45\textwidth}
        \includegraphics[width=\linewidth]{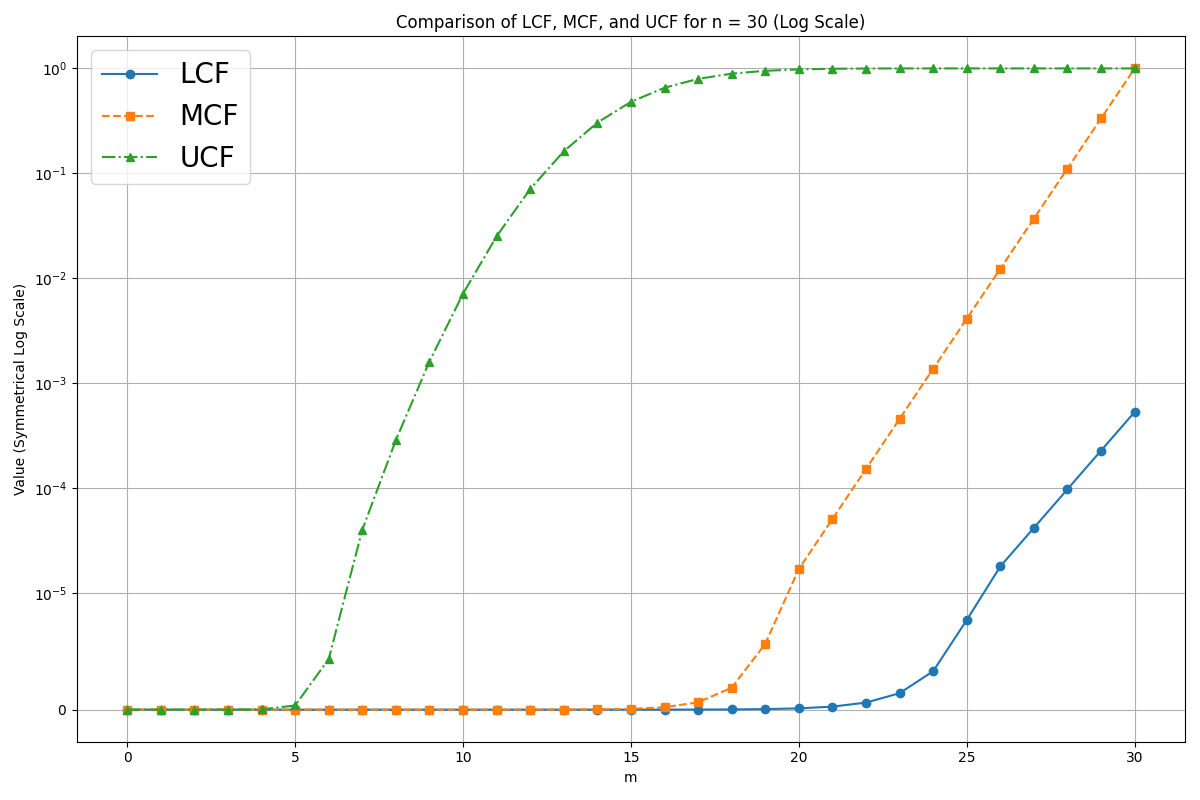}
        \caption{}
    \end{subfigure}\hfill
    \begin{subfigure}{0.45\textwidth}
        \includegraphics[width=\linewidth]{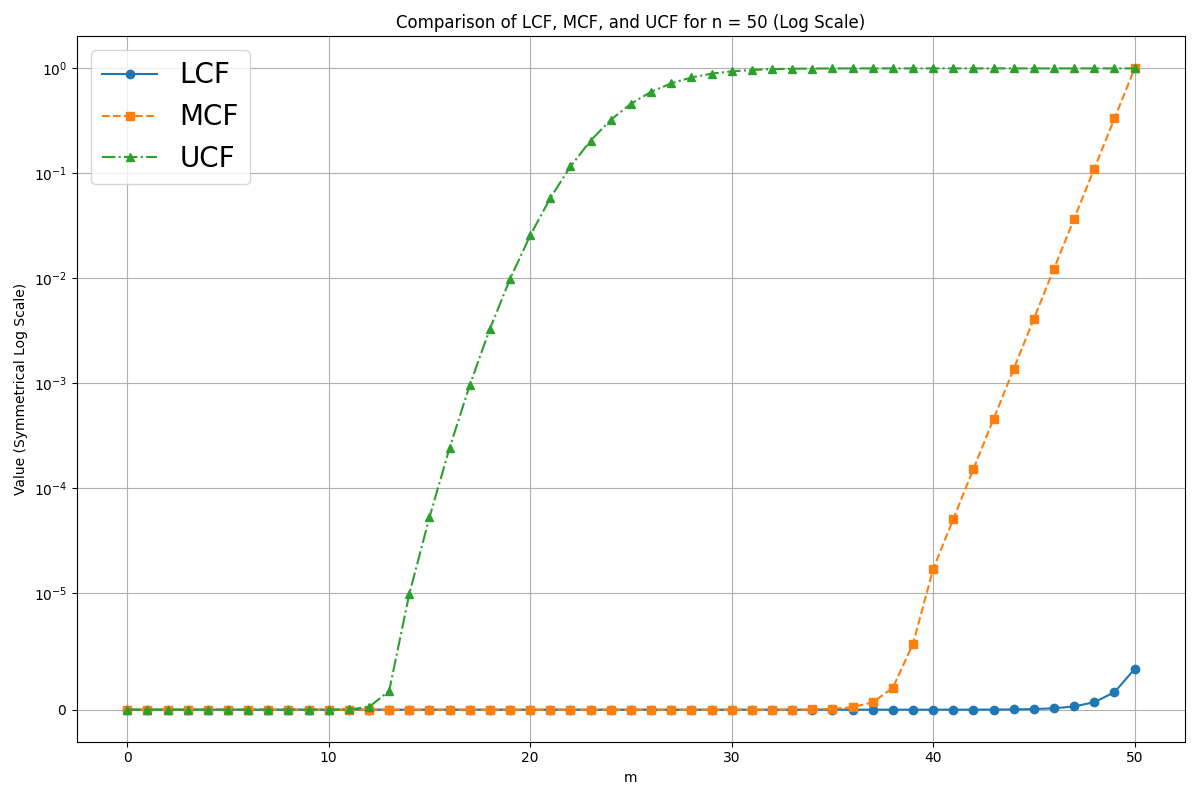}
        \caption{}
    \end{subfigure}

    \caption{Log plots of Lower, Middle and Upper bounds for all values of m upto a certain n, for n = 1, 5, 10, 20, 30, 50 in that order.}
    \label{fig:3x2grid}
\end{figure}

\begin{figure}[htbp]
    \centering

    \begin{subfigure}{0.4\textwidth}
        \centering
        \includegraphics[width=\linewidth]{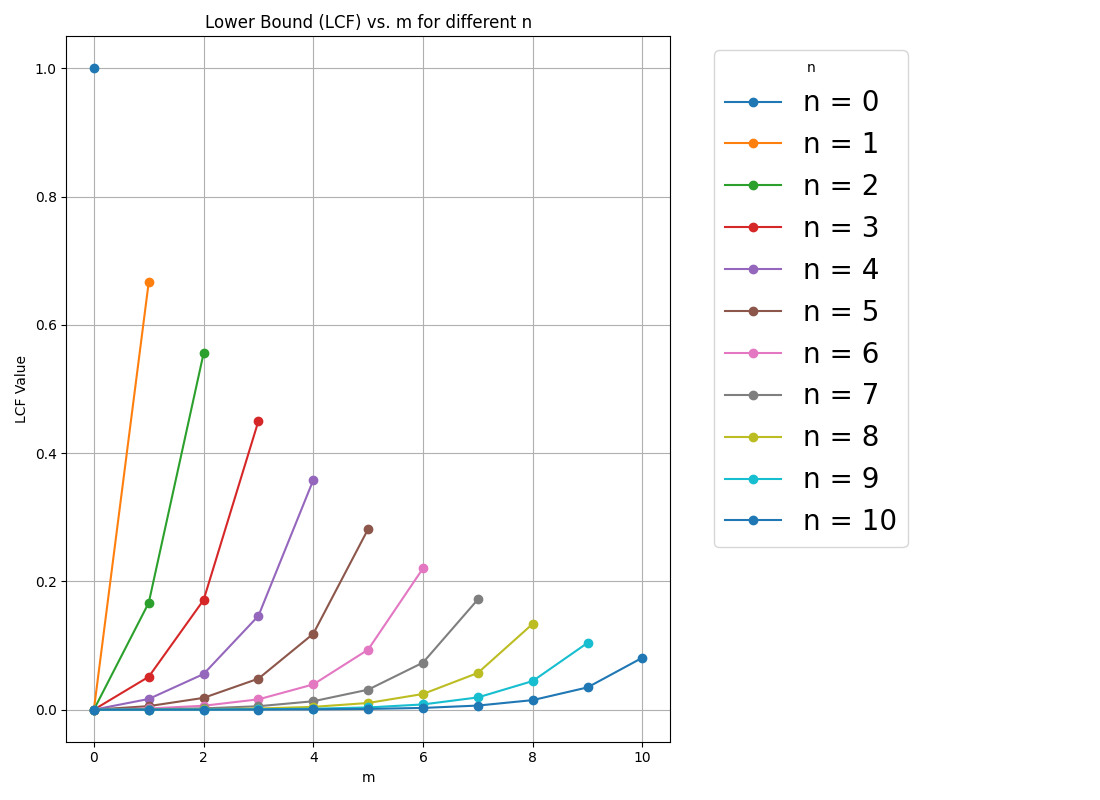}
        \caption{}
        \label{fig:image1}
    \end{subfigure}

    \vspace{1em}

    \begin{subfigure}{0.4\textwidth}
        \centering
        \includegraphics[width=\linewidth]{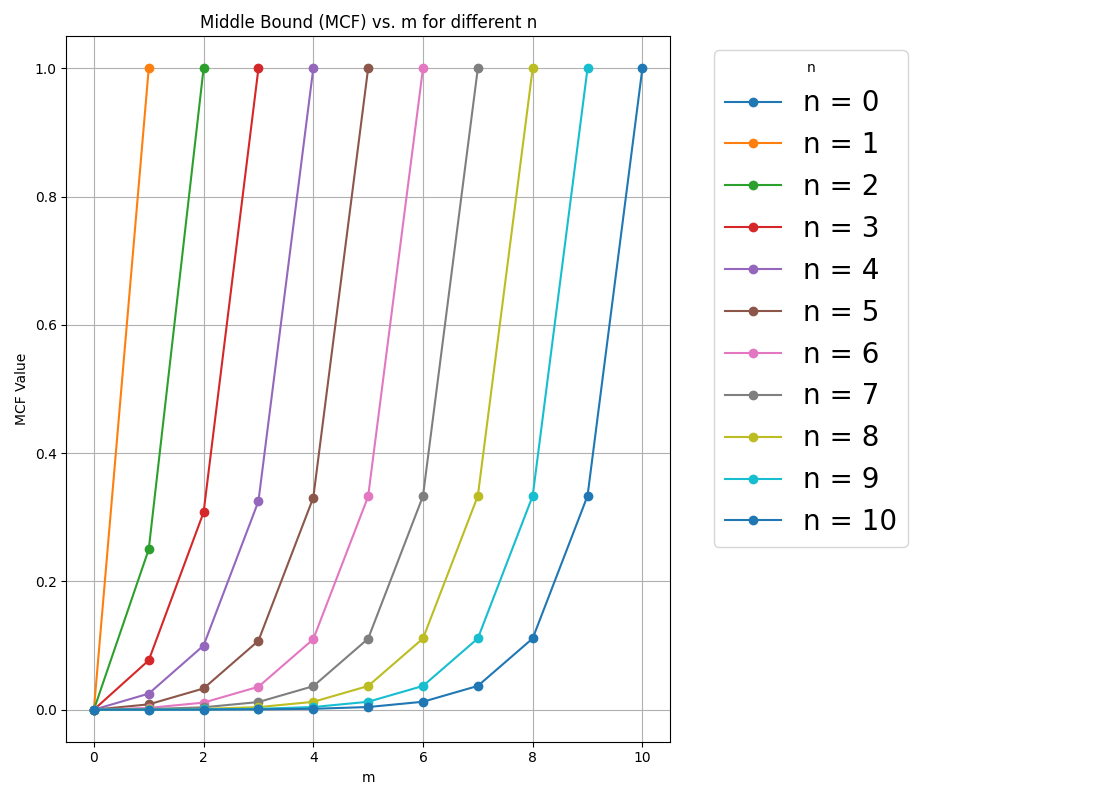}
        \caption{}
        \label{fig:image2}
    \end{subfigure}

    \vspace{1em}

    \begin{subfigure}{0.4\textwidth}
        \centering
        \includegraphics[width=\linewidth]{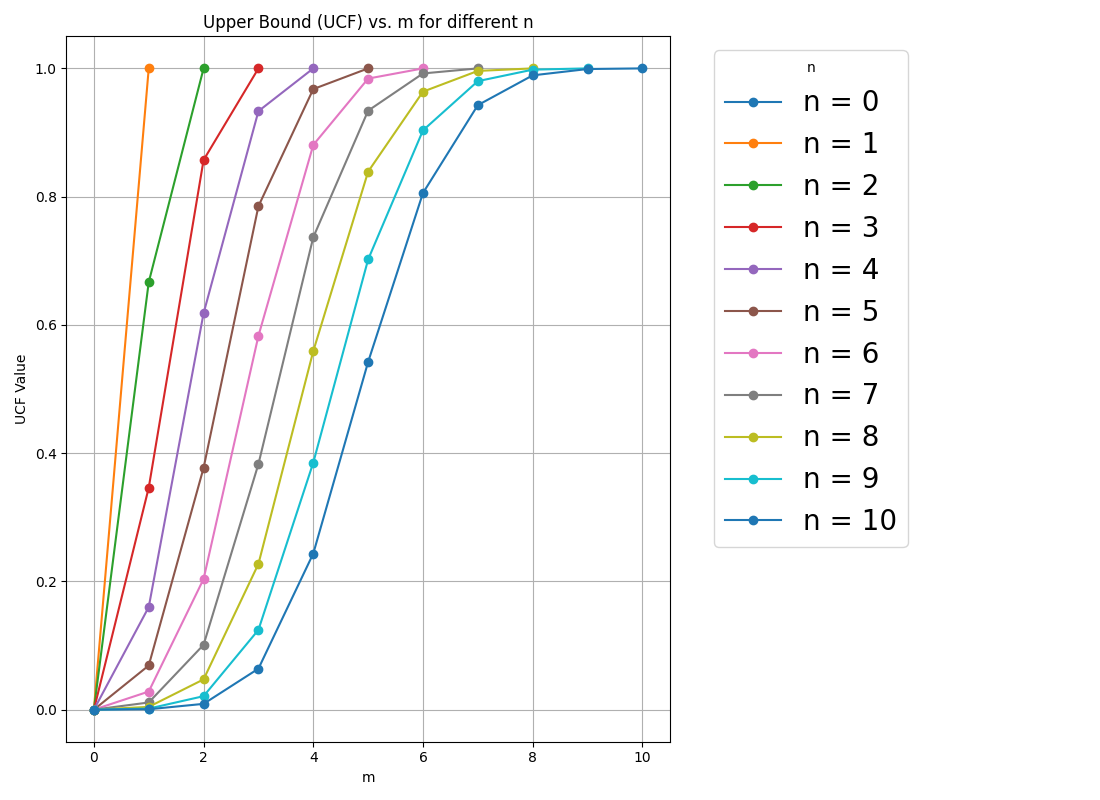}
        \caption{}
        \label{fig:image3}
    \end{subfigure}

    \caption{Plots of $\mathcal{LCF}$, $\mathcal{MCF}$ and $\mathcal{UCF}$ for various values of n (all values of m upto that value of n)}
    \label{fig:three_vertical_images}
\end{figure}

% PAGE 1 FOR CELL 1: INDIVIDUAL RESULTS (1k, 2k, 4k, 8k)
\begin{table}[H]
\centering
\captionsetup{font=small}

\section{Results for \texorpdfstring{$e\mathcal{DCF}$}{eDCF}:}\label{eDCFrunsvalues}

\subsection{Benchmark Datasets Ground Truths:}
\begin{table}[H]
\centering
\caption{Manifold Specifications}
\label{tab:manifolds}
\begin{tabular}{lrr}
\toprule
\textbf{Manifold} & \textbf{Intrinsic Dimension (ID)} & \textbf{Embedding Dimension} \\
\midrule
M1\_Sphere & 10 & 11 \\
M2\_Affine\_3to5 & 3 & 5 \\
M3\_Nonlinear\_4to6 & 4 & 6 \\
M4\_Nonlinear & 4 & 8 \\
M5a\_Helix1d & 1 & 3 \\
M5b\_Helix2d & 2 & 3 \\
M6\_Nonlinear & 6 & 36 \\
M7\_Roll & 2 & 3 \\
M8\_Nonlinear & 12 & 72 \\
M9\_Affine & 20 & 20 \\
M10a\_Cubic & 10 & 11 \\
M10b\_Cubic & 17 & 18 \\
M10c\_Cubic & 24 & 25 \\
M10d\_Cubic & 70 & 71 \\
M11\_Moebius & 2 & 3 \\
M12\_Norm & 20 & 20 \\
M13a\_Scurve & 2 & 3 \\
M13b\_Spiral & 1 & 13 \\
Mbeta & 10 & 40 \\
Mn1\_Nonlinear & 18 & 72 \\
Mn2\_Nonlinear & 24 & 96 \\
Mp1\_Paraboloid & 3 & 12 \\
Mp2\_Paraboloid & 6 & 21 \\
Mp3\_Paraboloid & 9 & 30 \\
\bottomrule
\end{tabular}
\end{table}

\subsection{Our Results:}
\textbf{Results for $e\mathcal{DCF}$ runs on benchmark datasets with 1 \% noise:}
\begin{subtable}[t]{0.48\textwidth}
    \centering\small
    \caption{Individual (1000 points)}
    \begin{tabular}{rrrr}
    \toprule
    \textbf{Known} & \textbf{MLE} & \textbf{TwoNN} & \textbf{$e\mathcal{DCF}$} \\ \midrule
    10 & 9 & 10 & 10 \\ 3 & 3 & 3 & 3 \\ 4 & 4 & 4 & 3 \\
    4 & 4 & 4 & 5 \\ 1 & 1 & 1 & 1 \\ 2 & 3 & 2 & 1 \\
    6 & 6 & 6 & 12 \\ 2 & 2 & 2 & 1 \\ 12 & 13 & 14 & 35 \\
    20 & 14 & 16 & 35 \\ 10 & 9 & 9 & 10 \\ 17 & 13 & 13 & 29 \\
    24 & 16 & 18 & 35 \\ 70 & 33 & 39 & 35 \\ 2 & 2 & 2 & 1 \\
    20 & 15 & 18 & 7 \\ 2 & 2 & 2 & 1 \\ 1 & 2 & 1 & 1 \\
    10 & 5 & 6 & 6 \\ 18 & 13 & 14 & 11 \\ 24 & 16 & 18 & 15 \\
    3 & 3 & 3 & 3 \\ 6 & 4 & 5 & 4 \\ 9 & 5 & 7 & 3 \\ \bottomrule
    \end{tabular}
\end{subtable}
\hfill
\begin{subtable}[t]{0.48\textwidth}
    \centering\small
    \caption{Individual (2000 points)}
    \begin{tabular}{rrrr}
    \toprule
    \textbf{Known} & \textbf{MLE} & \textbf{TwoNN} & \textbf{$e\mathcal{DCF}$} \\ \midrule
    10 & 9 & 10 & 10 \\ 3 & 3 & 3 & 3 \\ 4 & 4 & 4 & 4 \\
    4 & 4 & 4 & 5 \\ 1 & 1 & 2 & 0 \\ 2 & 3 & 2 & 1 \\
    6 & 6 & 6 & 12 \\ 2 & 2 & 2 & 1 \\ 12 & 14 & 14 & 41 \\
    20 & 14 & 15 & 23 \\ 10 & 9 & 9 & 10 \\ 17 & 13 & 14 & 19 \\
    24 & 17 & 18 & 41 \\ 70 & 35 & 40 & 39 \\ 2 & 2 & 2 & 1 \\
    20 & 15 & 17 & 18 \\ 2 & 2 & 2 & 1 \\ 1 & 2 & 1 & 0 \\
    10 & 6 & 6 & 7 \\ 18 & 13 & 14 & 11 \\ 24 & 17 & 18 & 15 \\
    3 & 3 & 3 & 3 \\ 6 & 5 & 5 & 5 \\ 9 & 6 & 7 & 4 \\ \bottomrule
    \end{tabular}
\end{subtable}

\vspace{1cm}

\begin{subtable}[t]{0.48\textwidth}
    \centering\small
    \caption{Individual (4000 points)}
    \begin{tabular}{rrrr}
    \toprule
    \textbf{Known} & \textbf{MLE} & \textbf{TwoNN} & \textbf{eDCF} \\ \midrule
    10 & 9 & 10 & 10 \\ 3 & 3 & 3 & 3 \\ 4 & 4 & 4 & 4 \\
    4 & 4 & 4 & 4 \\ 1 & 1 & 2 & 0 \\ 2 & 2 & 2 & 1 \\
    6 & 6 & 6 & 12 \\ 2 & 2 & 2 & 1 \\ 12 & 14 & 13 & 39 \\
    20 & 15 & 16 & 23 \\ 10 & 9 & 10 & 10 \\ 17 & 13 & 14 & 19 \\
    24 & 18 & 19 & 34 \\ 70 & 37 & 41 & 39 \\ 2 & 2 & 2 & 1 \\
    20 & 16 & 17 & 17 \\ 2 & 2 & 2 & 1 \\ 1 & 1 & 1 & 0 \\
    10 & 6 & 7 & 8 \\ 18 & 14 & 15 & 12 \\ 24 & 17 & 19 & 16 \\
    3 & 3 & 3 & 4 \\ 6 & 5 & 5 & 5 \\ 9 & 6 & 7 & 5 \\ \bottomrule
    \end{tabular}
\end{subtable}
\hfill
\begin{subtable}[t]{0.48\textwidth}
    \centering\small
    \caption{Individual (8000 points)}
    \begin{tabular}{rrrr}
    \toprule
    \textbf{Known} & \textbf{MLE} & \textbf{TwoNN} & \textbf{eDCF} \\ \midrule
    10 & 9 & 10 & 10 \\ 3 & 3 & 3 & 3 \\ 4 & 4 & 4 & 4 \\
    4 & 4 & 4 & 4 \\ 1 & 1 & 2 & 0 \\ 2 & 2 & 2 & 1 \\
    6 & 6 & 6 & 8 \\ 2 & 2 & 2 & 1 \\ 12 & 13 & 13 & 26 \\
    20 & 15 & 16 & 12 \\ 10 & 9 & 9 & 10 \\ 17 & 14 & 14 & 19 \\
    24 & 18 & 20 & 26 \\ 70 & 38 & 42 & 26 \\ 2 & 2 & 2 & 1 \\
    20 & 16 & 17 & 15 \\ 2 & 2 & 2 & 1 \\ 1 & 1 & 1 & 0 \\
    10 & 6 & 7 & 8 \\ 18 & 14 & 15 & 12 \\ 24 & 18 & 19 & 15 \\
    3 & 3 & 3 & 3 \\ 6 & 5 & 6 & 5 \\ 9 & 7 & 8 & 5 \\ \bottomrule
    \end{tabular}
\end{subtable}
\caption{Individual dimension estimates for 1 \% noise.}
\end{table}

% PAGE 2 FOR CELL 1: INDIVIDUAL RESULTS (16k, 32k, 64k) + MAE METRIC
\begin{table}[H]
\centering
\captionsetup{font=small}

\begin{subtable}[t]{0.48\textwidth}
    \centering\small
    \caption{Individual (16000 points)}
    \begin{tabular}{rrrr}
    \toprule
    \textbf{Known} & \textbf{MLE} & \textbf{TwoNN} & \textbf{$e\mathcal{DCF}$} \\ \midrule
    10 & 9 & 10 & 10 \\ 3 & 3 & 3 & 3 \\ 4 & 4 & 4 & 4 \\
    4 & 4 & 4 & 4 \\ 1 & 1 & 3 & 1 \\ 2 & 2 & 2 & 1 \\
    6 & 6 & 6 & 8 \\ 2 & 2 & 2 & 1 \\ 12 & 13 & 13 & 27 \\
    20 & 15 & 16 & 22 \\ 10 & 9 & 9 & 10 \\ 17 & 14 & 15 & 17 \\
    24 & 18 & 19 & 26 \\ 70 & 39 & 43 & 28 \\ 2 & 2 & 2 & 1 \\
    20 & 17 & 18 & 17 \\ 2 & 2 & 2 & 1 \\ 1 & 1 & 1 & 0 \\
    10 & 6 & 7 & 8 \\ 18 & 14 & 15 & 12 \\ 24 & 18 & 19 & 15 \\
    3 & 3 & 3 & 4 \\ 6 & 5 & 6 & 6 \\ 9 & 7 & 8 & 6 \\ \bottomrule
    \end{tabular}
\end{subtable}
\hfill
\begin{subtable}[t]{0.48\textwidth}
    \centering\small
    \caption{Individual (32000 points)}
    \begin{tabular}{rrrr}
    \toprule
    \textbf{Known} & \textbf{MLE} & \textbf{TwoNN} & \textbf{$e\mathcal{DCF}$} \\ \midrule
    10 & 9 & 10 & 10 \\ 3 & 3 & 3 & 3 \\ 4 & 4 & 4 & 4 \\
    4 & 4 & 4 & 4 \\ 1 & 2 & 3 & 1 \\ 2 & 2 & 2 & 1 \\
    6 & 6 & 6 & 8 \\ 2 & 2 & 2 & 1 \\ 12 & 13 & 13 & 26 \\
    20 & 16 & 16 & 22 \\ 10 & 9 & 9 & 9 \\ 17 & 14 & 15 & 18 \\
    24 & 19 & 20 & 26 \\ 70 & 41 & 44 & 31 \\ 2 & 2 & 2 & 1 \\
    20 & 17 & 18 & 14 \\ 2 & 2 & 2 & 1 \\ 1 & 1 & 1 & 0 \\
    10 & 7 & 7 & 8 \\ 18 & 15 & 15 & 12 \\ 24 & 18 & 19 & 15 \\
    3 & 3 & 4 & 4 \\ 6 & 5 & 6 & 6 \\ 9 & 7 & 8 & 7 \\ \bottomrule
    \end{tabular}
\end{subtable}

\vspace{1cm}

\begin{subtable}[t]{0.48\textwidth}
    \centering\small
    \caption{Individual (64000 points)}
    \begin{tabular}{rrrr}
    \toprule
    \textbf{Known} & \textbf{MLE} & \textbf{TwoNN} & \textbf{$e\mathcal{DCF}$} \\ \midrule
    10 & 10 & 10 & 10 \\ 3 & 3 & 3 & 3 \\ 4 & 4 & 4 & 4 \\
    4 & 4 & 4 & 4 \\ 1 & 2 & 3 & 1 \\ 2 & 2 & 2 & 1 \\
    6 & 6 & 6 & 8 \\ 2 & 2 & 2 & 1 \\ 12 & 13 & 13 & 16 \\
    20 & 16 & 16 & 20 \\ 10 & 9 & 10 & 9 \\ 17 & 14 & 15 & 18 \\
    24 & 19 & 20 & 16 \\ 70 & 42 & 45 & 32 \\ 2 & 2 & 3 & 1 \\
    20 & 18 & 18 & 16 \\ 2 & 2 & 2 & 1 \\ 1 & 1 & 2 & 1 \\
    10 & 7 & 7 & 5 \\ 18 & 15 & 15 & 13 \\ 24 & 19 & 19 & 16 \\
    3 & 3 & 4 & 4 \\ 6 & 6 & 6 & 6 \\ 9 & 8 & 8 & 7 \\ \bottomrule
    \end{tabular}
\end{subtable}
\hfill
\begin{subtable}[t]{0.48\textwidth}
    \centering
    \caption{MAE (absolute error)}
    \begin{tabular}{rrrr}
    \toprule
     & \textbf{MLE} & \textbf{TwoNN} & \textbf{$e\mathcal{DCF}$} \\ \midrule
    \textbf{1k} & 3.708 & 2.792 & 6.208 \\
    \textbf{2k} & 3.458 & 2.833 & 5.083 \\
    \textbf{4k} & 3.125 & 2.500 & 4.583 \\
    \textbf{8k} & 2.917 & 2.375 & 4.375 \\
    \textbf{16k} & 2.833 & 2.333 & 3.833 \\
    \textbf{32k} & 2.625 & 2.292 & 3.833 \\
    \textbf{64k} & 2.375 & 2.292 & 3.458 \\ \bottomrule
    \end{tabular}
\end{subtable}
\caption{Individual dimension estimates and MAE for 1 \% noise.}
\end{table}

% PAGE 3 FOR CELL 1: FINAL TWO METRIC TABLES
\begin{table}[H]
\centering
\captionsetup{font=small}

\begin{subfigure}[t]{0.48\textwidth}
    \centering
    \caption{Mean signed error (bias)}
    \begin{tabular}{rrrr}
    \toprule
     & \textbf{MLE} & \textbf{TwoNN} & \textbf{$e\mathcal{DCF}$} \\ \midrule
    \textbf{1k} & 3.458 & 2.625 & 0.542 \\
    \textbf{2k} & 3.125 & 2.583 & 0.250 \\
    \textbf{4k} & 2.958 & 2.333 & 0.500 \\
    \textbf{8k} & 2.833 & 2.208 & 2.708 \\
    \textbf{16k} & 2.750 & 2.083 & 2.000 \\
    \textbf{32k} & 2.458 & 1.958 & 2.000 \\
    \textbf{64k} & 2.208 & 1.792 & 2.792 \\ \bottomrule
    \end{tabular}
\end{subfigure}
\hfill
\begin{subfigure}[t]{0.48\textwidth}
    \centering
    \caption{\% exact matches}
    \begin{tabular}{rrrr}
    \toprule
     & \textbf{MLE} & \textbf{TwoNN} & \textbf{$e\mathcal{DCF}$} \\ \midrule
    \textbf{1k} & 37.5 & 50.0 & 25.0 \\
    \textbf{2k} & 37.5 & 45.8 & 20.8 \\
    \textbf{4k} & 45.8 & 50.0 & 20.8 \\
    \textbf{8k} & 45.8 & 50.0 & 25.0 \\
    \textbf{16k} & 45.8 & 50.0 & 33.3 \\
    \textbf{32k} & 41.7 & 45.8 & 25.0 \\
    \textbf{64k} & 50.0 & 41.7 & 33.3 \\ \bottomrule
    \end{tabular}
\end{subfigure}
\caption{Error metrics for 1 \% noise.}
\end{table}

\newpage

\textbf{Results for $e\mathcal{DCF}$ for benchmark datasets with 10 \% noise:}
% PAGE 1 FOR CELL 2: INDIVIDUAL RESULTS (1k, 2k, 4k, 8k)
\begin{table}[H]
\centering
\captionsetup{font=small}

\begin{subtable}[t]{0.48\textwidth}
    \centering\small
    \caption{Individual (1000 points)}
    \begin{tabular}{rrrr}
    \toprule
    \textbf{Known} & \textbf{MLE} & \textbf{TwoNN} & \textbf{$e\mathcal{DCF}$} \\ \midrule
    10 & 9 & 10 & 11 \\ 3 & 3 & 3 & 3 \\ 4 & 4 & 4 & 3 \\
    4 & 4 & 4 & 5 \\ 1 & 1 & 3 & 1 \\ 2 & 3 & 2 & 1 \\
    6 & 7 & 6 & 14 \\ 2 & 2 & 2 & 1 \\ 12 & 13 & 14 & 37 \\
    20 & 14 & 16 & 37 \\ 10 & 9 & 9 & 11 \\ 17 & 13 & 13 & 37 \\
    24 & 16 & 18 & 37 \\ 70 & 33 & 40 & 37 \\ 2 & 2 & 3 & 1 \\
    20 & 15 & 17 & 18 \\ 2 & 2 & 3 & 1 \\ 1 & 2 & 1 & 1 \\
    10 & 7 & 9 & 6 \\ 18 & 13 & 15 & 14 \\ 24 & 16 & 19 & 18 \\
    3 & 3 & 5 & 4 \\ 6 & 5 & 6 & 4 \\ 9 & 5 & 7 & 3 \\ \bottomrule
    \end{tabular}
\end{subtable}
\hfill
\begin{subtable}[t]{0.48\textwidth}
    \centering\small
    \caption{Individual (2000 points)}
    \begin{tabular}{rrrr}
    \toprule
    \textbf{Known} & \textbf{MLE} & \textbf{TwoNN} & \textbf{$e\mathcal{DCF}$} \\ \midrule
    10 & 9 & 9 & 10 \\ 3 & 3 & 3 & 3 \\ 4 & 4 & 4 & 3 \\
    4 & 4 & 4 & 5 \\ 1 & 1 & 3 & 0 \\ 2 & 3 & 2 & 1 \\
    6 & 7 & 6 & 16 \\ 2 & 2 & 2 & 1 \\ 12 & 14 & 14 & 41 \\
    20 & 14 & 15 & 24 \\ 10 & 9 & 9 & 10 \\ 17 & 13 & 14 & 10 \\
    24 & 17 & 19 & 37 \\ 70 & 35 & 40 & 41 \\ 2 & 2 & 3 & 1 \\
    20 & 15 & 17 & 16 \\ 2 & 2 & 3 & 1 \\ 1 & 2 & 1 & 0 \\
    10 & 8 & 10 & 6 \\ 18 & 14 & 15 & 14 \\ 24 & 17 & 18 & 16 \\
    3 & 4 & 6 & 4 \\ 6 & 5 & 7 & 5 \\ 9 & 6 & 8 & 4 \\ \bottomrule
    \end{tabular}
\end{subtable}

\vspace{1cm}

\begin{subtable}[t]{0.48\textwidth}
    \centering\small
    \caption{Individual (4000 points)}
    \begin{tabular}{rrrr}
    \toprule
    \textbf{Known} & \textbf{MLE} & \textbf{TwoNN} & \textbf{$e\mathcal{DCF}$} \\ \midrule
    10 & 9 & 10 & 10 \\ 3 & 3 & 3 & 3 \\ 4 & 4 & 5 & 3 \\
    4 & 4 & 4 & 4 \\ 1 & 2 & 3 & 1 \\ 2 & 2 & 2 & 1 \\
    6 & 6 & 6 & 14 \\ 2 & 2 & 2 & 1 \\ 12 & 14 & 14 & 39 \\
    20 & 15 & 16 & 12 \\ 10 & 9 & 10 & 11 \\ 17 & 13 & 14 & 19 \\
    24 & 18 & 19 & 39 \\ 70 & 37 & 41 & 39 \\ 2 & 2 & 3 & 1 \\
    20 & 16 & 17 & 18 \\ 2 & 2 & 3 & 1 \\ 1 & 1 & 1 & 0 \\
    10 & 8 & 11 & 7 \\ 18 & 14 & 15 & 13 \\ 24 & 18 & 19 & 18 \\
    3 & 4 & 7 & 4 \\ 6 & 6 & 7 & 5 \\ 9 & 7 & 8 & 5 \\ \bottomrule
    \end{tabular}
\end{subtable}
\hfill
\begin{subtable}[t]{0.48\textwidth}
    \centering\small
    \caption{Individual (8000 points)}
    \begin{tabular}{rrrr}
    \toprule
    \textbf{Known} & \textbf{MLE} & \textbf{TwoNN} & \textbf{$e\mathcal{DCF}$} \\ \midrule
    10 & 9 & 10 & 10 \\ 3 & 3 & 3 & 3 \\ 4 & 4 & 5 & 4 \\
    4 & 4 & 5 & 4 \\ 1 & 2 & 3 & 1 \\ 2 & 2 & 2 & 1 \\
    6 & 6 & 6 & 15 \\ 2 & 2 & 2 & 1 \\ 12 & 14 & 13 & 42 \\
    20 & 15 & 16 & 22 \\ 10 & 9 & 10 & 10 \\ 17 & 14 & 14 & 20 \\
    24 & 18 & 20 & 28 \\ 70 & 38 & 42 & 42 \\ 2 & 3 & 3 & 1 \\
    20 & 16 & 17 & 16 \\ 2 & 2 & 3 & 1 \\ 1 & 1 & 2 & 0 \\
    10 & 9 & 12 & 8 \\ 18 & 14 & 15 & 14 \\ 24 & 18 & 19 & 17 \\
    3 & 5 & 8 & 4 \\ 6 & 6 & 7 & 6 \\ 9 & 7 & 9 & 5 \\ \bottomrule
    \end{tabular}
\end{subtable}
\caption{Individual dimension estimates for 10 \% noise.}
\end{table}

% PAGE 2 FOR CELL 2: INDIVIDUAL RESULTS (16k, 32k/64k) + METRICS
\begin{table}[H]
\centering
\captionsetup{font=small}

\begin{subtable}[t]{0.48\textwidth}
    \centering\small
    \caption{Individual (16000 points)}
    \begin{tabular}{rrrr}
    \toprule
    \textbf{Known} & \textbf{MLE} & \textbf{TwoNN} & \textbf{$e\mathcal{DCF}$} \\ \midrule
    10 & 10 & 10 & 10 \\ 3 & 3 & 4 & 3 \\ 4 & 4 & 5 & 3 \\
    4 & 4 & 5 & 4 \\ 1 & 3 & 3 & 1 \\ 2 & 2 & 2 & 1 \\
    6 & 6 & 7 & 11 \\ 2 & 2 & 2 & 1 \\ 12 & 14 & 13 & 35 \\
    20 & 16 & 16 & 22 \\ 10 & 9 & 10 & 10 \\ 17 & 14 & 15 & 19 \\
    24 & 19 & 20 & 31 \\ 70 & 41 & 44 & 34 \\ 2 & 3 & 3 & 1 \\
    20 & 17 & 18 & 16 \\ 2 & 3 & 3 & 1 \\ 1 & 1 & 5 & 1 \\
    10 & 10 & 14 & 10 \\ 18 & 15 & 16 & 15 \\ 24 & 19 & 20 & 17 \\
    3 & 7 & 9 & 5 \\ 6 & 7 & 8 & 6 \\ 9 & 8 & 9 & 7 \\ \bottomrule
    \end{tabular}
\end{subtable}
\hfill
\begin{subtable}[t]{0.48\textwidth}
    \centering\small
    \caption{Individual (32k/64k points)}
    \begin{tabular}{rrrr}
    \toprule
    \textbf{Known} & \textbf{MLE} & \textbf{TwoNN} & \textbf{$e\mathcal{DCF}$} \\ \midrule
    10 & 10 & 10 & 10 \\ 3 & 3 & 4 & 3 \\ 4 & 5 & 5 & 3 \\
    4 & 5 & 5 & 4 \\ 1 & 3 & 3 & 1 \\ 2 & 2 & 2 & 1 \\
    6 & 6 & 7 & 10 \\ 2 & 2 & 2 & 1 \\ 12 & 13 & 13 & 31 \\
    20 & 16 & 17 & 21 \\ 10 & 9 & 10 & 9 \\ 17 & 14 & 15 & 20 \\
    24 & 19 & 20 & 27 \\ 70 & 42 & 45 & 42 \\ 2 & 3 & 3 & 1 \\
    20 & 18 & 18 & 17 \\ 2 & 3 & 3 & 1 \\ 1 & 2 & 7 & 1 \\
    10 & 11 & 15 & 10 \\ 18 & 15 & 16 & 16 \\ 24 & 19 & 20 & 19 \\
    3 & 7 & 9 & 5 \\ 6 & 7 & 8 & 7 \\ 9 & 8 & 10 & 8 \\ \bottomrule
    \end{tabular}
\end{subtable}

\vspace{1cm}

\begin{subtable}[t]{0.48\textwidth}
    \centering
    \caption{MAE (absolute error)}
    \begin{tabular}{rrrr}
    \toprule
     & \textbf{MLE} & \textbf{TwoNN} & \textbf{$e\mathcal{DCF}$} \\ \midrule
    \textbf{1k} & 3.625 & 2.792 & 6.208 \\
    \textbf{2k} & 3.417 & 2.833 & 5.292 \\
    \textbf{4k} & 3.000 & 2.750 & 5.000 \\
    \textbf{8k} & 2.958 & 2.750 & 4.292 \\
    \textbf{16k} & 2.708 & 2.875 & 4.083 \\
    \textbf{32k} & 2.750 & 2.958 & 3.250 \\
    \textbf{64k} & 2.750 & 2.958 & 3.250 \\ \bottomrule
    \end{tabular}
\end{subtable}
\hfill
\begin{subtable}[t]{0.48\textwidth}
    \centering
    \caption{Mean signed error (bias)}
    \begin{tabular}{rrrr}
    \toprule
     & \textbf{MLE} & \textbf{TwoNN} & \textbf{$e\mathcal{DCF}$} \\ \midrule
    \textbf{1k} & 3.292 & 2.125 & -1.042 \\
    \textbf{2k} & 2.917 & 2.000 & 0.458 \\
    \textbf{4k} & 2.667 & 1.667 & 0.500 \\
    \textbf{8k} & 2.458 & 1.417 & 0.208 \\
    \textbf{16k} & 1.792 & 0.792 & 0.667 \\
    \textbf{32k} & 1.583 & 0.542 & 0.500 \\
    \textbf{64k} & 1.583 & 0.542 & 0.500 \\ \bottomrule
    \end{tabular}
\end{subtable}
\caption{Individual estimates and error metrics for 10 \% noise.}
\end{table}

% PAGE 3 FOR CELL 2: FINAL METRIC TABLE
\begin{table}[H]
\centering
\captionsetup{font=small}

\begin{subtable}[t]{0.48\textwidth}
    \centering
    \caption{\% exact matches}
    \begin{tabular}{rrrr}
    \toprule
     & \textbf{MLE} & \textbf{TwoNN} & \textbf{$e\mathcal{DCF}$} \\ \midrule
    \textbf{1k} & 33.3 & 37.5 & 12.5 \\
    \textbf{2k} & 29.2 & 33.3 & 12.5 \\
    \textbf{4k} & 41.7 & 33.3 & 16.7 \\
    \textbf{8k} & 37.5 & 29.2 & 29.2 \\
    \textbf{16k} & 37.5 & 20.8 & 33.3 \\
    \textbf{32k} & 20.8 & 16.7 & 25.0 \\
    \textbf{64k} & 20.8 & 16.7 & 25.0 \\ \bottomrule
    \end{tabular}
\end{subtable}
\caption{Accuracy metric for 10 \% noise.}
\end{table}

\newpage

\textbf{Results for $e\mathcal{DCF}$ on benchmark datasets with 30 \% noise:}
% PAGE 1 FOR CELL 3: INDIVIDUAL RESULTS (1k, 2k, 4k, 8k)
\begin{table}[H]
\centering
\captionsetup{font=small}

\begin{subtable}[t]{0.48\textwidth}
    \centering\small
    \caption{Individual (1000 points)}
    \begin{tabular}{rrrr}
    \toprule
    \textbf{Known} & \textbf{MLE} & \textbf{TwoNN} & \textbf{$e\mathcal{DCF}$} \\ \midrule
    10 & 9 & 10 & 12 \\ 3 & 3 & 4 & 3 \\ 4 & 4 & 5 & 4 \\
    4 & 5 & 6 & 6 \\ 1 & 2 & 3 & 1 \\ 2 & 3 & 2 & 1 \\
    6 & 7 & 8 & 24 \\ 2 & 2 & 2 & 1 \\ 12 & 14 & 15 & 30 \\
    20 & 14 & 16 & 30 \\ 10 & 9 & 9 & 13 \\ 17 & 13 & 15 & 30 \\
    24 & 17 & 20 & 30 \\ 70 & 33 & 40 & 30 \\ 2 & 3 & 3 & 1 \\
    20 & 15 & 17 & 22 \\ 2 & 2 & 3 & 1 \\ 1 & 2 & 1 & 1 \\
    10 & 12 & 17 & 13 \\ 18 & 15 & 18 & 22 \\ 24 & 18 & 23 & 30 \\
    3 & 6 & 8 & 5 \\ 6 & 6 & 9 & 4 \\ 9 & 6 & 10 & 3 \\ \bottomrule
    \end{tabular}
\end{subtable}
\hfill
\begin{subtable}[t]{0.48\textwidth}
    \centering\small
    \caption{Individual (2000 points)}
    \begin{tabular}{rrrr}
    \toprule
    \textbf{Known} & \textbf{MLE} & \textbf{TwoNN} & \textbf{$e\mathcal{DCF}$} \\ \midrule
    10 & 9 & 9 & 11 \\ 3 & 3 & 4 & 3 \\ 4 & 5 & 5 & 3 \\
    4 & 5 & 6 & 6 \\ 1 & 2 & 3 & 1 \\ 2 & 3 & 2 & 1 \\
    6 & 8 & 8 & 25 \\ 2 & 2 & 2 & 1 \\ 12 & 15 & 16 & 41 \\
    20 & 14 & 14 & 31 \\ 10 & 9 & 10 & 12 \\ 17 & 13 & 14 & 29 \\
    24 & 17 & 19 & 41 \\ 70 & 35 & 40 & 41 \\ 2 & 3 & 3 & 1 \\
    20 & 15 & 17 & 25 \\ 2 & 3 & 3 & 1 \\ 1 & 2 & 2 & 0 \\
    10 & 14 & 18 & 17 \\ 18 & 16 & 18 & 25 \\ 24 & 19 & 22 & 29 \\
    3 & 7 & 9 & 5 \\ 6 & 7 & 10 & 5 \\ 9 & 8 & 11 & 4 \\ \bottomrule
    \end{tabular}
\end{subtable}

\vspace{1cm}

\begin{subtable}[t]{0.48\textwidth}
    \centering\small
    \caption{Individual (4000 points)}
    \begin{tabular}{rrrr}
    \toprule
    \textbf{Known} & \textbf{MLE} & \textbf{TwoNN} & \textbf{$e\mathcal{DCF}$} \\ \midrule
    10 & 9 & 10 & 11 \\ 3 & 3 & 4 & 3 \\ 4 & 5 & 5 & 4 \\
    4 & 5 & 6 & 6 \\ 1 & 3 & 3 & 1 \\ 2 & 3 & 2 & 1 \\
    6 & 8 & 9 & 17 \\ 2 & 2 & 2 & 1 \\ 12 & 15 & 15 & 37 \\
    20 & 15 & 16 & 27 \\ 10 & 9 & 10 & 11 \\ 17 & 14 & 15 & 22 \\
    24 & 18 & 19 & 37 \\ 70 & 36 & 42 & 37 \\ 2 & 3 & 3 & 1 \\
    20 & 16 & 17 & 17 \\ 2 & 3 & 3 & 1 \\ 1 & 1 & 3 & 0 \\
    10 & 15 & 20 & 15 \\ 18 & 17 & 19 & 16 \\ 24 & 20 & 23 & 22 \\
    3 & 8 & 9 & 6 \\ 6 & 8 & 11 & 6 \\ 9 & 9 & 12 & 5 \\ \bottomrule
    \end{tabular}
\end{subtable}
\hfill
\begin{subtable}[t]{0.48\textwidth}
    \centering\small
    \caption{Individual (8000 points)}
    \begin{tabular}{rrrr}
    \toprule
    \textbf{Known} & \textbf{MLE} & \textbf{TwoNN} & \textbf{$e\mathcal{DCF}$} \\ \midrule
    10 & 10 & 10 & 10 \\ 3 & 4 & 5 & 3 \\ 4 & 5 & 6 & 4 \\
    4 & 5 & 6 & 5 \\ 1 & 3 & 3 & 1 \\ 2 & 2 & 2 & 1 \\
    6 & 8 & 9 & 19 \\ 2 & 2 & 2 & 1 \\ 12 & 15 & 15 & 45 \\
    20 & 15 & 16 & 22 \\ 10 & 9 & 10 & 11 \\ 17 & 14 & 15 & 20 \\
    24 & 18 & 20 & 45 \\ 70 & 38 & 42 & 45 \\ 2 & 3 & 3 & 1 \\
    20 & 16 & 17 & 17 \\ 2 & 3 & 3 & 1 \\ 1 & 1 & 4 & 1 \\
    10 & 17 & 21 & 16 \\ 18 & 17 & 19 & 17 \\ 24 & 21 & 23 & 22 \\
    3 & 8 & 10 & 6 \\ 6 & 9 & 12 & 7 \\ 9 & 10 & 13 & 6 \\ \bottomrule
    \end{tabular}
\end{subtable}
\caption{Individual dimension estimates for 30 \% noise.}
\end{table}

% PAGE 2 FOR CELL 3: INDIVIDUAL RESULTS + METRICS
\begin{table}[H]
\centering
\captionsetup{font=small}

\begin{subtable}[t]{0.48\textwidth}
    \centering\small
    \caption{Individual (16k/32k points)}
    \begin{tabular}{rrrr}
    \toprule
    \textbf{Known} & \textbf{MLE} & \textbf{TwoNN} & \textbf{$e\mathcal{DCF}$} \\ \midrule
    10 & 10 & 10 & 10 \\ 3 & 4 & 5 & 3 \\ 4 & 5 & 6 & 4 \\
    4 & 6 & 7 & 5 \\ 1 & 3 & 3 & 1 \\ 2 & 2 & 2 & 1 \\
    6 & 8 & 10 & 25 \\ 2 & 2 & 3 & 1 \\ 12 & 15 & 15 & 46 \\
    20 & 15 & 16 & 29 \\ 10 & 9 & 10 & 10 \\ 17 & 14 & 15 & 27 \\
    24 & 19 & 20 & 38 \\ 70 & 41 & 44 & 46 \\ 2 & 3 & 3 & 2 \\
    20 & 17 & 18 & 23 \\ 2 & 3 & 3 & 1 \\ 1 & 1 & 6 & 1 \\
    10 & 18 & 22 & 16 \\ 18 & 18 & 20 & 16 \\ 24 & 22 & 24 & 29 \\
    3 & 9 & 10 & 7 \\ 6 & 10 & 13 & 7 \\ 9 & 11 & 14 & 7 \\ \bottomrule
    \end{tabular}
\end{subtable}
\hfill
\begin{subtable}[t]{0.48\textwidth}
    \centering\small
    \caption{Individual (64000 points)}
    \begin{tabular}{rrrr}
    \toprule
    \textbf{Known} & \textbf{MLE} & \textbf{TwoNN} & \textbf{$e\mathcal{DCF}$} \\ \midrule
    10 & 10 & 10 & 10 \\ 3 & 4 & 5 & 3 \\ 4 & 6 & 6 & 4 \\
    4 & 6 & 7 & 5 \\ 1 & 3 & 3 & 1 \\ 2 & 2 & 2 & 1 \\
    6 & 9 & 12 & 21 \\ 2 & 2 & 3 & 1 \\ 12 & 15 & 16 & 43 \\
    20 & 16 & 17 & 24 \\ 10 & 10 & 10 & 10 \\ 17 & 15 & 16 & 21 \\
    24 & 19 & 20 & 30 \\ 70 & 42 & 45 & 43 \\ 2 & 3 & 3 & 1 \\
    20 & 18 & 19 & 19 \\ 2 & 3 & 3 & 1 \\ 1 & 4 & 9 & 2 \\
    10 & 20 & 24 & 17 \\ 18 & 19 & 21 & 21 \\ 24 & 23 & 25 & 24 \\
    3 & 10 & 11 & 7 \\ 6 & 12 & 14 & 8 \\ 9 & 12 & 15 & 8 \\ \bottomrule
    \end{tabular}
\end{subtable}

\vspace{1cm}

\begin{subtable}[t]{0.48\textwidth}
    \centering
    \caption{MAE (absolute error)}
    \begin{tabular}{rrrr}
    \toprule
     & \textbf{MLE} & \textbf{TwoNN} & \textbf{$e\mathcal{DCF}$} \\ \midrule
    \textbf{1k} & 3.583 & 3.083 & 5.875 \\
    \textbf{2k} & 3.667 & 3.542 & 6.667 \\
    \textbf{4k} & 3.458 & 3.500 & 5.083 \\
    \textbf{8k} & 3.458 & 3.750 & 5.083 \\
    \textbf{16k} & 3.417 & 4.000 & 5.708 \\
    \textbf{32k} & 3.292 & 3.917 & 5.708 \\
    \textbf{64k} & 3.625 & 4.333 & 4.625 \\ \bottomrule
    \end{tabular}
\end{subtable}
\hfill
\begin{subtable}[t]{0.48\textwidth}
    \centering
    \caption{Mean signed error (bias)}
    \begin{tabular}{rrrr}
    \toprule
     & \textbf{MLE} & \textbf{TwoNN} & \textbf{$e\mathcal{DCF}$} \\ \midrule
    \textbf{1k} & 2.500 & 0.667 & -1.542 \\
    \textbf{2k} & 1.917 & 0.625 & -3.250 \\
    \textbf{4k} & 1.458 & 0.083 & -1.000 \\
    \textbf{8k} & 1.125 & -0.250 & -1.917 \\
    \textbf{16k} & 0.667 & -0.750 & -3.125 \\
    \textbf{32k} & 0.542 & -0.833 & -3.125 \\
    \textbf{64k} & -0.125 & -1.500 & -1.875 \\ \bottomrule
    \end{tabular}
\end{subtable}
\caption{Individual estimates and error metrics for 30 \% noise.}
\end{table}

% PAGE 3 FOR CELL 3: FINAL METRIC TABLE
\begin{table}[H]
\centering
\captionsetup{font=small}

\begin{subtable}[t]{0.48\textwidth}
    \centering
    \caption{\% exact matches}
    \begin{tabular}{rrrr}
    \toprule
     & \textbf{MLE} & \textbf{TwoNN} & \textbf{$e\mathcal{DCF}$} \\ \midrule
    \textbf{1k} & 20.8 & 20.8 & 16.7 \\
    \textbf{2k} & 8.3 & 16.7 & 8.3 \\
    \textbf{4k} & 16.7 & 16.7 & 16.7 \\
    \textbf{8k} & 16.7 & 16.7 & 20.8 \\
    \textbf{16k} & 20.8 & 16.7 & 29.2 \\
    \textbf{32k} & 20.8 & 16.7 & 29.2 \\
    \textbf{64k} & 16.7 & 12.5 & 25.0 \\ \bottomrule
    \end{tabular}
\end{subtable}
\caption{Accuracy metric for 30 \% noise.}
\end{table}
\end{document}